\documentclass[lettersize,journal]{IEEEtran}
\usepackage{amsmath,amsfonts}
\usepackage{algorithmic}
\usepackage{algorithm}
\usepackage{array}
\usepackage[caption=false,font=normalsize,labelfont=sf,textfont=sf]{subfig}
\usepackage{textcomp}
\usepackage{stfloats}
\usepackage{url}
\usepackage{verbatim}
\usepackage{graphicx}
\usepackage{cite}
\usepackage{multirow}
\usepackage[table,xcdraw]{xcolor}
\usepackage{makecell}
\begin{document}

\title{Neural Radiance Field-based Visual \\ Rendering: A Comprehensive Review}

\author{Mingyuan Yao,~
        Yukang Huo,~
        Yang Ran,~
        Qingbin Tian,~
        Ruifeng Wang~
        Haihua Wang$^{\S}$
\thanks{This work was supported by key common technologies for high-quality agricultural development (Grant No.21327401D-1) and key technology research and creation of digital fishery intelligent equipment (2021TZXD006).}

\thanks{Mingyuan Yao, Yukang Huo, Qingbin Tian, and Haihua Wang are with the National Innovation Center for Digital Fishery, Beijing 100083, P.R. China. (e-mail: {yaomingyuan, huoyukang, tianqingbin, wanghaihua}@cau.edu.cn).}

\thanks{Mingyuan Yao, Yukang Huo, Qingbin Tian, and Haihua Wang are with the Key Laboratory of Smart Farming Technologies for Aquatic Animal and Livestock, Ministry of Agriculture and Rural Affairs, Beijing 100083, P.R. China. (e-mail: {yaomingyuan, huoyukang, tianqingbin, wanghaihua}@cau.edu.cn).}

\thanks{Yang Ran is with the Key Laboratory of Agricultural Information Acquisition Technology, Ministry of Agriculture and Rural Affairs, China Agricultural University, Beijing 100083, China (e-mail: {ranyang@cau.edu.cn}).}

\thanks{Mingyuan Yao, Yukang Huo, Yang Ran, Qingbin Tian, and Haihua Wang are with the College of Information and Electrical Engineering, China Agricultural University, Beijing 100083, P.R. China. (e-mail: {yaomingyuan, huoyukang, ranyang, tianqingbin, wanghaihua}@cau.edu.cn).}

\thanks{Ruifeng Wang is with the College of Engineering, China Agricultural University, Beijing 100083, China (e-mail: {sweefongreggiewong@cau.edu.cn})}

\thanks{$\S$ denotes corresponding author.(e-mail: {wanghaihua@cau.edu.cn})}}




\maketitle

\begin{abstract}
  In recent years, Neural Radiance Fields (NeRF) has made remarkable progress in the field of computer vision and graphics, providing strong technical support for solving key tasks including 3D scene understanding, new perspective synthesis, human body reconstruction, robotics, and so on, the attention of academics to this research result is growing. As a revolutionary neural implicit field representation, NeRF has caused a continuous research boom in the academic community. Therefore, the purpose of this review is to provide an in-depth analysis of the research literature on NeRF within the past two years, to provide a comprehensive academic perspective for budding researchers. In this paper, the core architecture of NeRF is first elaborated in detail, followed by a discussion of various improvement strategies for NeRF, and case studies of NeRF in diverse application scenarios, demonstrating its practical utility in different domains. In terms of datasets and evaluation metrics, This paper details the key resources needed for NeRF model training. Finally, this paper provides a prospective discussion on the future development trends and potential challenges of NeRF, aiming to provide research inspiration for researchers in the field and to promote the further development of related technologies.
\end{abstract}


\vspace{25pt}

\IEEEraisesectionheading{\section{Introduction}}
\IEEEPARstart{T}{raditional} 3D reconstruction mainly includes Multi-View Stereo(MVS) \cite{paper001}, Light Field Photography \cite{paper002,paper003,paper004}, Structured Light \cite{paper005,paper006}, Laser Scanning, Time-of-Flight (ToF), Volumetric Reconstruction \cite{paper007,paper008,paper009,paper010,paper011,paper012}, Shape from Motion(SfM) \cite{paper013,paper014,paper015,paper016,paper017}, Stereo Vision \cite{paper018,paper019}, Point cloud-based reconstruction \cite{paper020,paper021,paper022}, and other methods. However, these methods still face many unresolved issues, such as sensitivity to data quality, limited generalization ability, blurry new view synthesis, and limitations in complex scene reconstruction.

The NeRF is a deep learning method used for reconstructing three-dimensional scenes and synthesizing new viewpoints. It was first introduced by Mildenhall et al. \cite{paper023} at the ECCV conference in 2020 and quickly became a popular research direction in the field of computer vision. The core idea of NeRF is to use a multi-layer perceptron (MLP) neural network to implicitly represent the radiance field of the three-dimensional scene, allowing for the synthesis of high-quality images from new perspectives.

\begin{figure*}[h!]
\centering
\includegraphics[width=0.9\textwidth]{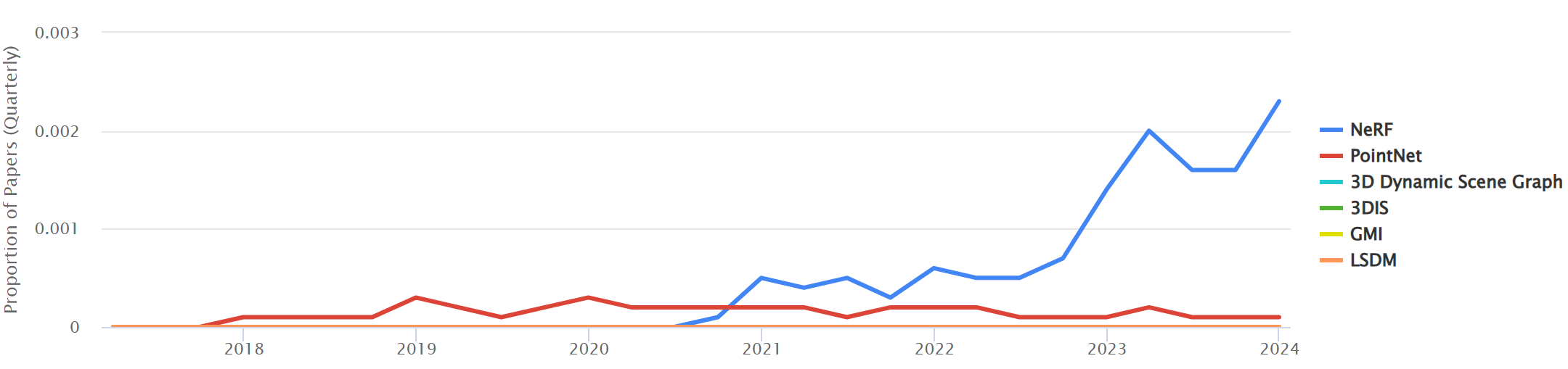}
\caption{Overview of recent high-quality publications of NeRF research papers from PapersWithCode.}
\label{Usage_Over_Time}
\end{figure*}

\begin{figure*}[h!]
\centering
\includegraphics[width=0.9\textwidth]{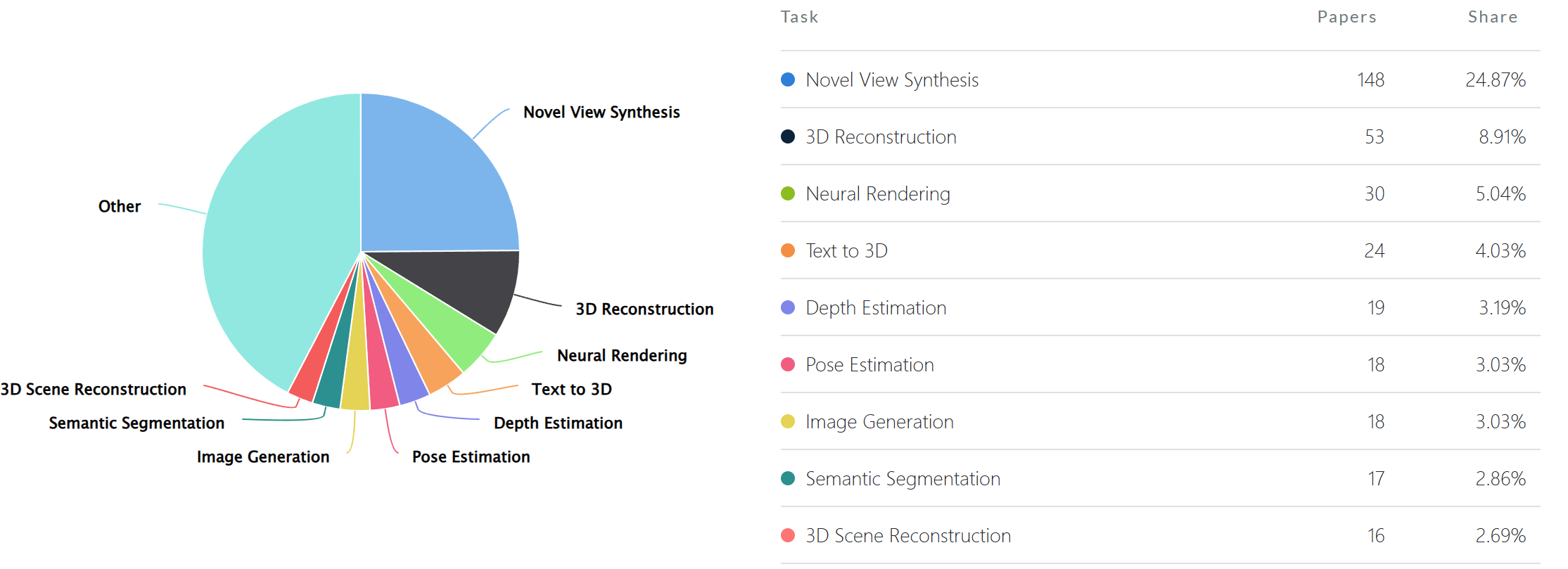}
\caption{Overview of recent high-quality publications of NeRF research papers from PapersWithCode.}
\label{Tasks}
\end{figure*}

With the emergence of NeRF, Neural Volume Representation based on NeRF and other Neural Volume Representation have become a compelling technique for learning to represent 3D scenes from images to render realistic scene images from unobserved viewpoints, with an exponential growth of related articles, as shown in Fig. \ref{Usage_Over_Time}. NeRF is now widely used in Novel View Synthesis, 3D Reconstruction, Neural Rendering, Depth Estimation, Pose Estimation, and other scenarios, as shown in Fig. \ref{Tasks}.


Given the rapid advances based on the NERF method, tracking new research developments is becoming increasingly challenging. Therefore, a comprehensive review of the latest developments in this research field is essential, which will have a positive impact on researchers in this field. This article details the latest advances in NERF. Our main contributions are as follows:

\begin{itemize}
    \item  Section \ref{section:related}: A comprehensive review of the existing NeRF-related literature was first conducted, which included a compendium of earlier work as well as an analysis of recent research trends.
    \item  Section \ref{section:theory}: Detailed descriptions are provided for the elements of the initial NeRF model, encompassing its network structure, loss function, and the method of rendering.
    \item  Section \ref{section:datasets} and \ref{section:evaluation}: Collected and analyzed multiple datasets in detail, summarizing the current commonly used NeRF evaluation metrics.
    \item Section \ref{section:methods} and \ref{section:applications}: We classified the variants of NeRF and detailed their innovations in improving rendering quality, accelerating computation, and their applications in the indoor, outdoor, human body, interactive scenes, etc. We also compared the performance of different models in speed, accuracy, and other key performance metrics such as rendering quality, memory usage, and generalization ability.
    \item  Section \ref{section:discussion}: We have identified the main obstacles in current research, such as the demand for computational resources, model scalability, and the ability to handle complex scenarios. We further explore possible solutions to these challenges and propose potential directions for future research.
    \item  Section \ref{section:conclusion}: We have summarized the main contributions and impacts of NeRF, as well as our outlook on the future development of this field.
\end{itemize}

\section{Related Previous Reviews and Surveys}\label{section:related}
NeRF, transforming the realms of computer vision and 3D reconstruction, provides fresh perspectives on how scenes are implicitly represented and views are formed. Since its introduction, NeRF has been extensively researched and reviewed, enhancing our comprehension of its capabilities and obstacles. Early reviews, such as those by Mildenhall et al. \cite{paper023} (2020) and Tewari et al. \cite{paper024} (2022), provide foundational insights into NeRF principles and initial applications. However, with the rapid advancement in this field, there is an urgent need for a comprehensive analysis. The goal of this segment is to amalgamate earlier analyses and studies, underscore the progression path of NeRF research, and pinpoint the deficiencies that our present study aims to resolve.

Chang et al. \cite{paper025} (2021) explored techniques for synthesizing multiple views using NeRF technology. Initially, the paper presented an overview and fundamental concepts of NeRF, followed by an examination of studies aimed at enhancing and refining NeRF, focusing on its synthesis precision, illustration effectiveness, and the broad applicability of its models. Subsequently, the paper delves into the use of NeRF in intricate scenarios, including the creation of unrestricted image viewpoints, techniques for relighting, and the synthesis of dynamic scene viewpoints. The article concludes by outlining NeRF's prospects and challenges, emphasizing crucial aspects like instantaneous rendering and the clarity of the model.

Gao et al. \cite{paper026} (2022) showcased the latest progress in NeRF technology for three-dimensional vision. Initially, the article examines NeRF's fundamental concepts and instructional techniques, followed by an in-depth analysis of different approaches to enhance the synthesized views' quality, the speed of training, and the generalization capacity of the model. Furthermore, the paper explored how NeRF can be applied in areas like robotic navigation, urban design, self-governing navigation, and virtual and augmented reality, and offered perspectives on upcoming research avenues.

Zhu et al. \cite{paper027} (2023) provided an in-depth analysis of recent research advances on NeRF, a deep neural network for learning and representing objects or scenes with five main features, including volumetric rendering, new view synthesis, decomposable embedding space, multi-view consistency, and weighted importance sampling. The article categorized and discussed NeRF according to the progress it has made in these five aspects, and presented some attempts of NeRF in the innovation of new applications, as well as pointing out the limitations of NeRF research and making suggestions for future research directions.

Croce et al. \cite{paper028} (2023) investigated the capabilities of NeRF applications within the realm of digital cultural heritage. The paper juxtaposed NeRF against conventional photometric methods and explored NeRF's benefits in handling complex items like metals, semi-transparent or clear surfaces, uniform textures, blockages, and intricate details. The article, using case studies, highlighted NeRF's capabilities in digitizing cultural heritage, particularly in handling items challenging to process via conventional methods. Ultimately, the paper delved into the prospective trajectory of NeRF in digitizing cultural heritage.

Cheng et al. \cite{paper029} (2023) investigated how NeRF technology can be utilized in self-driving situations. This paper delved into the difficulties faced by NeRF in autonomous driving, focusing on issues like sparse view reconstruction, extensive scene processing, dynamic scene modeling, and enhancing training speed, while also examining proposed solutions from current studies. Furthermore, the paper explored the prospective advancements in NeRF technology, focusing on enhancing model generalization and integrating conventional 3D reconstruction methods, multi-sensor integration, and semantic information fusion.

Remondino et al. \cite{paper030} (2023) conducted a thorough examination of NeRF-based techniques for 3D image reconstruction, contrasting them with traditional photometric methods. Scientists assessed NeRF's efficacy in handling objects varying in size and surface features, such as those without texture, metal, transparency, and transparency. Additionally, the article explored the practical use of NeRF in real-life situations and outlined potential research pathways.

Yang et al. \cite{paper031} (2024) investigated how NeRF could be applied in systems for simultaneous localization and mapping (SLAM). The article analyzed the development history of NeRF and SLAM, described NeRF-based SLAM approaches including different strategies for mapping and tracking phases, and discusses current challenges and future research directions. The researchers evaluated the performance of these approaches through experiments and datasets, pointing out key issues such as dynamic environment modeling, trade-offs between SDF and bulk density, and loop detection and localization performance improvement. The article provided a valuable reference for future research in this area.

Although the earlier referenced review articles provide an in-depth analysis of NeRF technology, their limited time availability leads to insufficient documentation of the latest research outcomes. Given this, this paper aims to fill this gap by systematically organizing and reviewing the latest research developments in the field of NeRF since March 2024. We will explore in detail the innovative applications of NeRF technology in various application scenarios, as well as the latest advances in algorithm optimization, model generalization capability, and computational efficiency. In addition, this paper will provide an in-depth analysis of the major challenges facing NeRF technology and look forward to its potential directions in future research, to provide comprehensive references and inspiration for researchers in both academia and industry.

\section{NeRF Theory Explanation}\label{section:theory}
The NeRF function is a representation of a continuous scene as a function whose input is a 5D vector $(X,d)$, The output is the viewpoint-dependent color c of this 3D point and the density $\sigma$ of the corresponding position (voxel).
\begin{equation}\label{eq:nerf}
    F_{\Theta}:(X, d) \rightarrow(c, \sigma)
\end{equation}
The inputs to the above equation are a 5D vector $(X,d)$, Here $X$ represents the 3D coordinates of the camera observation $(x, y, z)$, $d$ represents a 3D Cartesian unit vector in the camera observation direction $(\Theta, \Phi)$, and the output is a 4d vector where c represents the color values $(r, g, b)$ and $\sigma$ represents the volume density. This 5D function is approximated by one or more multilayer perceptrons (MLPs), denoted as $F_{\Theta}$. since voxel density is only related to spatial location, while color is related to spatial location as well as the viewing angle of observation. Therefore again in specific implementations, a two-layer MLP is usually used for training, where $X$ is first input to the MLP network and outputs $\sigma$ and a 256-dimensional intermediate feature, followed by the intermediate feature and d which are then input together to an additional fully connected layer (128 dimensions) to predict the color, as shown in Fig. \ref{nerf_network} below.

\begin{figure}[h!]
    \centering
    \includegraphics[width=0.45\textwidth]{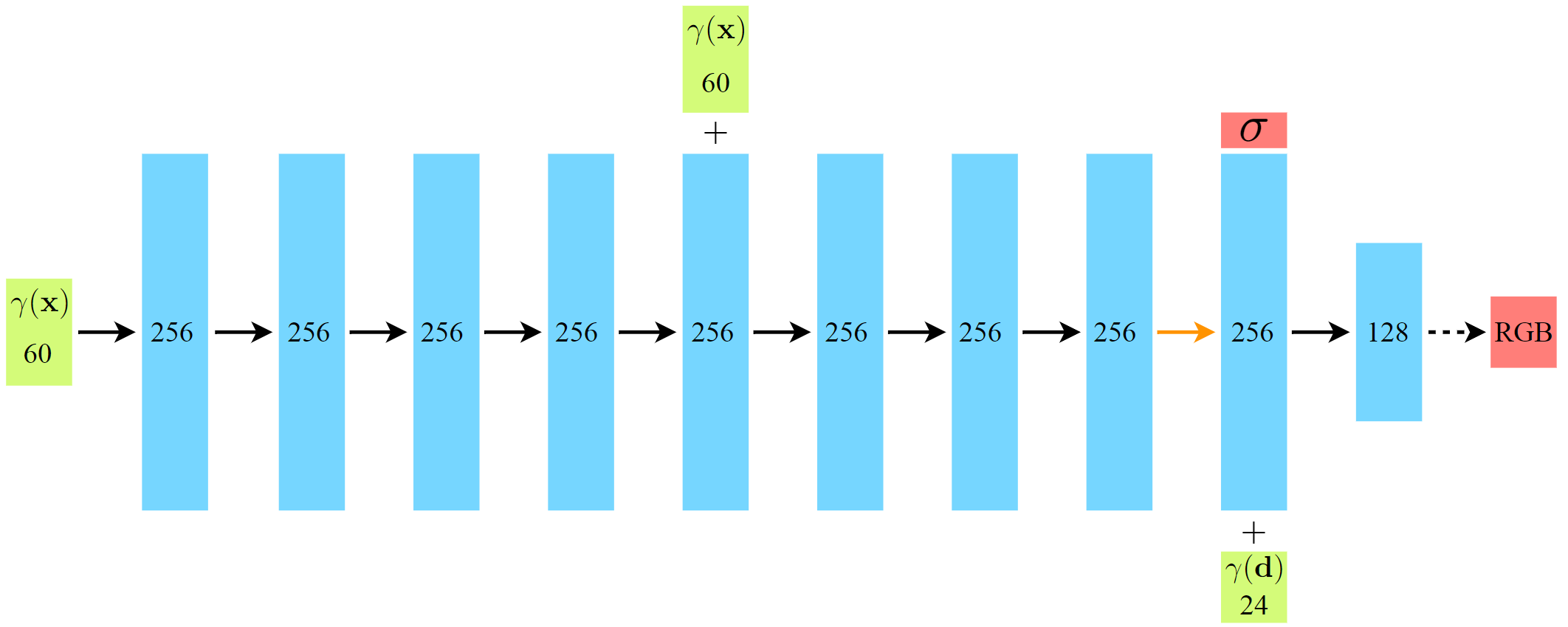}
    \caption{NeRF Network Architecture.}
    \label{nerf_network}
\end{figure}

NeRF renders the color of any light rays passing through the scene by representing the scene as the bulk density and directional emission radiation at any point in space, based on the classical body rendering technique \cite{paper032}, as shown in the implementation flow in Fig. \ref{nerf_rendering_process}:

\begin{figure}[h!]
    \centering
    \includegraphics[width=0.45\textwidth]{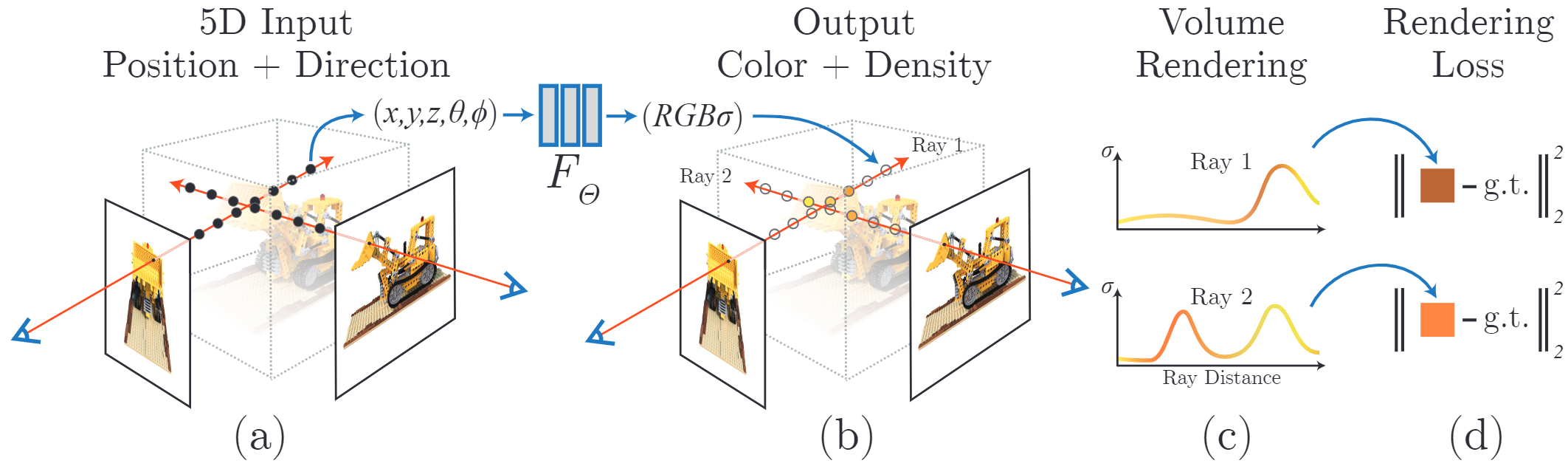}
    \caption{NeRF Rendering Process.}
    \label{nerf_rendering_process}
\end{figure}

To obtain a recognized perspective, the writer projects beams through each two-dimensional point, subsequently sampling various points along the direction of the ray (namely, the depth direction) for the NeRF network's input, and transforms the resulting voxel density $\sigma$ and the color value $c=(r, g, b)$ related to direction into a two-dimensional image that aligns with the ultimate viewing angle. Optimal outcomes are attained through the refinement of the rendering loss. Within the NeRF structure, the methodology for creating new perspectives can be encapsulated thus:

\begin{itemize}
    \item Light sampling: For each pixel in the target image, a light ray is generated through the pixel based on camera parameters and scene geometry. Along this ray, a series of spatial points are systematically sampled, which constitute a discrete representation of the ray's interaction with the scene. (see (a) in Fig. \ref{nerf_rendering_process}).
    \item For each sampled point, a trained NeRF multilayer perceptron (MLP) network is used to predict the color and bulk density of the sampled point based on the spatial coordinates and viewing direction of that point. degree (as shown in (b) in Fig. \ref{nerf_rendering_process}).
    \item The color and density information of all sampled points are accumulated to synthesize the color of the final pixel by the volume rendering technique (shown as (c)(d) in Fig. \ref{nerf_rendering_process}).
\end{itemize}

\subsection{Traditional Volume Rendering Technology}
In traditional volume rendering technology, given the volume density and color distribution of the scene, the observation ray of the camera in any direction can be defined. The ray $r(t)=o+t d$ is defined by the starting point o of the ray and the direction vector $d$, where t extends from the proximal boundary $t_n$ to the distal boundary $t_f$ The color on the ray can be expressed by integral to continuously accumulate the radiance along the ray path:
\begin{equation}
    \begin{gathered}
        C(\mathbf{r})=\int_{t_{\mathrm{n}}}^{t_f} T(t) \sigma(\mathbf{r}(t)) \mathbf{c}(\mathbf{r}(t), \mathbf{d}) d t, \\
        \quad \text { where } T(t)=\exp \left(\int_{t_{\mathrm{n}}}^t \sigma(r(s)) d s\right)
    \end{gathered}
\end{equation}
In the continuous radiation field, the integral $C(r)$ is used to calculate the cumulative radiance along the light path of the virtual camera to render the image from a specific perspective. The function $T(t)$ describes the cumulative transmittance from the ray starting point $t_n$ to $t$ reflecting the probability that the light does not interact with the medium during travel. The volume density $\sigma(r(t))$ and the color $c(r(t),d)$ represent the spatial density and color of the points $c(r(t),d)$ on the light, respectively, while $d_{t}$ represents the small displacement of the light during the integration process. By performing this integration process along the ray of each pixel, the color of the pixel can be determined, thereby synthesizing the image of the target perspective.
\subsection{Fragmental Approximate Volume Rendering Method}
Since MLP can only query at a fixed set of offline locations, the authors propose a method of Stratified Sampling: first, dividing the region $[t_n,t_f]$ that needs to be integrated into   parts uniformly, and in order to avoid limiting the resolution of NeRF due to the frequency of sampling points as much as possible, then uniformly randomly sampling in each small region, instead of uniformly sampling, the corresponding formula is shown below:
\begin{equation}
    t_{i}\sim\mathcal{U}\bigg[t_{n}+\frac{i-1}{N}(t_{f}-t_{n}),t_{n}+\frac{i}{N}\left(t_{f}-t_{n}\right)\bigg]
\end{equation}
Based on these sampling points, the above integrals can be reduced to the form of summation, and since the method of obtaining the ray rendering color by summation from the set of $(c_i,\sigma_i)$ of all sampling points is also differentiable, it can be simplified to be replaced by the traditional transparency blending algorithm $\alpha_{i}$. The relevant formulas are shown below:
\begin{equation}
    \hat{C}(r)=\sum_{i=1}^NT_i\alpha_ic_i
\end{equation}
\begin
{equation}T_{i}=\exp\left(-\sum_{j=1}^{i-1}\sigma_{j}\delta_{j}\right)\end{equation}
\begin{equation}
    \alpha_i=1-\exp\left(-\sigma_i\delta_i\right)
\end{equation}
where $\delta_{i}=t_{i+1}-t_{i}$ is the distance between neighboring sampling points of the camera ray.

\subsection{Position Encoding}
In order to enable the NeRF neural network to better capture and represent the high-frequency details in the scene, NeRF maps the 5D inputs into a high-dimensional space using positive cosine positional coding, which is formulated as shown below:
\begin{equation}
    \begin{gathered}
        \gamma(v)=(\sin{(2^0\pi v)},\cos{(2^0\pi v)},\sin{(2^1\pi v)},\cos{(2^1\pi v)},\\\cdots,\sin{(2^{N-1}\pi v)},\cos{(2^{N-1}\pi v)})
    \end{gathered}
\end{equation}
where $v$ is the normalized coordinate value and $N$ is the dimension of the position encoding (in the original paper, the $L$ value was 10 for the 3D coordinate $X$ and 4 for the 2D view direction $d$).

The positional encoding helps the MLP to better approximate functions with high-frequency variations, which is crucial for capturing fine textures and details in the scene. In this way, NeRF can generate high-quality images of new viewpoints while maintaining the continuity and consistency of the scene.

\subsection{Multi-Stage Voxel Sampling and Loss Optimization}
NeRF's approach to rendering involves repeatedly sampling each camera ray, typically N times, to create an authentic image. To deduce the ray's color, the hues of the sampled points are combined and adjusted for weight. Yet, this approach falls short in efficiency when handling numerous vacant or hidden areas, which have a negligible impact on the ultimate image yet are essential for the computational process.

To enhance the efficiency of rendering, scientists have suggested a layered voxel sampling approach, namely, a multi-tiered sampling technique ranging from rough to detailed. Initially, this technique gathers a collection of sparse sampling points (labeled $N_c$) evenly across the scene, followed by calculating the rendering outcomes of the coarse network using these points. More precisely, the discrete summation function can be reformulated as:
\begin{equation}
    \hat{C}_{c}(\mathbf{r})=\sum_{N_{\epsilon}}^{i=1}w_{i}c_{i},\quad w_{i}=T_{i}(1-\exp{(-\sigma_{i}\delta_{i})})
\end{equation}
where $\hat{C}_c(\mathbf{r})$ is the rough color estimate of the ray $r$, $c_i$ is the color of the $i$th sampling point, and $w_i$ is the weight of each rough sampling point, which is computed from the cumulative transmittance $T_i$ and the bulk density $\sigma_i$. This is normalized so that the weights $\hat{w}_{i}$ form a segmented constant probability density function (PDF) along the ray:
\begin{equation}
    \hat{w}_i=w_i/\sum_{N_c}^{j=1}w_j
\end{equation}
Next, $N_{f}$ additional fine sampling points are sampled from this distribution using the Inverse Transform Sampling method, based on the PDF obtained from the coarse sampling. These points are mainly concentrated in areas expected to contain visible content. The "fine" network is then evaluated on the combined set of coarse and fine sampled points to obtain more detailed color and volume density information. Finally, the final rendered color $\hat{C}_{f}(r)$ of the ray is calculated using all $N_c+N_f$ sample points:
\begin{equation}
    \hat{C}_{f}\left(r\right)=\sum_{i=1}^{N_{\epsilon}+N_{\ell}}\left(\hat{w}_{i}\cdot c_{i}\right)
\end{equation}
where $\hat{w}_{i}$ is the normalized weight and $c_{i}$ is the color of the sampled point. The loss function is the total squared error between the rendered color and the true pixel color, including the results of rough and fine rendering, with the following formula:
\begin{equation}
    L=\sum_{r\in R}\left[\parallel\hat{C}_{c}(r)-C(r)\parallel_{2}^{2}+\parallel\hat{C}_{f}(r)-C(r)\parallel_{2}^{2}\right]
\end{equation}
where $R$ is the set of rays in each batch, and $C(r)$, $\hat{C}_c(r)$, and $\hat{C}_f(r)$ are the true RGB color, rough volume-predicted color, and fine volume-predicted color of ray r, respectively. It is worth noting that while the final rendering comes from the output of the fine network $\hat{C}_f(r)$, the loss of the coarse network $\hat{C}_c(r)$ is also minimized so that the weight distribution of the coarse network can be used to distribute samples in the fine network.

In summary, NeRF acquires an ongoing volumetric depiction of the scene through neural network optimization, utilizes hierarchical sampling and volumetric rendering methods to create images from a novel viewpoint, and applies a pixel-level loss function to steer the network's training, aiming to reduce the disparity between the actual and rendered images, thereby enabling NeRF to produce superior images in this updated perspective.

\section{NeRF Common Datasets}\label{section:datasets}
This part will thoroughly examine the datasets intimately linked to the training of NeRF models. For the successful training and assessment of NeRF models, scientists depend on a range of meticulously crafted artificial datasets alongside actual image compilations. The datasets serve a dual purpose: offering abundant training examples for the models and facilitating a more uniform assessment of their performance. Below is an in-depth summary of these crucial datasets:

\subsection{Synthetic Datasets}
{\bf{NeRF synthetic dataset(Blender dataset)}} \cite{paper023}: Proposed in the initial NERF paper, the dataset comprises intricate 3D scenarios crafted using Blender software, featuring a variety of items like chairs, drums, plants, and more. Additionally, images of high resolution (800x800 pixels) are supplied, with every scene furnished with an appropriate collection of images for training, validation, and testing. Furthermore, the dataset includes depth and normal maps, along with comprehensive camera transformation data, offering essential geometric and lighting details for training the NeRF model.

{\bf{Local Light Field Fusion (LLFF) dataset}} \cite{paper033}: This serves as a tool for innovative view synthesis studies, merging artificial and actual images, aimed at facilitating the depiction of intricate scenarios in virtual exploration. The dataset comprises artificial images created using SUNCG \cite{paper034} and UnrealCV \cite{paper035}, along with 24 authentic scene photos taken from a portable mobile phone. The LLFF dataset is apt for broad new view synthesis activities and is apt for the training and assessment of deep learning models, particularly in managing the synthesis of new views in real-world scenarios. Furthermore, LLFF offers an effective algorithm for view synthesis, creating novel perspectives by merging multi-plane image (MPI) scene depictions with local light fields, thereby broadening the conventional plenoptic sampling theory.

{\bf{Mip-NeRF Synthetic 360° dataset (NeRF-360-V2 dataset)}} \cite{paper036}: This dataset is a synthetic dataset extended from Mip-NeRF\cite{paper146}, is crafted to address the challenge of 3D reconstruction in scenarios with limitless possibilities. The dataset addresses the difficulties of limitless scenes by employing nonlinear scene parameterization, real-time distillation, and innovative distortion-oriented regularization techniques. The Mip-NeRF 360 has the ability to create lifelike artificial perspectives and intricate depth charts for extremely intricate, limitless real-life scenarios. Within the dataset, there are 9 scenarios, split evenly between outdoor and indoor settings, each featuring a multifaceted main object or space, along with an intricate backdrop.

{\bf{NVS-RGBD dataset}} \cite{paper037}: Includes rough depth charts of real-world scenes recorded by consumer-level depth sensors. The goal of this dataset is to establish a novel standard for NeRF in assessing the efficacy of creating new perspectives using a limited set of views. Comprising 8 scenes, the NVS-RGBD dataset gathers rough depth maps from consumer-grade sensors like Azure Kinect, ZED 2, and iPhone 13 Pro. The artifacts in these depth maps might vary from those in sensor noise.

{\bf{DONeRF dataset}} \cite{paper038}: The dataset encompasses a variety of three-dimensional settings, including bulldozers, woodlands, educational spaces, San Miguel, kiosks, and hairdressing establishments, among others. A range of developers created these situations using Blender, establishing a practical foundation for studying neural radiation fields and resources, particularly for instantaneous rendering and interactive uses.

\subsection{Real-world Datasets}
{\bf{Tanks and Temples dataset}} \cite{paper039}: The dataset includes standard sequences gathered from outside the lab in actual conditions, offering high-definition video footage of both indoor and outdoor environments for input. The video sequences facilitate the creation of innovative pipelines that utilize video inputs to enhance the accuracy of reconstruction. An industrial laser scanner is used to gather the dataset's authentic data, encompassing scenes from both the outdoor and indoor settings. Additionally, this dataset offers both training and testing datasets, segmenting the test data into intermediate and advanced categories to suit reconstruction activities of varying complexity.

{\bf{DTU dataset}} \cite{paper040}: The dataset presents a multi-view stereo format, featuring a tenfold increase in scenes compared to its predecessors and a notable rise in diversity. More precisely, it comprises 80 scenes with a wide range of diversity. Every scene is composed of either 49 or 64 accurate camera placements and structured light reference scans, resulting in RGB imagery with a resolution of 1200×1600 pixels.

{\bf{Euroc dataset}}\cite{paper041}: The dataset encompasses both indoor and outdoor data, featuring an array of sensor information, such as camera and IMU readings. The data set finds extensive application in various research areas, including robot vision, determining camera angles, calibrating cameras, and positioning and navigation. The primary characteristics of this approach are its ability to deliver sensor data with high precision and authentic indoor settings, aiding in assessing our method's reconstruction and localization precision through the use of grayscale imagery and closely integrated IMU measurements.

{\bf{Replica dataset}} \cite{paper042}: The dataset represents a superior 3D reconstruction of indoor scenes, crafted by Facebook. The collection includes 18 intricately lifelike indoor settings, each meticulously crafted and depicted to maintain visual realism. Every dataset scene features a compact three-dimensional mesh, detailed high dynamic range   (HDR) textures, data on glass and mirror surfaces, along semantic classifications and segmentation of instances.

{\bf{BlendedMVS dataset}} \cite{paper043}: This extensive dataset is tailored for multi-view stereo matching   (MVS) networks, to supply ample training instances to facilitate algorithms based on learning MVS. The BlendedMVS collection boasts over 17,000 detailed images, encompassing diverse landscapes like urban areas, structures, sculptures, and miniature items. The vastness and variety of this dataset render it a crucial asset for MVS studies.

{\bf{Amazon Berkeley Objects dataset(ABO dataset)}} \cite{paper044}: The dataset serves as an extensive 3D object comprehension collection, crafted to connect the realms of reality and virtual 3D. The dataset comprises around 147,702 listings of products, each linked to 398,212 distinct images in the catalog, with every product possessing as many as 18 unique metadata features including category, color, material, weight, and size. Included in the ABO dataset are 360-degree images of 8,222 items and artist-crafted 3D mesh representations of 7,953 products. The dataset is perfectly suited for 3D reconstruction, estimating materials, and retrieving multi-view objects across domains, as these 3D models feature intricate geometrical designs and incorporate materials based on physical properties.

{\bf{Common Objects in 3D dataset(CO3Dv2 dataset)}} \cite{paper045}: The dataset comprises 1.5 million multi-view image frames, spanning 50 MS-COCO categories, offering abundant image resources, precise camera positions, and 3D point cloud annotations. The vastness and variety of CO3Dv2 render it perfect for assessing innovative view synthesis and 3D reconstruction techniques, propelling advancements in 3D computer vision research.

{\bf{3D-FRONT dataset}} \cite{paper046}: An extensive artificial indoor scene dataset collaboratively created by Alibaba's Taobao Technology Department, Simon Fraser University, and the Chinese Academy of Sciences' Institute of Computing Technology. The dataset offers expertly crafted room designs along with an extensive array of 3D models that are both style-compatible and of high quality. The 3D-FRONT facility houses 18,797 rooms, each uniquely organized with 3D elements, alongside 7,302 furniture pieces featuring superior textures. The dataset's characteristics encompass an extensive range from the semantics of layouts to the intricate textures of each object, designed to aid studies in fields like 3D scene comprehension, SLAM, and the reconstruction and segmentation of 3D scenes. Furthermore, the dataset includes Trescope, a streamlined rendering tool, to facilitate the fundamental rendering of 2D images and their annotations.

{\bf{SceneNet RGB-D dataset}} \cite{paper047}: This dataset is a collection of 5 million photorealistic images of synthetic indoor scenes with corresponding ground truth data. The scenes in the dataset are randomly generated and contain 255 different categories, which are usually regrouped into 13 categories similar to the NYUv2 dataset. These synthetic scenes provide rich viewpoint and illumination variations, making the dataset ideal for indoor scene understanding tasks such as semantic segmentation, instance segmentation, object detection, and geometric computer vision such as optical flow, depth estimation, camera pose estimation, and 3D reconstruction. question.

\subsection{Face Datasets}
{\bf{CelebV-HQ dataset}} \cite{paper048}: An extensive, superior, and varied video collection, elaborately marked with facial features, encompassing 35,666 clips with a minimum resolution of 512x512, featuring 15,653 different identities. Each video clip is hand-tagged with 83 distinct facial characteristics, encompassing looks, motion, and feelings, to serve as a valuable asset in research fields like facial identification, expression study, and video comprehension.

{\bf{CelebAMask-HQ dataset}} \cite{paper049}: The dataset in question is an extensive, high-definition facial image collection, featuring 30,000 images chosen from the CelebA dataset. Every picture is equipped with a segmentation mask measuring 512*512 pixels. Researchers receive detailed facial region data by manually labeling these masks with 19 types of facial features such as skin, eyes, nose, mouth, and more.

{\bf{VoxCeleb dataset}} \cite{paper050}: This dataset is a large-scale speaker recognition dataset developed by researchers at the University of Oxford. It contains approximately 100,000 voice clips of 1,251 celebrities from YouTube videos. The VoxCeleb dataset is designed to support research on speaker identification and verification, providing a realistic, diverse, and large-scale data resource. The speech clips in the data set cover different ages, genders, accents, and occupations, as well as a variety of different recording environments and background noise. VoxCeleb is divided into two subsets: VoxCeleb1 and VoxCeleb2. The audio sampling rate of the data set is 16kHz, 16bit, mono, PCM-WAV format.

{\bf{Labeled Faces in the Wild (LFW) datasets}} \cite{paper051}: The dataset in question is publicly accessible and extensively utilized in facial recognition studies. It was compiled by the Computer Vision Laboratory at the University of Massachusetts Amherst and contains more than 13,000 face images collected from the Internet. The images cover 1,680 different individuals, with at least two images for each person. The purpose of the LFW dataset is to improve the accuracy of face recognition under natural conditions, so it contains face images taken in a variety of different environments, such as different lighting, expressions, postures, and occlusion situations.

{\bf{MPIIGaze dataset}} \cite{paper052}: The dataset was collected by 15 users over several months of daily laptop use and contains 213,659 full-face images and their corresponding true gaze locations. Experienced sampling techniques guarantee consistent gaze and head positions, along with authentic alterations in the appearance and lighting of the eyes. To facilitate cross-dataset evaluation, 37,667 images were manually annotated with eye corners, mouth corners, and pupil centers. The dataset stands out for its variety in personal looks, surroundings, and photographic equipment, coupled with an extended duration of data gathering, offering a crucial asset for examining the broad applicability of gaze estimation techniques.

{\bf{GazeCapture dataset}} \cite{paper053}: The dataset is a large dataset for eye-tracking technology, containing approximately 2.5 million frames of images from more than 1,450 volunteers. Gathered via mobile devices, this dataset aims to aid in eye-tracking studies and train associated convolutional neural networks (CNNs), like iTracker.Characteristics of the GazeCapture dataset include its scalability, trustworthiness, and variability, which ensure data diversity and quality.

{\bf{Flickr-Faces-HQ (FFHQ) dataset}} \cite{paper054}: This collection of facial images is of superior quality, comprising 70,000 images in PNG format, each with a resolution of 1024*1024. FFHQ encompasses a diverse range of age groups, ethnicities, and cultural heritages, along with an assortment of accessories like eyewear, sunglasses, hats, and more, offering extensive variety.

\subsection{Human Datasets}
{\bf{Thuman dataset}} \cite{paper055}: The dataset represents an extensive public collection for 3D human body reconstruction, encompassing around 7,000 data points.Each data item includes a surface mesh model with materials, RGBD images, and the corresponding SMPL model. Contains human models in a variety of poses and costumes, captured and reconstructed using DoubleFusion technology. The release of the data set provides valuable resources for research in 3D human body modeling, virtual reality, augmented reality, and other fields.
{\bf{HuMMan dataset}} \cite{paper056}: The HuMMan dataset is a large-scale multi-modal 4D human body dataset containing 1,000 human subjects, 400,000 sequences, and 60 million frames of data. Features of this dataset include multi-modal data and annotations (such as color images, point clouds, key points, SMPL parameters, and textured mesh models), a sensor suite including popular mobile devices, and a sensor suite designed to cover basic motion. 500 action collections, and supports a variety of tasks such as action recognition, pose estimation, parametric human body restoration, and texture mesh reconstruction. The HuMMan dataset is designed to support diverse sensing and modeling research, including challenges such as fine-grained action recognition, dynamic human mesh sequence reconstruction, point cloud-based parametric human estimation, and cross-device domain gaps.

{\bf{H36M dataset}} \cite{paper057}: The Human3.6M dataset is a widely used 3D human pose estimation research dataset. Showcasing approximately 3.6 million images, the collection displays 11 artists (6 men and 5 women) participating in 15 standard activities in 7 different situations, such as walking, eating, and talking, among others. Simultaneously, the data was recorded using 4 high-resolution cameras and a rapid motion capture system, offering accurate information on 3D joint positions and angles. Each actor's BMI ranges from 17 to 29, ensuring diversity in body shapes.

{\bf{Multi-Garment dataset}} \cite{paper058}: The dataset for reconstructing 3D clothing includes 356 images, each depicting individuals with diverse body shapes, poses, and clothing styles. Originating from authentic scans, it offers 2078 reconstructed models based on real clothing, encompassing 10 categories and 563 instances of clothing. Each garment in the dataset is richly annotated, including 3D feature lines (such as collar, cuff outline, hem, etc.), 3D body pose, and corresponding multi-view real images.

{\bf{MARS dataset}} \cite{paper059}: The dataset under review is a comprehensive video-based person re-identification (ReID) compilation, featuring 1,261 unique pedestrians, captured by six cameras operating nearly concurrently, each snapped by at least two cameras. Characteristics of the MARS dataset include changes in walking posture, clothing color, and lighting, coupled with less-than-ideal image sharpness, making its recognition more challenging. Moreover, the dataset comprises 3,248 distractors to mimic the intricacies of real-life scenarios.

\subsection{Other Datasets}
{\bf{InterHand2.6M dataset}} \cite{paper060}: This dataset is a large-scale gesture recognition dataset that contains over 2.6 million gesture instances captured by 21 different people in a controlled environment. The data set provides annotations for 21 gesture categories, including common gestures such as fists, palm spread, thumbs up, etc. Each gesture has multiple variations, such as different hand postures, backgrounds, and lighting conditions. The InterHand2.6M dataset is designed to support the development and evaluation of hand gesture recognition algorithms, especially in complex scenes and diverse gesture expressions.

{\bf{TartanAir dataset}} \cite{paper061}: This dataset was developed by Carnegie Mellon University to challenge and advance the limits of visual SLAM technology. The dataset is generated in a highly realistic simulated environment, containing diverse lighting, weather conditions, and moving objects to simulate the complexity of the real world. TartanAir provides rich multi-modal sensor data, including RGB stereo images, depth images, segmentation labels, optical flow, and camera pose information. This data helps researchers develop and test SLAM algorithms, especially when dealing with challenging scenarios.

{\bf{SUN3D dataset}} \cite{paper062}: The assortment features an extensive range of RGB-D videos, displaying scenes from various sites and structures. The dataset includes 415 sequences recorded across 254 different locations and 41 unique structures, with each frame detailing the semantic division of objects in the scene and the camera's location.

\section{NeRF Commonly Used Evaluation Indicators}\label{section:evaluation}
In the initial NeRF research, the researchers employed diverse quantitative measures to assess the model's efficacy in the novel view synthesis task. The metrics encompass Peak Signal-to-Noise Ratio (PSNR), Learned Perceptual Image Block Similarity \cite{paper063} (LPIPS), and Structural Similarity \cite{paper064} (SSIM), assessing the disparity in quality between the created image and the intended image from various angles.

Beyond these metrics for evaluating image quality, scientists have developed Absolute Trajectory Error \cite{paper065} (ATE) and Relative Pose Error \cite{paper065} (RPE) to measure the precision of poses in 3D synthesized views. Furthermore, the DISTS evaluation metric \cite{paper066} offers a method based on distance to gauge the disparities between artificial and actual images. Conversely, Mean Angular Error (MAE°) \cite{paper067} emphasizes the measurement of angular error. Negative Log Likelihood (NLL) \cite{paper068,paper069,paper070,paper071} assesses metrics that quantify the uncertainty in the model's predictions by determining the variance between the forecasted and actual distributions. S3IM \cite{paper072} appraises metrics for gauging the resemblance between two-pixel groups and gathers non-local structural similarity data from randomly selected pixels.

Integrating these evaluation metrics yields a thorough analysis of the NeRF model's efficacy, focusing not just on image clarity but also on the precision of pose and the unpredictability of predictions in three-dimensional space. The aforementioned metrics are elaborated upon in the following manner:

\subsection{Peak Signal-to-Noise Ratio (PSNR)}
PSNR quantifies the variance between an image with noise and one that is clear, and commonly employed to assess the impact of image processing activities like compression, noise reduction, and enhanced resolution. The computation of PSNR involves contrasting the noise-free original image with its processed counterpart, which might include noise. As the value increases, the quality of the image improves while minimizing distortion. PSNR computation relies on the mean square error(MSE), a metric that calculates the mean squared variance in pixel values between the initial and processed images. The calculation formula is:
\begin{equation}
    MSE=\frac{1}{mn}\sum_{m-1}^{i=0}\sum_{n-1}^{j=0}[I(i,j)-K(i,j)]^{2}
\end{equation}
Among them, $I$ refers to the clean image, $K$ refers to the noisy image, $i$, $j$ represent the pixel coordinates, the size of the image is $m*n$, and the calculation formula of PSNR is:
\begin{equation}
    PSNR=10\cdot\log_{10}\left(\frac{MAX_I^2}{MSE}\right)
\end{equation}
where $MAX$ is the maximum value of the dynamic range, that is the maximum value of the pixel range at a certain point $(i,j)$. The value of PSNR is usually expressed in decibels (dB), which provides a quantitative way to evaluate image quality. In practical applications, the higher PSNR value, usually means that the image quality is closer to the original image and the distortion is smaller. However, PSNR does not always fully reflect the human eye's perception of image quality because it is based on mathematical calculations of pixel values rather than visual perception. Despite this, PSNR is still a widely used metric in image quality assessment.

\subsection{Learning Perceptual Image Patch Similarity (LPIPS)}
Learning Perceptual Image Patch Similarity (LPIPS) is a deep learning-driven method used to evaluate image quality. LPIPS trains a deep neural network to learn feature representations of image patches and calculates similarity between images based on these features. This approach aims to more accurately reflect the human visual system's perception of image quality.

The calculation process of LPIPS usually includes the following steps:
\begin{enumerate}
    \item Feature extraction: Use a pre-trained deep convolutional neural network (such as VGG16) to extract features of image patches.
    \item Feature comparison: Calculate the difference between two sets of features, which usually involves Euclidean distance or other similarity measures of feature vectors.
    \item Similarity Score: Merging disparities in features into a unified score, a lower score indicates greater similarity between the two images.
\end{enumerate}

The core idea of LPIPS is to use deep neural networks (e.g. VGG networks) to extract high-level features of an image and then compute the differences between these features. Compared to traditional image quality assessment metrics (e.g., PSNR and SSIM), LPIPS can better capture the perceived quality of an image because it takes into account the deep semantic information of the image.

LPIPS is widely used in the field of image processing, especially in Generative Adversarial Networks (GANs) and image generation tasks, to evaluate the similarity between a generated image and a target image. It is capable of revealing subtle differences in images that may not be easily detected in traditional pixel-level evaluations.

\subsection{Structural Similarity (SSIM)}
Structural Similarity Index Measure (SSIM) is a metric used to measure the similarity between two images. It aims to more accurately reflect the perception of image quality by the Human Visual System (HVS). Compared to traditional pixel-level difference measures (such as Mean Squared Error MSE), SSIM takes into account three dimensions of the image: luminance, contrast, and structure, to comprehensively evaluate image quality. The calculation formula is shown below:
\begin{equation}
    SSIM\left(x,y\right.)=\frac{\left(2\mu_x\mu_y+c_1\right)\left(2\sigma_{xy}+c_2\right)}{\left(\mu_x^2+\mu_y^2+c_1\right)\left(\sigma_x^2+\sigma_y^2+c_2\right)}
\end{equation}
where $\mu_{x}$ and $\mu_{y}$ are the means of the images $x$ and $y$, respectively, $\sigma_x^2$ and $\sigma_{y}^2$ are the variances of the images $x$ and $y$, respectively, and $\sigma_{xy}$ is the covariance of $x$ and $y$. $c_{1}$ and $c_{2}$ are constants added to avoid zero denominators and usually take the values of $\left(k_1L\right)^2$ and $(k_{2}L)^{2}$. $L$ is the dynamic range of the pixel (for an 8-bit image, $L=255$), $k_{1}=0.01$ and $k_{2}=0.03$.

The SSIM value ranges from -1 to 1, where 1 indicates that two images are identical, 0 means no correlation, and negative values imply that images x and y are structurally dissimilar. SSIM is commonly used in image compression, transmission, encoding, and processing to assess image quality loss. In practical applications, SSIM is usually calculated in the form of local windows, where SSIM values are calculated for different regions of the image and then averaged, known as Multi-Scale SSIM (MS-SSIM). MS-SSIM considers the local characteristics of the image, aligning more with the human visual system's local features.

\subsection{Absolute Trajectory Error (ATE) and Relative Pose Error (RPE)}
Absolute Trajectory Error (ATE, Absolute Trajectory Error) and Relative Pose Error (RPE, Relative Pose Error) are two important metrics for evaluating the performance of visual SLAM (Simultaneous Localization and Mapping) systems.
ATE measures the difference between the trajectory estimated by the SLAM system and the true trajectory. It directly calculates the difference between the true value of the camera's position and the SLAM system's estimated value.ATE is commonly used to evaluate the overall performance of a SLAM system, as it reflects the global consistency of the entire trajectory. To compute ATE, the true and estimated trajectories need to be aligned first, which is usually achieved by least squares or other alignment algorithms. The formula for its calculation is:
\begin{equation}
    F_{i}=(Q_{i}^{-1}SP_{i})
\end{equation}
where $Q_i^{-1}$ is the true position of frame $i$, ${p}_{i}$ is the estimated bit position of frame $i$, and $S$ is the aligned matrix.

RPE measures the error in the amount of change in position between two consecutive frames. It describes the localized accuracy at fixed time intervals apart and is well-suited for assessing the drift of a visual odometer system.RPE is commonly used to assess the short-term performance of a system, such as drift per second. Its calculation formula is:
\begin{equation}
    E_i=(Q_i^{-1}Q_{i+\Delta})^{-1}(P_i^{-1}P_{i+\Delta})
\end{equation}
where ${Q}_{i}$ and $P_{i}$ are the true and estimated bit positions of the $i$th frame, respectively, and $\Delta$ is the time interval.

\subsection{DISTS Evaluation Indicators}
DISTS serves as a metric for assessing image quality, relying on the differentiable image saliency transform. The goal is to calculate image distances through feature mapping learning for assessing image quality. DISTS's fundamental concept involves converting the input image into a format based on pixel saliency, differentiable for direct training of neural networks to enhance image quality evaluation's scalability and portability. DISTS typically involves these steps:
\begin{enumerate}
    \item Significance Score Calculation: the significance score of each pixel in the image is first calculated, which can be achieved by a pre-trained deep learning model (e.g. VGG network).
    \item Pixel sorting: Pixels are sorted according to the saliency score to highlight the most important parts of the image.
    \item Constructing a new image representation: the sorted pixel order is used to construct a new image representation that can be used for subsequent image quality assessment.
    \item Distance calculation: An assessment score of the image quality is calculated by comparing the difference in the new representation between the original image and the processed image.
\end{enumerate}

The DISTS evaluation metrics provide a new perspective in the field of image quality assessment, especially in the application of deep learning and neural network modeling, which can provide more accurate and efficient evaluation results compared to traditional methods.

\subsection{Mean Angular Error (MAE°)}
Mean Angular Error (MAE°) is used to measure the average error between the predicted angle and the true angle. This metric is commonly used in the fields of computer vision and robot navigation, especially in tasks involving angle estimation, such as camera pose estimation, target tracking, motion capture, etc. Its calculation formula is:
\begin{equation}
    MAE°=\frac1n\sum_{i=1}^n|\theta_i-\hat{\theta}_i|
\end{equation}
Where $n$ is the number of observation points, $\theta_{i}$ is the true angle of the  $i$th observation point, and  $\hat{\theta}_i$ is the predicted angle of the  $i$th observation point.

A lower MAE° value results in a reduced discrepancy between the forecasted and actual angles, enhancing the model's efficacy. The utility of this metric lies in its ability to assess a model's precision in estimating angles, as it accurately mirrors the inaccuracies in predicting angles. Practically, MAE° aids scientists and engineers in comprehending how the model fares in angle estimation and in fine-tuning and fine-tuning the model as needed.

\subsection{Negative Log Likelihood (NLL)}
The NLL serves as a tool to assess the alignment between the model's output uncertainty and the actual observations. Within NeRF, the NLL method is commonly employed to assess the unpredictability of color values, with the formula displayed below:
\begin{equation}
    NLL=-\frac{1}{N}\sum_{N}^{i=1}\log\left(P(c_{i}|\mathbf{x}_{i})\right)
\end{equation}
where $N$ represents the sample count, $c_{i}$ represents the measured color value, and $P(c_i|x_i)$ the likelihood distribution of color $c_{i}$ at a specific point $x_{i}$. This likelihood distribution is derived from the color values forecasted by the model and the related uncertainty (for instance, by combining the model's weight and scale parameters.) A lower NLL value suggests a more accurate alignment of the model's forecasts with the actual data observed.

\subsection{S3IM Evaluation Indicators}
S3IM (Stochastic Structural SIMilarity) represents an innovative metric for evaluating Neural Fields' efficacy, particularly in NeRF and similar neural field techniques. Its goal is to gather non-local structural data by analyzing various data points collectively, instead of handling multiple inputs separately, to obtain non-local structural details. This method diverges from the conventional point-to-point loss functions like mean square error MSE, commonly employed in NeRF training.

S3IM's fundamental concept revolves around assessing the structural resemblance between two-pixel groups through a random assortment of pixels. The method encompasses the distant structural data from adjacent and remote pixels. S3IM is suggested to tackle the issue of current neural field techniques not fully utilizing the extensive structural details among pixels in training. Enhancing the model through the acquisition of non-local data results in better performance across various tasks. Implementing this metric offers fresh viewpoints and instruments for the study and utilization of neural field techniques.

\section{NERF-based Correlation Methods}\label{section:methods}
In recent years, NeRF's research has focused on improving rendering efficiency, optimizing few-view synthesis, enhancing rendering quality, and developing self-supervised learning and other key areas, aiming to address real-time rendering demands, improve the accuracy and generalization ability of 3D reconstruction, and reduce reliance on a large amount of annotated data. In this section, we will detail the latest developments up to March 2024. The relevant improvement methods are shown in Table \ref{methods_table}:

\begin{table*}[!h]
    \centering
    \caption{Qverview of NeRF-based Correlation Methods}
    \label{tab:methods_table}
    \begin{tabular}{c||l} \hline
    Methods                          & \multicolumn{1}{c}{Related Models}                                                         \\ \hline
    Rendering Quality Improvement    & UHDNeRF\cite{paper073}, BAD-NeRF\cite{paper074}, NeRFVS\cite{paper075}, RefSR-NeRF\cite{paper076}, NeRFLix\cite{paper077}, NeRFLiX++\cite{paper078}, etc.                            \\ \hline
    Less View Synthesis              & FlipNeRF\cite{paper079}, Single-Stage Diffusion NeRF\cite{paper080}, WaveNeRF\cite{paper081}, MixNeRF\cite{paper082}, FreeNeRF\cite{paper083}, etc                    \\ \hline
    Rendering Efficiency Improvement & Zip-NeRF\cite{paper084}, Re-ReND\cite{paper085}, Recursive-NeRF\cite{paper086}, F2-NeRF\cite{paper087}, NerfAcc\cite{paper088}, NeRFLight\cite{paper089}, etc.                       \\ \hline
    Improvement of Imaging Obstacles & ScatterNeRF\cite{paper090}, NeRFrac\cite{paper091}, NeRF-MS\cite{paper092}, DiffusioNeRF\cite{paper093}, ABLE-NeRF\cite{paper094}, Ref-NeRF\cite{paper095}, etc.                     \\ \hline
    Self-Supervised Learning         & SceneRF\cite{paper096}, MIMO-NeRF\cite{paper097}, CaFi-Net\cite{paper098}, CI-NeRF\cite{paper099}, etc.                                                \\ \hline
    Monocular New View Synthesis     & MonoNeRD\cite{paper100}, NeRDi\cite{paper101}, MonoNeRF\cite{paper102}, HOSNeRF\cite{paper103}, NeRF-DS\cite{paper104}, etc.                                          \\ \hline
    Posture Estimation               & IR-NeRF\cite{paper105}, DBARF\cite{paper106}, LU-NeRF\cite{paper107}, NoPe-NeRF\cite{paper108}, etc.                                                   \\ \hline
    Others                           & SparseNeRF\cite{paper109}(depth prior), NeRF-Det\cite{paper110}(end-to-end detection), IntrinsicNeRF\cite{paper111}(Unsupervised), etc.  \\
    \hline
    \end{tabular}
\end{table*}


\subsection{NeRF-based Rendering Quality Enhancement}
By employing deep learning techniques to create superior 3D scenes, NeRF has significantly advanced the synthesis of new views. Following the introduction of NeRF, scientists have been investigating methods to enhance its efficiency in view synthesis and understanding geometric properties. Tables \ref{tab:quality_table} and \ref{tab:quality_method_effects_table} below illustrate several principal areas of enhancement and their respective impacts:

\begin{table*}[!h]
    \centering
    \caption{Overview of Articles Related to Rendering Quality Improvement}
    \label{tab:quality_table}
    \begin{tabular}{c||c||l} \hline
    Category                                          & Model                         & \multicolumn{1}{c}{Highlight}                                                                                                                                                                                                                                                                                   \\ \hline
    \multirow{6}{*}{High-Resolution Rendering}     & UHDNeRF\cite{paper073}                       & \multicolumn{1}{m{11cm}}{Adaptive representation of scenes, both implicit and explicit, and strategies for separating frequencies in ultra-high-resolution imagery.}                                                                                                                                                                       \\  \cline{2-3}
                                                      & RefSR-NeRF\cite{paper076}                    & \multicolumn{1}{m{11cm}}{Innovative synthesis of views featuring superior quality and super-resolution, achieved by merging a low-resolution NeRF depiction with a high-resolution reference image.}                                                                                                                                       \\ \cline{2-3}
                                                      & E2NeRF\cite{paper145}                           & \multicolumn{1}{m{11cm}}{By merging information from an event camera and a conventional RGB camera, a precise NeRF is effectively derived from fuzzy images through the introduction of blur rendering loss and event rendering loss.}                                                                                                    \\ \hline
    \multirow{11}{*}{Grid-based Approach}            & Zip-NeRF\cite{paper084}                         & \multicolumn{1}{m{11cm}}{A combination of Mip-NeRF\cite{paper146} and mesh-based models is used to address antialiasing issues such as jaggedness and artifacts in NeRF.}                                                                                                                                                         \\ \cline{2-3}
                                                      & F2-NeRF\cite{paper087}                          & \multicolumn{1}{m{11cm}}{The creation of new perspectives is realized through the introduction of an innovative method for perspective deformation, adept at training superior neural radiation fields for extensive, limitless scenes with varied camera paths within minutes.}                                                       \\ \cline{2-3}
                                                      & NeRF2Mesh\cite{paper147}                        & \multicolumn{1}{m{11cm}}{Superior surface meshes featuring intricate textures are reassembled using multi-view RGB images, starting with the geometry and visual aspects from NeRF, followed by the extraction of rough mesh and progressive refinement of the surfaces to dynamically modify vertex locations and facial densities.} \\ \cline{2-3}
                                                      & DMAR\cite{paper148}                             & \multicolumn{1}{m{11cm}}{The suggestion is made for a dynamic mesh-aware radiation field to ensure uniform physical rendering and simulation of polygonal mesh assets in conventional graphics pipelines, enhancing visual realism through a bi-directional integration of   NeRF and surface depiction.}                                 \\ \hline
    \multirow{7}{*}{Others}                          & BAD-NeRF\cite{paper074}                         & \multicolumn{1}{m{11cm}}{The introduction of a new bundle-adjusted deblurring neural radiation field is suggested, designed to effectively handle intense motion blur and incorrect camera positioning.}                                                                                                                                  \\ \cline{2-3}
                                                      & StructNeRF\cite{paper113}                       & \multicolumn{1}{m{11cm}}{Enhance the accuracy of geometry estimation and view synthesis by utilizing uniformity across multiple views and planar limitations, bypassing the need for pre-training external data.}                                                                                                                         \\ \cline{2-3}
                                                      & NeRFLiX++\cite{paper078}                        & \multicolumn{1}{m{11cm}}{A hybrid recursive inter-viewpoint fusion method called IVM is proposed to enhance NeRF-rendered views by fusing useful information from the selected most relevant reference viewpoints.}                                                                                                                       \\ \cline{2-3}
                                                      & NeRC\cite{paper149}                             & \multicolumn{1}{m{11cm}}{A novel learning method based on implicit neural representation is proposed for efficiently rendering diverse focalization effects in scenes that vary with camera and light source movements.}   \\ \hline                                                                                                        
    \end{tabular}
\end{table*}
    

\begin{table}[!h]
    \centering
    \caption{Rendering Quality Improvement Related Method Effects}
    \label{tab:quality_method_effects_table}
    \begin{tabular}{c||cccc} \hline
    Dataset                                                           & Model                       & PSNR$\uparrow$  & SSIM$\uparrow$  & LPIPS$\downarrow$ \\ \hline
                                                                      & Zip-NeRF\cite{paper084}                    & 32.52  & 0.954  & 0.037  \\
                                                                      & AligNeRF\cite{paper215}                    & 24.55  & 0.703  & 0.263  \\
    \multirow{-3}{*}{\thead{Mip-NeRF 360 \\ Dataset\cite{paper036}}}             & CamP\cite{paper217} & 28.86  & 0.843  & 0.182  \\ \hline
                                                                      & E2NeRF\cite{paper145}                      & 29.77  & 0.96   & 0.0725 \\
                                                                      & Exact-NeRF\cite{paper216}                  & 34.707 & 0.9705 & 0.0667 \\
    \multirow{-3}{*}{NSD\cite{paper023}}           & CamP\cite{paper217}                        & 28.29  & 0.93   & 0.064  \\ \hline
                                                                      & NeRFLix\cite{paper077}                     & 27.26  & 0.863  & 0.159  \\
                                                                      & NeRFLiX++\cite{paper078}                   & 27.25  & 0.858  & 0.17   \\
                                                                      & UHDNeRF\cite{paper073}                     & 29.03  & 0.834  & 0.325  \\
                                                                      & RefSR-NeRF\cite{paper076}                  & 26.23  & 0.874  & 0.243  \\
    \multirow{-5}{*}{LLFF\cite{paper033}}                             & ContraNeRF\cite{paper218}                  & 25.44  & 25.44  & 0.178  \\ \hline
    \end{tabular}
    \end{table}
    

\subsubsection{High-Resolution Rendering}
{\bf{UHDNeRF}} \cite{paper073} (2023) introduces an adaptive implicit-explicit scene representation that improves performance when modeling details by combining implicit volume rendering and explicit sparse point clouds. Specifically, UHDNeRF uses a frequency separation strategy where the implicit volume is responsible for learning low-frequency attributes of the whole scene, while the sparse point cloud is used to reproduce high-frequency details. In addition, the article proposes an extraction method for global structural features and local point features and introduces a patch-based sampling strategy to reduce the computational cost, and the model diagram is shown in Fig. \ref{UHDNeRF} below.
\begin{figure}[h!]
    \centering
    \includegraphics[width=0.45\textwidth]{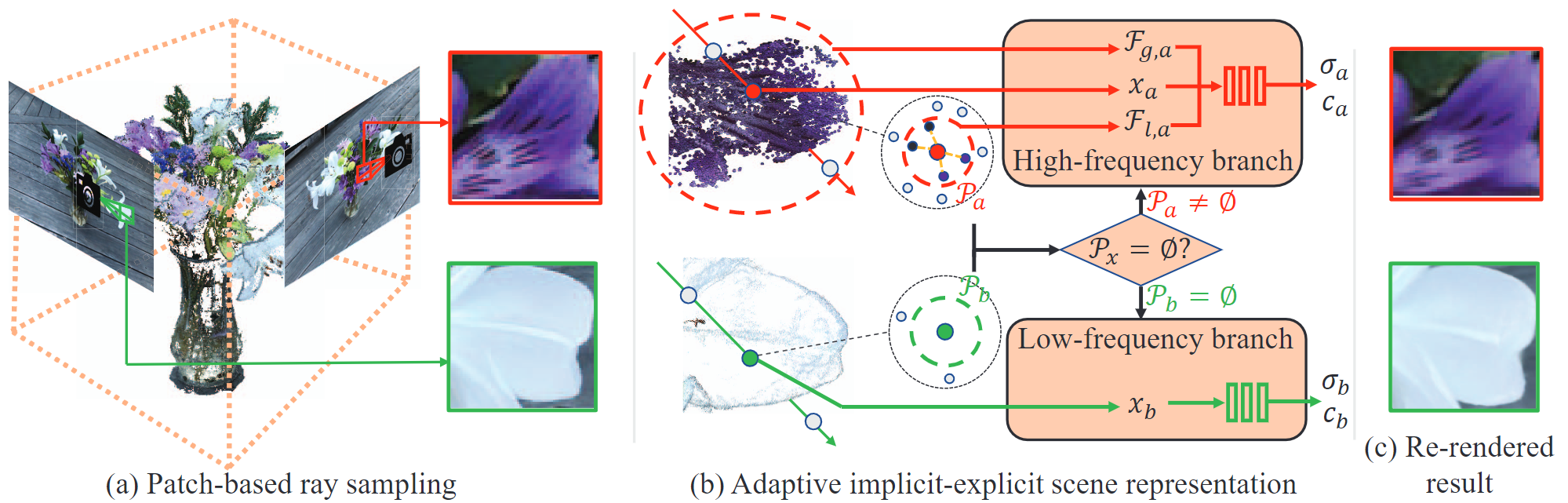}
    \caption{Overview of UHDNeRF.}
    \label{UHDNeRF}
\end{figure}

{\bf{RefSR-NeRF}} \cite{paper076} (2023) has developed a streamlined RefSR model to adapt the Inverse Degradation Process (IDP) from NeRF rendering to the desired high-resolution image, aiming to rectify the unsatisfactory distortions caused by pre-trained models. Regarding assessment outcomes, RefSR-NeRF exhibits a remarkable balance among the quality of rendering, speed, and memory consumption in numerous benchmark evaluations. In contrast to NeRF and its derivatives, RefSR-NeRF attains a speed increase of 52 times and a minor rise in memory consumption, all the while preserving or surpassing the efficacy of current techniques.

{\bf{E2NeRF}} \cite{paper145} (2023) is used to learn sharp neural 3D representations from blurred images and event data. The innovation of the framework is the combination of data from a bio-inspired event camera and a standard RGB camera. To efficiently introduce event streams into the learning process of neural volumetric representations, the authors propose blur rendering loss and event rendering loss, which guides the network by modeling the actual blurring process and event generation process. In addition, to deal with real-world data, the authors constructed an event-stream-based camera pose estimation framework to generalize the method to real-world applications. In terms of evaluation results, extensive experiments of E2NeRF on synthetic and real-world data show that it can effectively learn sharp NeRF from blurred images, especially in complex and low-light scenes.

\subsubsection{Grid-based Approach}
{\bf{Zip-NeRF}} \cite{paper084} (2023) is a NeRF model that combines the scale-awareness and anti-aliasing properties of Mip-NeRF 360 \cite{paper036} with the fast meshing training of Instant-NGP \cite{paper150}. The spatial anti-aliasing problem is solved by utilizing multi-sampling and pre-filtering techniques, and a new loss function is introduced to solve the "z-aliasing" (aliasing along a ray) problem. This approach not only improves the rendering quality but also significantly accelerates the training speed, which is 24 times faster than Mip-NeRF 360. In terms of evaluation results, Zip-NeRF outperforms all previous techniques on the Mip-NeRF 360 benchmarks, with error rate reductions ranging from 0.08 to 0.77.

\begin{figure}[h!]
    \centering
    \includegraphics[width=0.45\textwidth]{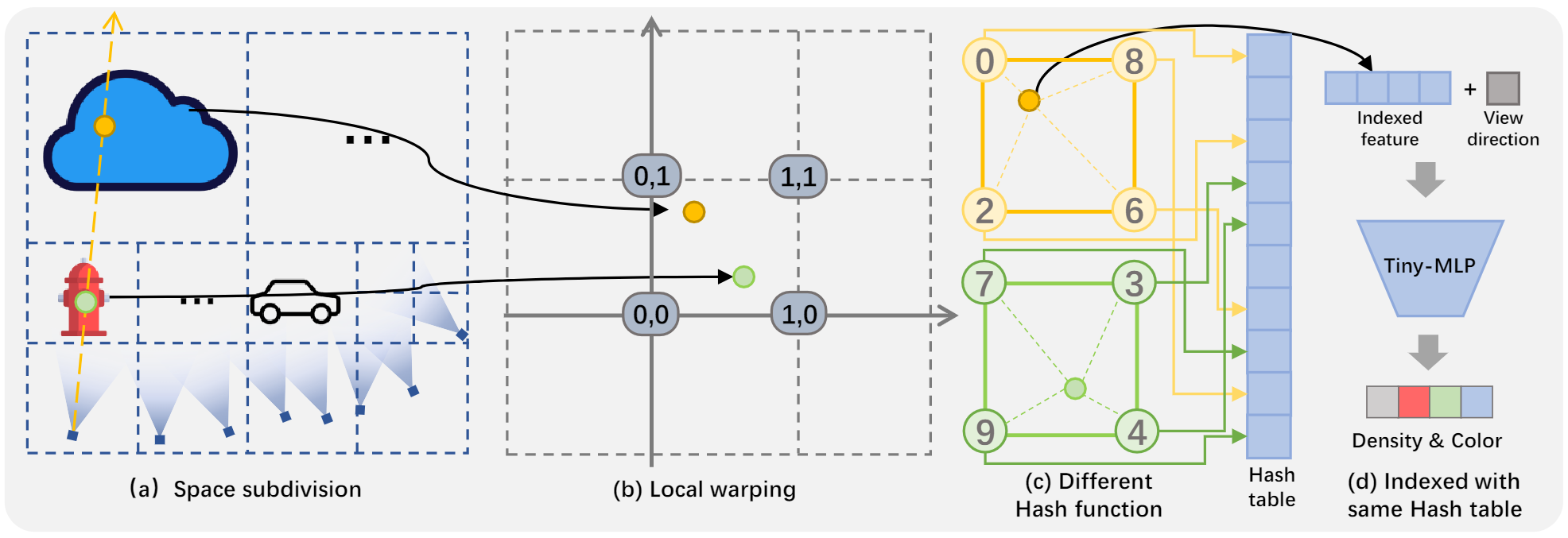}
    \caption{Pipeline of F2-NeRF.}
    \label{F2_NeRF}
\end{figure}

{\bf{F2-NeRF}} \cite{paper087} (2023) introduces an innovative grid-oriented NeRF framework for the creation of fresh perspectives. Contrasting with current rapid grid-oriented NeRF training systems (such as Instant-NGP \cite{paper150}, Plenoxels \cite{paper151}, DVGO \cite{paper152}, or TensoRF \cite{paper153}), which primarily handle limited scenarios and depend on spatial distortion for limitless situations, F2-NeRF introduces an innovative technique known as perspective warping   (PW), capable of managing various trajectories through depth analysis of spatial distortion mechanisms. a new technique for spatial distortion known as perspective warping, adept at managing various trajectories, as depicted in Fig. \ref{F2_NeRF}. Regarding assessment outcomes, F2-NeRF demonstrates the capability to produce superior images through identical perspective warping methods, and with the introduction of the new free trajectory dataset, it surpasses the standard gridded NeRF technique in image clarity, requiring merely around 12 minutes for training.

{\bf{NeRF2Mesh}} \cite{paper147} (2023) is employed for the reconstruction of surface meshes, incorporating intricate textures from various RGB images. Initially, the system effectively sets up the geometry and perspective-separated visual using NeRF, subsequently removing the coarse mesh and dynamically modifying vertex locations and facial densities through a repetitive surface enhancement algorithm reliant on reprojection rendering mistakes. Furthermore, this technique simultaneously enhances both the shape and visual aspect, transforming these elements into texture images for immediate rendering. This technique can recreate superior surface meshes using a reduced number of vertices and faces, and these meshes work well with standard 3D hardware and software for instantaneous rendering and interactive editing. Despite its constraints in handling uncertain lighting and intricate effects based on perspective, this technique proves to be a viable option for deriving superior meshes from volumetric NeRF reconstructions.

\subsubsection{Others}
{\bf{BAD-NeRF}} \cite{paper074} (2023) is employed to derive precise 3D scene shapes and visuals from a collection of fuzzy images. Through simultaneous learning of the 3D representation and fine-tuning of the camera's position, one can effectively clear the blur, produce fresh perspective images, and precisely reconstruct the camera's movement path during the exposure period. During the assessment, BAD-NeRF surpasses earlier studies in both artificial and actual datasets, particularly in cases of intense motion blur. Experimental findings indicate that BAD-NeRF enhances image clarity by accurately simulating the actual imaging of motion-blurred images. Furthermore, this technique demonstrates remarkable precision in predicting the camera's path.

\begin{figure}[h!]
    \centering
    \includegraphics[width=0.45\textwidth]{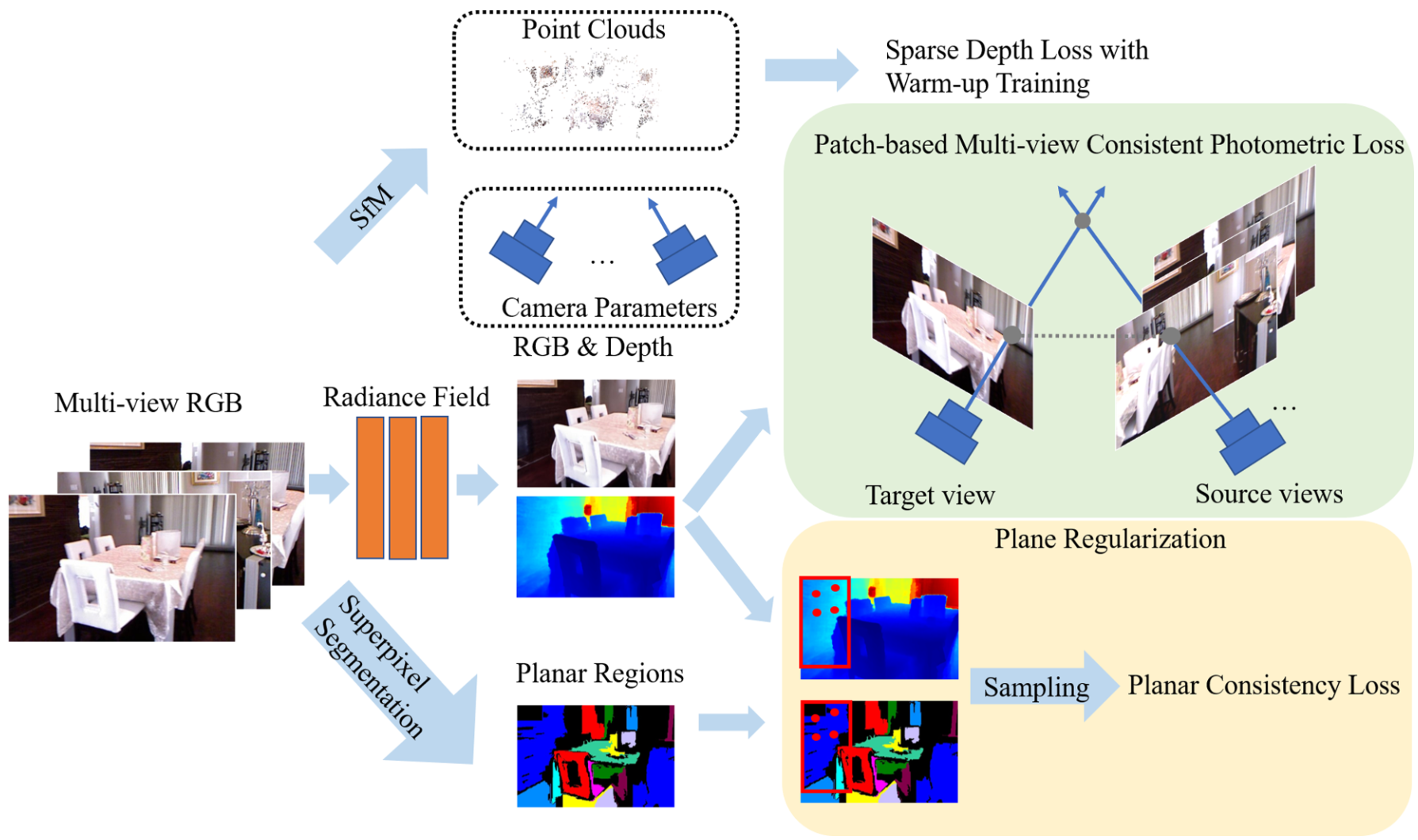}
    \caption{Overview of StructNeRF.}
    \label{StructNeRF}
\end{figure}

{\bf{StructNeRF}} \cite{paper113} (2022) introduces an innovative NeRF technique for indoor settings, enhancing the creation of new views in limited inputs through the use of structural indicators in multi-view inputs. Additionally, it integrates structural indicators into the self-guided depth estimation technique by limiting the shape of textured areas via patch-based multi-view coherent photometric loss and applying planar coherence loss to areas without texture to maintain three-dimensional coherent planes. Next, an innovative training approach, namely progressive training, has been implemented to reduce noise from the limited depth point cloud in structure-from-motion (SfM), as depicted in the model's rendering process in Fig. \ref{StructNeRF}.

{\bf{NeRFLiX++}} \cite{paper078} (2023) proposed NeRFLiX and NeRFLiX++ for improving the quality of novel view synthesis. Netflix learns degradation-driven multiview blending by designing a NeRF-style degradation simulator (NDS) and a multiview mixer (IVM) to efficiently remove native rendering in NeRF rendering artifacts.NeRFLiX++ further introduces a two-stage degradation simulation strategy and a more efficient bootstrapped multi-view mixer to achieve better performance and significantly improved computational efficiency.

\subsection{NeRF-based Less View synthesis}
In the field of computer vision and graphics, Few-Shot View Synthesis (FVS) is a challenging task to generate new, realistic 3D scene views from a limited number of input views. The significance of this task extends to the realms of virtual reality (VR), augmented reality (AR), and the creation of special effects in movies. In recent years, with the rise of NeRF technology, the research on less view synthesis has made significant progress. The following are some key improvement directions and evaluation results, as shown in Tables \ref{tab:less_view_synthesis_table}, and \ref{tab:less_view_synthesis_evaluation_table}:

\begin{table*}[!h]
    \centering
    \caption{Overview of Articles Related to Less View Synthesis}
    \label{tab:less_view_synthesis_table}
    \begin{tabular}{c||c||l} \hline
    Category                                & Model                          & \multicolumn{1}{c}{Highlight}                                                                                                                                                                                                                           \\ \hline
    \multirow{3}{*}{Single View Synthesis} & NeuralLift-360\cite{paper154}                 & \multicolumn{1}{m{11cm}}{Converting a single 2D photo taken in the field into a 3D object with a 360° view improves the quality and efficiency of 3D content generation by combining depth-aware Neural Radiation Field (NeRF) and denoising diffusion models.}                   \\ \cline{2-3}
                                            & SHERF\cite{paper123}                          & \multicolumn{1}{m{11cm}}{The first generalizable human NeRF model for recovering animatable 3D humans from a single input image.}                                                                                                                                                 \\ \hline
    \multirow{8}{*}{Sparse View Synthesis} & MixNeRF\cite{paper082} & \multicolumn{1}{m{11cm}}{To create novel perspectives from limited data, simulate light beams using a mixed-density approach, enhance the precision of learning 3D scene geometry, and show exceptional training and inference skills in various standard benchmark evaluations.} \\ \cline{2-3}
                                            & VM-NeRF\cite{paper155}                        & \multicolumn{1}{m{11cm}}{An innovative approach that integrates view morphing methods for the sparse view issue in NeRF markedly enhances the synthesis quality of new views in limited views through the creation of geometrically uniform image alterations in training.}       \\ \cline{2-3}
                                            & FreeNeRF\cite{paper083}                       & \multicolumn{1}{m{11cm}}{Improving Sample-Few Neural Rendering (NeRF) by Frequency Regularization and Masking Regularization.}                                                                                                                                                    \\ \cline{2-3}
                                            & ViP-NeRF\cite{paper156}                       & \multicolumn{1}{m{11cm}}{An approach to enhance NeRF efficiency in scenarios of limited input by standardizing the training procedure with visibility prior derived from Planar Scanning Volume(PSV).}     \\ \hline                                                                     
    \end{tabular}
    \end{table*}
    

\begin{table}[!h]
    \centering
    \caption{Evaluation of Methods Related to Less View Synthesis}
    \label{tab:less_view_synthesis_evaluation_table}
    \begin{tabular}{c||c||cccc} \hline
    Dataset                              & View                    & Model      & PSNR$\uparrow$  & SSIM$\uparrow$  & LPIPS$\downarrow$ \\ \hline
    \multirow{10}{*}{DTU\cite{paper040}}                & \multirow{4}{*}{3-View} & FlipNeRF   & 19.55 & 0.767  & 0.18   \\
                                         &                         & SparseNeRF & 19.55 & 0.769  & 0.201  \\
                                         &                         & MixNeRF    & 19.95 & 0.744  & 0.203  \\
                                         &                         & FreeNeRF   & 19.92 & 0.787  &        \\ \cline{2-6}
                                         & \multirow{3}{*}{6-View} & FlipNeRF   & 22.45 & 0.839  & 0.098  \\
                                         &                         & MixNeRF    & 22.3  & 0.835  & 0.102  \\
                                         &                         & FreeNeRF   & 23.25 & 0.844  &        \\ \cline{2-6}
                                         & \multirow{3}{*}{9-View} & FlipNeRF   & 25.12 & 0.882  & 0.062  \\
                                         &                         & MixNeRF    & 25.03 & 0.879  & 0.065  \\
                                         &                         & FreeNeRF   & 25.38 & 0.888  &        \\ \cline{1-6}
    \multirow{8}{*}{LLFF\cite{paper033}}                & \multirow{4}{*}{3-View} & FlipNeRF   & 19.34 & 0.631  & 0.2325 \\
                                         &                         & SparseNeRF & 19.86 & 0.624  & 0.328  \\
                                         &                         & MixNeRF    & 19.27 & 0.629  & 0.236  \\
                                         &                         & FreeNeRF   & 19.63 & 0.612  & 0.308  \\ \cline{2-6}
                                         & \multirow{2}{*}{6-View} & MixNeRF    & 23.76 & 0.791  & 0.115  \\
                                         &                         & FreeNeRF   & 23.73 & 0.779  & 0.195  \\ \cline{2-6}
                                         & \multirow{2}{*}{9-View} & MixNeRF    & 25.2  & 0.833  & 0.087  \\
                                         &                         & FreeNeRF   & 25.13 & 0.827  & 0.16   \\ \cline{1-6}
    \multirow{4}{*}{NSD\cite{paper023}} & \multirow{2}{*}{4-View} & FlipNeRF   & 20.6  & 0.882  & 0.159  \\
                                         &                         & VM-NeRF    & 16.9  & 0.7563 & 0.2461 \\ \cline{2-6}
                                         & \multirow{2}{*}{8-View} & FlipNeRF   & 24.38 & 0.883  & 0.095  \\
                                         &                         & VM-NeRF    & 24.39 & 0.8768 & 0.1146 \\ \hline
    \end{tabular}
    \end{table}
    

\subsubsection{Single View Synthesis}
{\bf{NeuralLift-360}} \cite{paper154} (2023) can transform a solitary 2D image from a natural setting into a three-dimensional object offering a complete 360° perspective. It merges depth-aware NeRF with noise-free diffusion modeling and can estimate rough depths when rough depth estimation is present, by implementing a sampling approach involving ranking loss and guided by CLIP for efficient navigation. Furthermore, the system adjusts to individual images by refining the diffusion model, yet retains the capacity to produce varied content. Nonetheless, NeuralLift-360 continues to face constraints regarding the resolution of targets and the management of intricate scenarios, such as obscured multiple objects.

\begin{figure}[h!]
    \centering
    \includegraphics[width=0.45\textwidth]{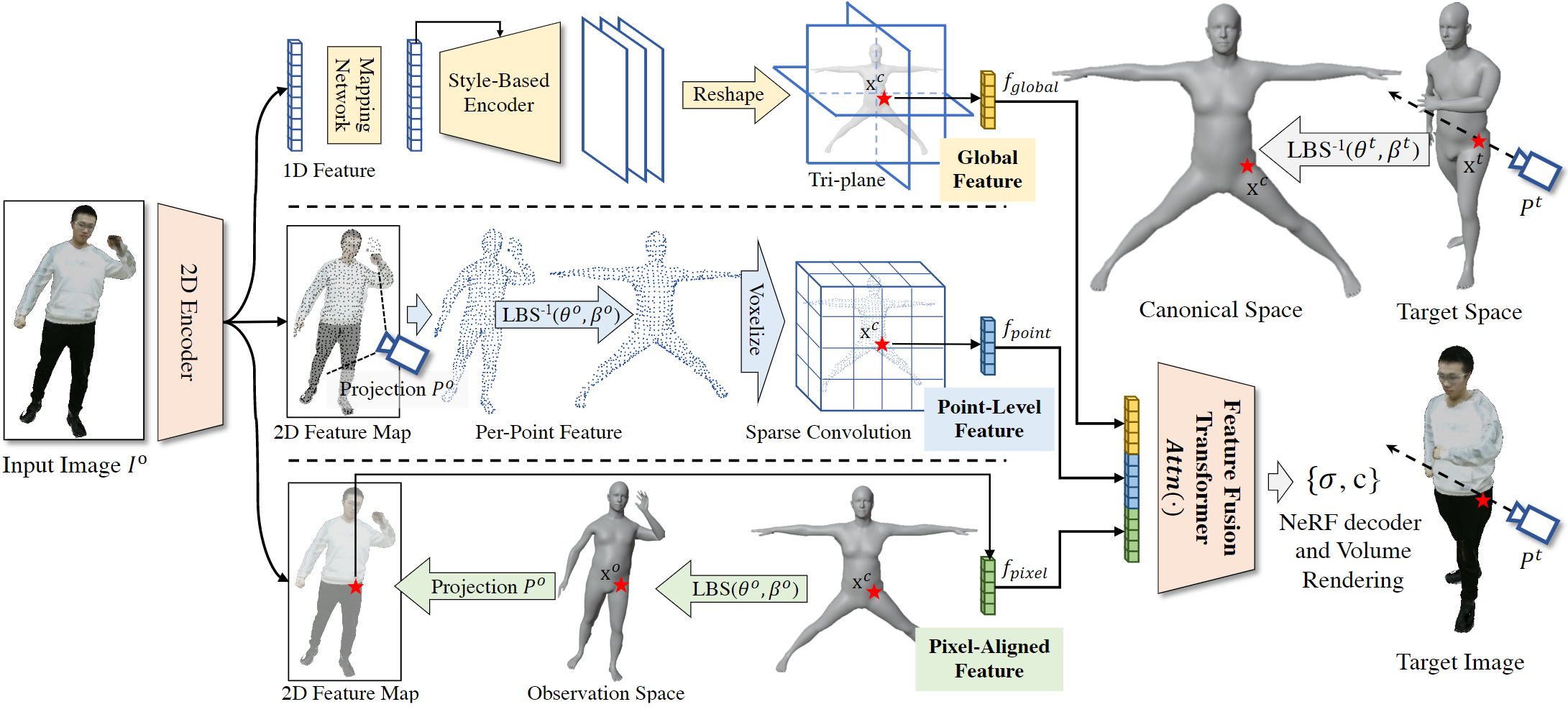}
    \caption{Overview of SHERF Framework.}
    \label{SHERF}
\end{figure}

{\bf{SHERF}} \cite{paper123} (2023)  introduces an innovative method for forecasting ongoing, varied depth assessments from each perspective, addressing the intrinsic uncertainty in single-eye depth estimation via conditional implicit maximum likelihood estimation(cIMLE). Furthermore, a novel spatial sculpting loss technique is presented, capable of merging various theoretical depth maps across different perspectives and distilling them into a unified geometry that aligns with all views, thereby enhancing the synthesis of new perspectives in finite views, as depicted in the model's framework in Fig. \ref{SHERF}.

\subsubsection{Sparse View Synthesis}
\begin{figure}[h!]
    \centering
    \includegraphics[width=0.45\textwidth]{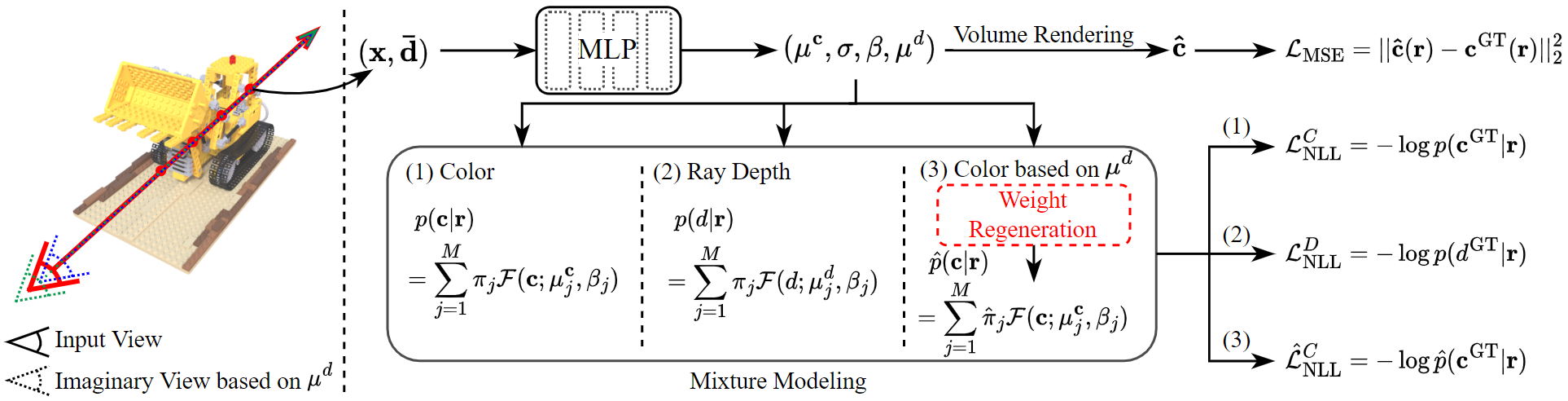}
    \caption{Overview of MixNeRF.}
    \label{MixNeRF}
\end{figure}

{\bf{MixNeRF}} \cite{paper082} (2023)  serves the purpose of creating fresh perspectives from infrequent inputs. This system replicates light using a mixing density model to effectively understand the combined RGB color distribution in light samples. MixNeRF introduces an additional task, estimating light depth, linked to the 3D scene's geometric height, with the model's structure depicted in Fig. \ref{MixNeRF}. Furthermore, MixNeRF enhances the blending weights using estimated ray depths, bolstering the sturdiness of colors and perspectives, and surpasses other advanced techniques in training and inference effectiveness across numerous standard benchmark evaluations.

{\bf{VM-NeRF}}\cite{paper155} (2023) addresses the issue of NeRF overfitting when dealing with limited viewpoints. Through a View Morphing method, VM-NeRF produces geometrically uniform image transitions, eliminating the need for a pre-existing understanding of the scene's structure due to its foundation in projective geometry basics. This technique seamlessly incorporates the process of generating geometric views into the conventional NeRF training, markedly enhancing the synthesis of new views when limited viewpoints are present. Trials reveal that VM-NeRF enhances PSNR by 1.8 dB (with eight perspectives) and 1.0 dB (with four) when compared to current techniques for handling sparse viewpoints in NeRF models.

{\bf{FreeNeRF}\cite{paper083}} (2023) significantly improves the rendering performance of NeRF with fewer samples through frequency regularization. The method achieves comparable performance to complex methods with minimal modifications by introducing two regularization terms: frequency range regularization and near-camera density field penalties. The key to FreeNeRF is that it does not require additional pre-training or supervised signals and does not add additional computational cost.FreeNeRF has been tested on several datasets, including Blender \cite{paper023}, DTU \cite{paper040}, and LLFF \cite{paper033}, and achieves state-of-the-art performance that outperforms existing methods both in terms of image quality and geometric accuracy. Experimental results show that FreeNeRF can generate high-quality new-view images even with only a small number of input views, and the training efficiency is comparable to that of the original NeRF.

\subsection{NeRF-based Rendering Efficiency Improvement}
NeRF's rendering procedure typically demands substantial computational power, constraining its applicability in real-time scenarios. Researchers have suggested multiple techniques to improve NeRF's rendering efficiency as a solution to this issue. As an illustration, enhancing network architecture, refining sampling methods, and optimizing computational resource utilization are among the strategies, with crucial directions for improvement detailed and the results of the assessment, as shown in Tables \ref{tab:rendering_efficiency_table} and \ref{tab:rendering_efficiency_evaluation_table}:

\begin{table*}[!h]
    \centering
    \caption{ Overview of Articles Related to Rendering Efficiency Improvements}
    \label{tab:rendering_efficiency_table}
    \begin{tabular}{c||c||l} \hline
    Category                                                             & Model          & \multicolumn{1}{c}{Highlight}                                                                                                                                                                                                     \\ \hline
    \multirow{5}{*}{Model Simplification}                               & HollowNeRF\cite{paper157}     & \multicolumn{1}{m{11cm}}{The effective trimming of features is achieved through the training of a rough 3D saliency mask and the application of trainable collision mitigation, which automatically narrows the feature mesh during training, thereby greatly decreasing the parameter count while preserving the quality of rendering.} \\ \cline{2-3}
                                                                         & MIMO-NeRF\cite{paper097}      & \multicolumn{1}{m{11cm}}{A Multiple-Input Multiple-Output NeRF (MIMO-NeRF) is proposed, which reduces the number of operational MLPs by replacing SISO MLPs with MIMO MLPs and mapping them in groups.}                                                                                                                                  \\ \cline{2-3}
                                                                         & Pyramid NeRF\cite{paper158}   & \multicolumn{1}{m{11cm}}{By using image pyramids and frequency-guided optimization strategies, training and inference speeds are significantly improved while maintaining high-quality renderings.}                                                                                                                                      \\ \hline
     \multirow{8}{*}{{\thead{Rendering Speed \\ Improvement}}}          & NerfAcc\cite{paper088}        & \multicolumn{1}{m{11cm}}{An efficient sampling strategy is provided by introducing the concept of a unified transmittance estimator that unifies different types of sampling methods.}                                                                                                                                                   \\ \cline{2-3}
                                                                         & F2-NeRF\cite{paper087}        & \multicolumn{1}{m{11cm}}{Fast neural radiation field training method with building a neural representation of an unbounded scene and realizing the new view synthesis task by volumetric rendering.}                                                                                                                                     \\ \cline{2-3}
                                                                         & SteerNeRF\cite{paper159}      & \multicolumn{1}{m{11cm}}{A low-resolution feature map is rendered by a volume rendering method, and a 2D neural renderer is used to generate an output image at the target resolution by combining the features from the previous frames.}                                                                                               \\ \cline{2-3}
                                                                         & Recursive-NeRF\cite{paper086} & \multicolumn{1}{m{11cm}}{The model divides complex scenes into different sub-networks by predicting uncertainty to learn implicit geometric and appearance representations more efficiently.}                                                                                                                                            \\ \hline
    \multirow{8}{*}{\thead{Real Time \\ Rendering}}                      & LNeDRF\cite{paper160}            & \multicolumn{1}{m{11cm}}{Efficient 3D scene representation is achieved by reducing the number of sampling points on each ray while preserving the realistic details of complex scenes.}                                                                                                                                                  \\ \cline{2-3}
                                                                         & MobileNeRF\cite{paper161}     & \multicolumn{1}{m{11cm}}{Efficiently compositing new perspective images on mobile devices using polygonal textures and standard rendering pipelines achieves a 10x rendering speedup compared to existing techniques.}                                                                                                                   \\ \cline{2-3}
                                                                         & VI-NeRF-SLAM\cite{paper162}   & \multicolumn{1}{m{11cm}}{A real-time visual-inertial SLAM system that combines a NeRF map representation with a conventional SLAM optimization process for real-time dense reconstruction.}                                                                                                                                              \\ \cline{2-3}
                                                                         & NeX360\cite{paper163}         & \multicolumn{1}{m{11cm}}{A real-time panorama synthesis method that uses neural basis function extensions to construct viewpoint-dependent pixel representations that are rendered using an implicit-explicit modeling strategy.}        \\ \hline                                                                                                
    \end{tabular}
    \end{table*}
    

\begin{table*}[!h]
    \centering
    \caption{Evaluation of Methods Related to Rendering Efficiency Improvement}
    \label{tab:rendering_efficiency_evaluation_table}
    \begin{tabular}{c||ccccccc} \hline
    Dataset                                                                 & Model                    &Test GPU    & PSNR$\uparrow$  & SSIM$\uparrow$  & LPIPS$\downarrow$ & Render Time(s) & FPS   \\ \hline
                                                                            & Zip-NeRF\cite{paper084}                 &1xRTX3090   & 32.52     & 0.954 & 0.037  & 0.89           &         \\ \cline{2-8}
                                                                            & NeRF2Mesh\cite{paper147}                &1xV100      & 22.33     & 0.538 & 0.481  &                &         \\ \cline{2-8}
                                                                            & F2-NeRF\cite{paper087}                  &1x2080Ti    & 26.39     & 0.746 & 0.361  &                &         \\ \hline
    \multirow{-5}{*}{\thead{mip-NeRF 360 \\Dataset\cite{paper036}}}         & MobileNeRF\cite{paper161}               &1xRTX2070   & 21.95     & 0.47  & 0.47   &                & 279.7   \\ \cline{2-8}
                                                                            & Pyramid NeRF\cite{paper158}             &1x2080Ti    & 30.72     & 0.948 & 0.039  & 0.033          &         \\ \cline{2-8}
                                                                            & CG-NeRF\cite{paper164}                  &1xV100      & 34.12     & 0.98  & 0.04   & 0.2            &         \\ \cline{2-8}
    \multirow{-4}{*}{NSD\cite{paper023}}                                    & MIMO-NeRF\cite{paper097}                &1x3080Ti    & 31.26     & 0.953 & 0.054  & 5.15           &         \\  \hline
                                                                            & MIMO-NeRF\cite{paper097}                &1x3080Ti    & 27.7      & 0.87  & 0.155  & 4.55           &         \\ \cline{2-8}
                                                                            & F2-NeRF\cite{paper087}                  &1x2080Ti    & 26.54     & 0.844 & 0.189  &                &         \\ \cline{2-8}
    \multirow{-3}{*}{LLFF\cite{paper033}}                                   & MobileNeRF\cite{paper161}               &1xRTX2070   & 25.91     & 0.825 & 0.183  &                & 349.34  \\ \hline
                                                                            & DOT\cite{paper165}                      &1xA100      & 32.113    & 0.959 & 0.053  &                & 531.144 \\ \cline{2-8}
                                                                            & SteerNeRF\cite{paper159}                &1xA100      & 28.65     & 0.924 & 0.121  &                & 27.24   \\ \cline{2-8}
                                                                            & LNeDRF\cite{paper160}                   &1xA100      & 28.12     & 0.915 & 0.089  &                & 27.49   \\ \cline{2-8}
    \multirow{-4}{*}{Tanks \& Temples\cite{paper039}}                       & NeRFLight\cite{paper089}                &1xA100      & 27.85     & 0.939 & 0.084  &                & 78      \\ \hline
    \end{tabular}
\end{table*}


\subsubsection{Model Simplification}
\begin{figure}[h!]
    \centering
    \includegraphics[width=0.45\textwidth]{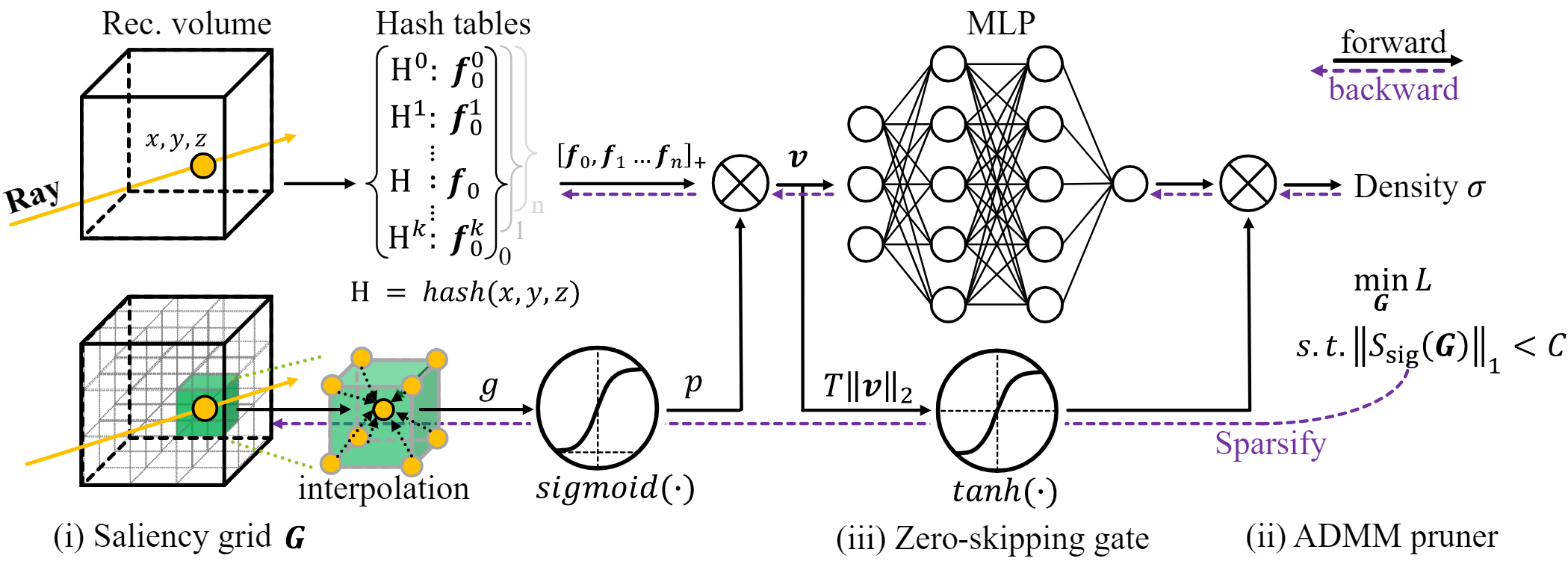}
    \caption{Overview of the HollowNeRF workflow.}
    \label{HollowNeRF}
\end{figure}

{\bf{HollowNeRF}} \cite{paper157} (2023) introduces an innovative compression method for NeRF using a hash grid. The process of efficient feature pruning involves educating a rough 3D saliency mask and reducing its size during the training phase through an alternating direction multiplier method (ADMM) pruner. The HollowNeRF technique leverages the 3D scene's sparsity to reallocate hash conflicts, enhancing the quality of rendering with fewer parameters than current advanced solutions, leading to an improved balance between cost and accuracy, as depicted in the model architecture diagram in Fig. \ref{HollowNeRF}. Nonetheless, HollowNeRF encounters difficulties in handling objects that have reflective surfaces and are rendered inadequately.

{\bf{MIMO-NeRF}} \cite{paper097} (2023) minimizes MLP operations in rendering by employing a multi-input, multiple-output (MIMO) multilayer perceptron (MLP) to enhance the speed of rendering. Furthermore, the paper introduces a self-directed learning approach for standardizing the MIMO MLP using several rapidly adaptable MLPs, addressing the issue of saliency ambiguity arising from the choice of input coordinate sets, independent of pre-trained models. Studies indicate that MIMO-NeRF outperforms the original NeRF in image quality metrics like PSNR, SSIM, and LPIPS, often with a notably reduced inference duration.

{\bf{Pyramid NeRF}} \cite{paper158} (2023) accomplishes rapid NeRF training and inference through the application of an image pyramid and a strategy based on frequency optimization. This technique utilizes a  "low-frequency-first, high-frequency-second" approach, substituting large MLPs with thousands of smaller ones via progressive segmentation, markedly enhancing both training and inference efficiency. Furthermore, the introduction of a sampling approach that is aware of depth aims to enhance the quality of rendering and hasten the convergence of training. The experimental findings reveal that Pyramid NeRF enhances training and inference speeds by 15x and 805x, respectively, and notably betters structural and perceptual resemblance relative to the original NeRF.

\subsubsection{Rendering Speed Improvement}
{\bf{NerfAcc}} \cite{paper088} (2023) serves to amalgamate sophisticated sampling techniques to hasten NeRF training. It harmonizes various sampling approaches via a cohesive transmittance estimator concept, simplifying the model's migration and integration. Furthermore, NerfAcc accommodates a variety of formats, such as explicit voxels, MLP, or mixed representations, and can manage scenes without limits. Research indicates that NerfAcc markedly shortens the training duration for various NeRF correlation techniques, achieving speeds between 1.5x and 20x, and typically enhances image clarity.

{\bf{SteerNeRF}} \cite{paper159} (2022) leverages the reality that shifts in perspective during interactive viewpoint management are typically seamless and unbroken. This technique cuts down on rendering duration and memory usage by initially creating low-resolution feature maps, followed by the use of a streamlined 2D neural renderer to produce output images at the desired resolution. Research indicates that SteerNeRF markedly enhances the speed of rendering while preserving image clarity comparable to current techniques, bridging the speed disparity between low-memory and cache-based approaches.

\begin{figure}[h!]
    \centering
    \includegraphics[width=0.45\textwidth]{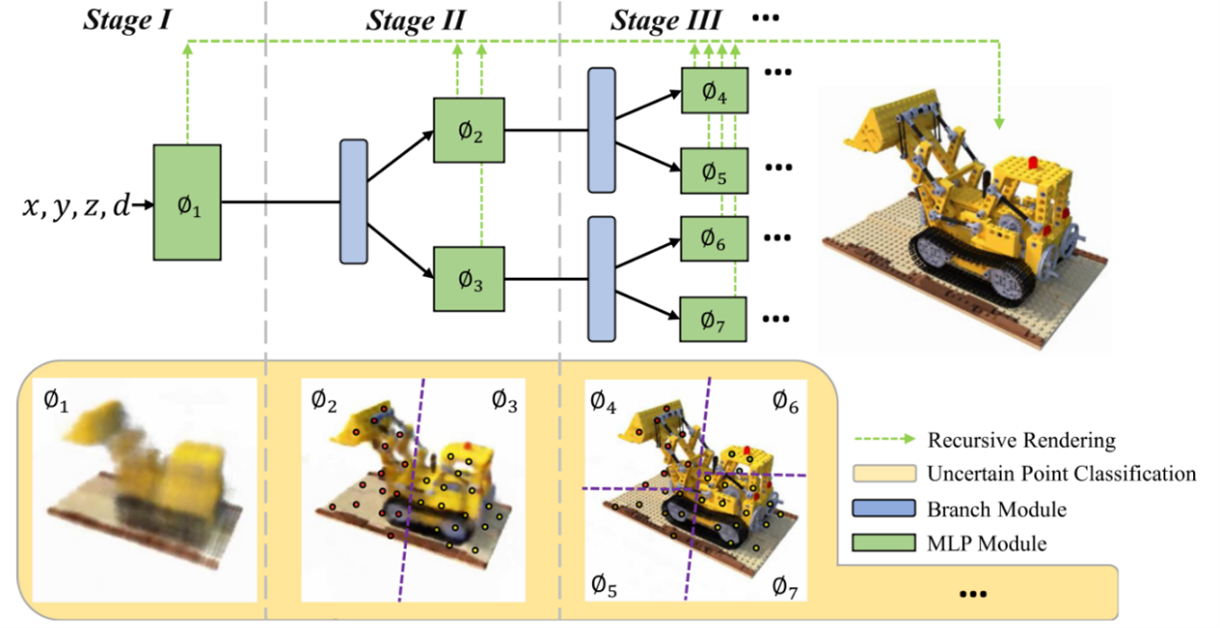}
    \caption{Pipeline of Recursive-NeRF.}
    \label{Recursive_NeRF}
\end{figure}

{\bf{Recursive-NeRF}} \cite{paper086} (2022) adjusts to scene intricacies by continuously implementing the NeRF framework and expanding the network dynamically. It forecasts uncertainty at every phase and determines the transfer of query coordinates to a more advanced neural network, depending on the uncertainty degree, as depicted in Fig. \ref{Recursive_NeRF}. The experimental findings indicate that Recursive-NeRF surpasses the conventional NeRF approach in performance across various public datasets and extensive scene datasets. Furthermore, Recursive-NeRF possesses the capability to dynamically modify the network's architecture across various image resolutions, catering to tasks of diverse complexity.

\subsubsection{Real Time Rendering}
{\bf{LNeDRF}} \cite{paper160} (2023) adeptly surpasses the natural imprecision in 3D surface reconstruction by employing a dual-layer duplex mesh depiction of the scene, attaining a superior visual quality via screen-space convolution as opposed to MLPs. Furthermore, a novel optimization approach for multi-view distillation is suggested, markedly enhancing the overall framework's efficiency. The runtime efficiency of the NeRF model has been enhanced by a factor of 10,000, thanks to comprehensive testing on standard datasets. Analyzing the Synthetic \cite{paper023} and Tanks \& Temples \cite{paper039} datasets quantitatively reveal the superiority of this technique over KiloNeRF \cite{paper166} and SNeRG \cite{paper167} in terms of PSNR and SSIM metrics. This superiority is attributed to the complexities in deriving an effective duplex mesh from NeRF, making its application more arduous in scenarios involving transparent or semi-transparent environments.

{\bf{MobileNeRF}} \cite{paper161} (2023) introduces an innovative NeRF representation technique for the effective creation of novel perspective images on mobile devices, utilizing textured polygons and conventional rendering processes. By depicting NeRF as a collection of polygons with binary opacity and feature vectors, and employing z-buffer and fragment shaders for rendering, this method attains a speed increase of 10 times over SNeRG  \cite{paper167}, all while preserving identical output quality. Furthermore, MobileNeRF now operates on mobile phones and tablets, previously incapable of offering interactive NeRF visualization. The system struggles with scenes that are semi-transparent, owing to its reliance on binary opacity.

{\bf{VI-NeRF-SLAM}} \cite{paper162} (2024)  merges the benefits of a traditional SLAM method with NeRF mapping, facilitating immediate, dense 3D scene reconstruction online without prior training. The estimation of pixel depth and pose is based on constraints in reprojecting feature points, whereas the depiction of 3D scenes is refined through the use of ray integration and networks. Furthermore, the proposal includes a dynamic error correction technique for IMUs and a method for managing closed loops in NeRF mapping. Tests on Euroc \cite{paper041} and Replica \cite{paper042} datasets reveal that VI-NeRF-SLAM surpasses current benchmark techniques in reconstruction and tracking precision, particularly in scenarios limited to grayscale imagery. For real-time reconstruction of 3D scenes, this technique eliminates the regularization loss and mesh depiction in NeRF mapping, leading to an increase in artifacts and flaws in the mesh representation relative to certain current mapping approaches.

\subsection{NeRF-based Improvement of Imaging Barriers}
NeRF technology harnesses intricate lighting and geometric nuances in a scene through deep learning models, facilitating the creation of superior new views. Nonetheless, NeRF continues to face significant hurdles in imaging, including occlusion, reflection, and handling transparent objects. In response to these challenges, scientists have suggested multiple approaches, encompassing enhanced network structures, creative loss functions, and sophisticated rendering methods. Below are a few principal areas for enhancement, as illustrated in Table \ref{tab:imaging_barriers_table}.

\begin{table*}[!h]
    \centering
    \caption{Overview of Articles Related to Improvement of Imaging Barriers.}
    \label{tab:imaging_barriers_table}
    \begin{tabular}{c||c||l} \hline
    Category                                                                                                     & Model        & \multicolumn{1}{c}{Highlight}                                                                                                                                                                                \\ \hline
    \multirow{12}{*}{\thead{Reduction of 3D \\ imaging blur}}                                                     & ScatterNeRF\cite{paper090}                                       & \multicolumn{1}{m{12cm}}{High-quality scene reconstruction and de-fogging effects are achieved by decomposing the scattering medium and scene objects.}                                                                                                                                                             \\ \cline{2-3}
                                                                                                                 & NeRF-MS\cite{paper092}      & \multicolumn{1}{m{12cm}}{Utilizes ternary loss to normalize the distribution of appearance codes and proposes sequential transient codes for better separation of non-static objects, leading to 3D consistent rendering and controlled appearance effects.}                                                      \\ \cline{2-3}
                                                                                                                 & DP-NeRF\cite{paper168}      & \multicolumn{1}{m{12cm}}{Geometric and appearance consistency in 3D space is maintained by introducing two physical scene priors, and an Adaptive Weight Proposal (AWP) module is proposed to improve the color combination weights to improve perceptual quality and 3D consistency when synthesizing new views.} \\ \cline{2-3}
                                                                                                                 & ABLE-NeRF\cite{paper094}    & \multicolumn{1}{m{12cm}}{The rendering quality is significantly improved by introducing Learnable Embeddings to capture viewpoint-dependent effects in the scene.}                                                                                                                                                 \\ \cline{2-3}
                                                                                                                 & E2NeRF\cite{paper145}       & \multicolumn{1}{m{12cm}}{By merging information from a bio-inspired event camera with a conventional RGB camera, precise three-dimensional volumetric models are efficiently derived from fuzzy images through the introduction of blur rendering loss and event rendering loss.}                                  \\ \hline
    \multirow{9}{*}{\thead{Reduction of 3D \\ imaging artifacts}}                                                & Nerfbusters\cite{paper169}  & \multicolumn{1}{m{12cm}}{Reducing distortions in NeRF optimization through the application of a local 3D prior and an innovative density-oriented fractional distillation sampling loss.}                                                                                                                          \\ \cline{2-3}
                                                                                                                 & Strata-NeRF\cite{paper170}  & \multicolumn{1}{m{12cm}}{Scenes of a hierarchical nature, marked by sudden alterations in their structure, are efficiently represented through the use of vector quantization(VQ) latent representation, subtly encapsulating multi-layered scenes.}                                                             \\ \cline{2-3}
                                                                                                                 & DiffusioNeRF\cite{paper093} & \multicolumn{1}{m{12cm}}{An approach to standardize NeRF through the application of Denoising Diffusion Models(DDMs) for learning, aiming to enhance the scene's geometry and color schemes.}                                                                                                                     \\ \cline{2-3}
                                                                                                                 & RobustNeRF\cite{paper171}   & \multicolumn{1}{m{12cm}}{An approach that enhances NeRF efficiency in reconstructing static scenes through a sturdy loss function, designed to disregard distractions like moving objects, alterations in lighting, shadows, and so on. within the training dataset.}                                              \\ \hline
    \multirow{6}{*}{\thead{Improvement of \\ refraction}}                                                        & NeRFrac\cite{paper091}      & \multicolumn{1}{m{12cm}}{Innovative synthesis of perspectives from refracted surfaces, like underwater ones, is accomplished by integrating a multilayer perceptron (MLP) refractive field to calculate the distances between light rays and refracted surfaces, utilizing Snell's law for these calculations.}    \\ \cline{2-3}
                                                                                                                 & MS-NeRF\cite{paper172}      & \multicolumn{1}{m{12cm}}{The suggestion is for a Multi-Spatial Neural Radiation Field(MS-NeRF), employing a series of eigenfields in concurrent subspaces to depict the scene, thereby enhancing the neural network's comprehension of objects that are reflected or refracted.}                                  \\ \cline{2-3}
                                                                                                                 & SeaThru-NeRF\cite{paper173} & \multicolumn{1}{m{12cm}}{A novel rendering model for NeRFs in scattering media, derived from the SeaThru image formation model, has been created, along with a proposed architecture for acquiring scene details and media parameters.}                                                                            \\ \hline
    \multirow{9}{*}{\thead{Improvement of\\ reflection}}                                                         & ScaNeRF\cite{paper174}      & \multicolumn{1}{m{12cm}}{A scalable neural radiation field framework, markedly enhances the precision of extensive scene depiction and camera pose accuracy through the use of a tiled hybrid neural field model and a parallel distributed optimization approach.}                                                \\ \cline{2-3}
                                                                                                                 & Ref-NeRF\cite{paper095}     & \multicolumn{1}{m{12cm}}{The implementation of a reflection direction parameterization and an organized view-dependent visual representation markedly enhances the realism and precision in depicting 3D scenes with shiny surfaces.}                                                                              \\ \cline{2-3}
                                                                                                                 & VDN-NeRF\cite{paper175}     & \multicolumn{1}{m{12cm}}{Resolving the ambiguity in shape-radiance during the reconstruction of geometries using NeRF on non-Lambertian surfaces and under varying lighting conditions is achieved by standardizing the visible dependency and isolating the constant information.}                                \\ \cline{2-3}
                                                                                                                 & Season-NeRF\cite{paper176}  & \multicolumn{1}{m{12cm}}{The NeRF model has been enhanced by incorporating seasonal variations and sun angle factors, facilitating the creation of new perspectives with distinct seasonal traits and accurate shading derived from satellite images.}                                                  \\ \hline                                              
    \end{tabular}
\end{table*}


\subsubsection{Reduction of 3D imaging blur}
{\bf{ScatterNeRF}} \cite{paper090} (2023)  introduces a physics-driven reverse neural rendering technique for reconstructing scenes of high quality in foggy environments. By integrating the Koschmieder [178] scattering model into the process of volume rendering, this technique acquires a distinct understanding of the scattering volume and scene objects, utilizing physical heuristic losses for scene reconstruction learning. Utilizing this method allows for optimization across various perspectives in a multi-view car sequence, eliminating the need for extensive training data. Confirmed through the field and controlled environment multi-view data, ScatterNeRF surpasses current techniques in the rendering, de-fogging, and depth estimation of foggy environments. Nonetheless, it might encounter difficulties in managing severe foggy environments, particularly with the high level of fog inconsistency.

{\bf{NeRF-MS}} \cite{paper092} (2023) diminishes the vagueness in visual alterations by implementing a ternary loss to standardize the spread of visual codes across each image, and suggests a temporary decomposition unit to more effectively distinguish still scenes from dynamic objects. Trials using both real-life outdoor and artificial datasets reveal that NeRF-MS attains cutting-edge results in multi-sequence scenarios, markedly enhancing rendering efficiency through the reconstruction of 3D-consistent visuals and minimizing ghost effects. The technique might encounter difficulties in managing significant sequence discrepancies, like severe alterations in the visual aspect of scenes caused by the activation of lights.

\begin{figure}[h!]
    \centering
    \includegraphics[width=0.45\textwidth]{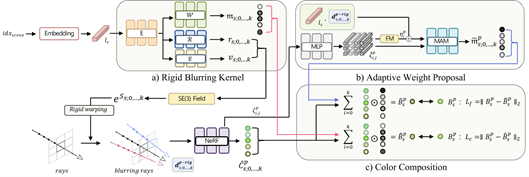}
    \caption{Overall pipeline for DP-NeRF.}
    \label{DP_NeRF}
\end{figure}

{\bf{DP-NeRF}} \cite{paper168} (2023)  is employed to create sharp three-dimensional visuals from fuzzy images. This technique employs two foundational physical concepts, the Rigid Blur Kernel(RBK) and the Adaptive Weighting Proposal(AWP), to maintain uniformity in geometry and appearance across three-dimensional space, with the model's architectural layout depicted in Fig. \ref{DP_NeRF}. Tests conducted on both artificial and actual scene datasets reveal that DP-NeRF surpasses current techniques in assessment metrics like PSNR, SSIM, and LPIPS, achieving sharp, high-quality images with superior perceptual clarity and three-dimensional coherence. Challenges persist in managing motion blur within dynamic environments, and subsequent research might investigate incorporating a time-based modeling element into DP-NeRF to overcome these difficulties.

\subsubsection{Reduction of 3D imaging artifacts}
\begin{figure}[h!]
    \centering
    \includegraphics[width=0.45\textwidth]{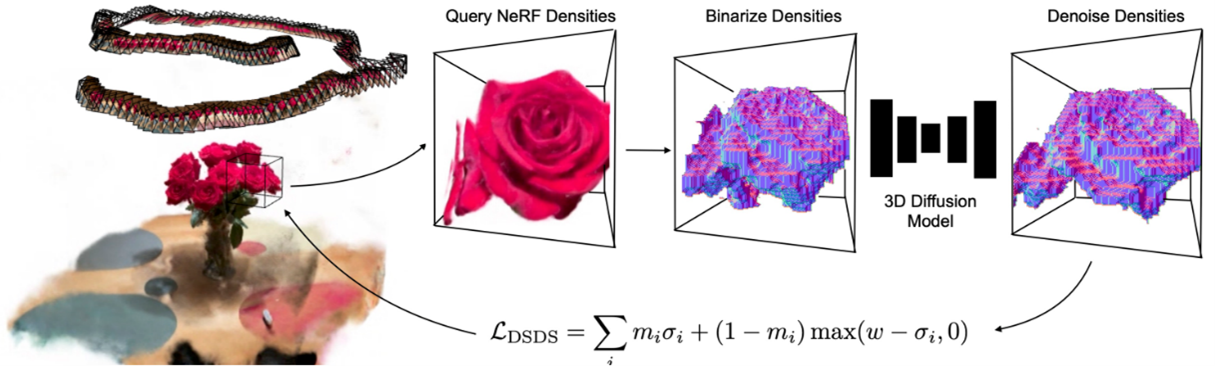}
    \caption{Overview of Nerfbusters Model.}
    \label{Nerfbusters}
\end{figure}

{\bf{Nerfbusters}} \cite{paper169} (2023) suggested a method based on 3D diffusion to enhance NeRF geometry and minimize floating-point anomalies, employing a local 3D prior and an innovative density-based score distillation sampling loss, with the model's structure depicted in Fig. \ref{Nerfbusters}. The experimental findings indicate that this technique excels in image quality indicators like PSNR, SSIM, and LPIPS, enhancing the precision of depth and standard estimation. Presently, this technique can only refine geometric patterns without altering textures, indicating its inability to colorize areas of low confidence in NeRF. Subsequent research might investigate the integration of 2D diffusion prior and 3D consistency repair methods to tackle these problems.

{\bf{Strata-NeRF}} \cite{paper170} (2023) introduces a representation based on potential vector quantization   (VQ), adept at implicitly depicting multi-layered scenes and simulating sudden alterations in their structure through a conditional NeRF. The technique undergoes assessment using both a multi-layered synthetic dataset and a real-world dataset, demonstrating its proficiency in depicting stratified scenes, reducing artifacts, and creating high-fidelity views, surpassing current methods.

{\bf{DiffusioNeRF}} \cite{paper093} (2023) standardizes the NeRF through the integration of a denoising diffusion model (DDM) as an initial learning factor. This technique employs the gradient data from RGBD patches, acquired by the DDM through training on an artificial dataset, to steer the NeRF in aligning more closely with the scene's geometry and color fields throughout the training phase. Tests conducted on the LLFF \cite{paper033} and DTU \cite{paper040} datasets reveal that DiffusioNeRF outperforms the current method (SOTA) in the novel perspective synthesis task.

\subsubsection{Improvement of Refraction}
{\bf{NeRFrac}} \cite{paper091} (2023) tackles NeRF's constraints in handling refractive surfaces, like subaquatic environments, through the integration of a multilayer perceptron   (MLP)-based refractive field. This technique calculates the gap between a ray and its refracted counterpart, calculating the refracted ray in line with Snell's law. This allows for precise depiction of subaquatic environments and the rebuilding of these refracted areas, ensuring uniformity across multiple views. NeRFrac successfully generates scenes devoid of refraction impacts by eliminating the acquired refraction field, showcasing its proficiency in creating and reconstructing these underwater scenes. Upcoming research aims to broaden the technique to address intricate multi-layered refraction and various light phenomena like scattering and focal dispersion.

{\bf{MS-NeRF}} \cite{paper172} (2023) proposes a multi-space neural radiation field approach that better handles reflective and refractive objects by representing the scene as a set of eigenfields in a parallel subspace. This approach serves as an enhancement to existing NeRF methods and requires only a small additional computational overhead to train and infer the outputs of the additional space. Experiments on datasets containing complex reflections and refractions show that MS-NeRF significantly outperforms existing single-space NeRF methods in terms of rendering quality, especially when dealing with specular objects. Although MS-NeRF has made significant progress in dealing with reflective surfaces, it may require more subspaces to accurately model when dealing with more complex reflective and refractive scenes.

{\bf{SeaThru-NeRF}} \cite{paper173} (2023) introduces an innovative NeRF model for visualizing images in dispersed mediums like fog, haze, and subaquatic settings. Incorporating a model for creating scattering images into the NeRF rendering formulas enables the model to distinguish between the  "unblemished" and the backscattered segments of the scene, thereby producing a lifelike new image perspective while maintaining the medium effect, or eliminating the medium effect and reinstating a distinct view of the scene. SeaThru-NeRF showcases its proficiency in depicting scenes in both simulated and actual settings, resulting in cutting-edge, realistic new-view composites. Furthermore, this technique can restore the scene's hue and deduce its three-dimensional configuration, particularly in far-off areas that are typically hard to detect in scattering media.

\subsubsection{Improvement of Reflection}
{\bf{ScaNeRF}} \cite{paper174} (2023) presents a scalable bundle-tuned neural radiation field (NeRF) method for large-scale scene rendering. The method combines a tile-based hybrid neural field and parallel distributed optimization to achieve global consistency of the camera pose via the alternating direction multiplier method (ADMM). In addition, the method efficiently handles complex reflections in both indoor and outdoor scenes by representing the local radiation field of each tile through multi-resolution hash grids and shallow chained multilayer perceptrons (MLPs). Compared with existing neural scene rendering methods, ScaNeRF improves PSNR and SSIM metrics by 5\%-10\% on average. In experiments on both indoor and outdoor scenes, the method demonstrates superior performance, especially when dealing with high-frequency textured regions and thin objects. However, since its representation only allows reflections to be represented at each 3D point via background encoding or foreground MLP, it may not be accurate enough when dealing with reflective and transmissive parts of glass windows.

{\bf{Ref-NeRF}} \cite{paper095} (2022) significantly improves the realism and accuracy of NeRF in rendering complex scenes with glossy surfaces by introducing a new parameterization method and a structured view-dependent appearance representation. The method improves the ability to interpolate glossy and reflective appearances by using reflection directions about local normals as input instead of view directions. In addition, Ref-NeRF introduces an integrated direction encoding technique and decomposes the output radiation into explicit diffuse and specular reflection components, allowing smooth interpolation even in scenes with material and texture variations. Experiments on synthetic datasets and real captured scenes show that Ref-NeRF outperforms previous top neural view synthesis methods in terms of rendering quality, especially on glossy objects. In addition, Ref-NeRF's structured output radiation enables scene editing, which is not possible with standard NeRF. However, it is computationally expensive and the integrated orientation encoding for evaluating reflection directions is slower than standard positional encoding.

{\bf{VDN-NeRF}} \cite{paper175} (2023) addresses the challenge of geometrically reconstructing under non-Lambertian surfaces and varying lighting conditions via normalization dependent on the viewpoint. This technique standardizes perspective reliance by deriving consistent data from the acquired NeRF, as opposed to directly simulating intricate viewpoint-dependent occurrences. The technique is straightforward and efficient in reducing the effects of shape-radiation uncertainty on reconstructing geometries. This technique attains superior view creation and geometric rebuilding by maintaining the volume rendering method constant via the self-distillation feature, trained alongside the frequently utilized photo-reconstruction loss. While VDN-NeRF excels in handling dynamic lighting and non-Lambertian surfaces, its capability to adapt to intricate lighting scenarios might be restricted. Furthermore, this technique depends on NeRF's learning potential, necessitating additional adjustments and enhancements for scenarios with significant alterations in lighting or intricate surface characteristics.

\subsection{NeRF-based Self-Supervised Learning}
The use of self-supervised learning for training models sans costly labeled data has garnered significant interest in computer vision and machine learning domains. Nonetheless, conventional NeRF training typically depends on accurate camera settings and labeling at the pixel level, a method frequently impractical for real-world use. In response to these constraints, scientists are investigating NeRF-based autonomous learning techniques that employ the scene's geometric and lighting details as training cues, bypassing direct labeling. Below are a few principal areas for enhancement, as displayed in Table \ref{tab:self-supervised_learning_table}.

\begin{table*}[!h]
    \centering
    \caption{Overview of Articles Related to Self-Supervised Learning.}
    \label{tab:self-supervised_learning_table}
    \begin{tabular}{c||c||l} \hline
    Category                                    & Model     & \multicolumn{1}{c} {Highlight}                                                                                                                                                                                                                             \\ \hline
    \multirow{10}{*}{\thead{Self-Supervised \\Learning}} & SceneRF\cite{paper096}   & \multicolumn{1}{m{12cm}}{By optimizing the NeRF and combining explicit depth optimization with a novel probabilistic sampling strategy, it is possible to achieve 3D reconstruction of complex scenes using only image sequences with pose for training without depth supervision.} \\ \cline{2-3}
                                                & MIMO-NeRF\cite{paper097} & \multicolumn{1}{m{12cm}}{A self-supervised learning method is introduced to solve the ambiguity problem of color and volume densities induced by the input coordinate sets, enabling fast and high-quality synthesis of new perspectives.}                                          \\ \cline{2-3}
                                                & CI-NeRF\cite{paper099}   & \multicolumn{1}{m{12cm}}{Interpretable and editable relighting on a single image is realized by complementarily estimating the intrinsic properties of the scene, improving the quality and realism of relighting outdoor scenes.}                                                  \\ \cline{2-3}
                                                & EventNeRF\cite{paper178} & \multicolumn{1}{m{12cm}}{The ability to self-supervise the learning of NeRF from a stream of events captured by a single color event camera to enable 3D-consistent, dense, and realistic synthesis of new views in RGB space at the time of testing.}                              \\ \cline{2-3}
                                                & CaFi-Net\cite{paper098}  & \multicolumn{1}{m{12cm}}{A self-supervised learning method is used for 3D position and orientation (pose) normalization of object classes represented as NeRF.}  \\ \hline                                                                                                                    
    \end{tabular}
\end{table*}
    

{\bf{SceneRF}} \cite{paper096} (2023) proposes a self-supervised monocular 3D scene reconstruction method trained using only image sequences with poses. The method efficiently handles large scenes through explicit depth optimization and a novel probabilistic sampling strategy. Experiments show that SceneRF outperforms existing methods for both new view synthesis and scene reconstruction tasks. In particular, for new depth view synthesis, SceneRF achieves the best performance in all metrics, including absolute relative error, root mean square error, and IoU.

{\bf{CI-NeRF}} \cite{paper099} (2023) utilizes NeRF for volumetric rendering to deduce essential scene characteristics like diffuse albedo, surface normals, shadows, and lighting settings, employing CNN for forecasting interpretable and modifiable lighting parameters from an individual image. Utilizing this method allows for the physical interpretation of outdoor scene relighting through a unified image inference process, enhancing both the quality and realism of the relighting. The experimental findings indicate that this technique surpasses current methods in terms of inherent image breakdown and real-world landmark scene relighting.

\begin{figure}[h!]
    \centering
    \includegraphics[width=0.45\textwidth]{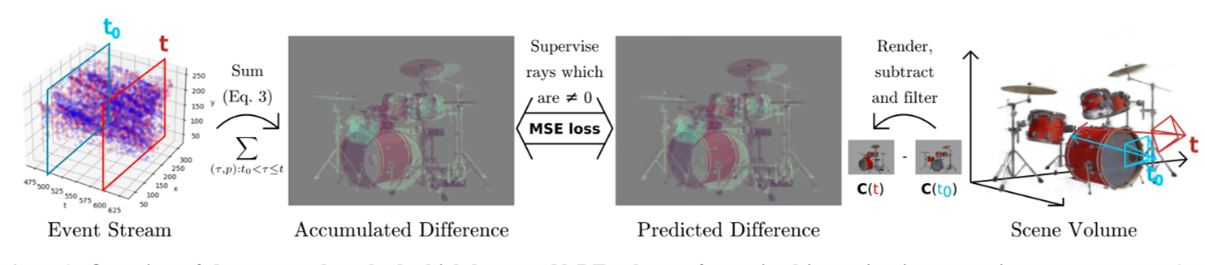}
    \caption{Overview of the proposed method which learns a NeRF volume of a static object using just a moving event camera.}
    \label{EventNeRF}
\end{figure}

{\bf{EventNeRF}} \cite{paper178} (2023) stands as the inaugural method for synthesizing views that maintain consistency, density, and realism in 3D, utilizing a singular color event stream as the input. Training of the NeRF occurs independently, maintaining the clarity of the initial color event channel. Furthermore, the paper introduces a tailored light sampling approach for events, facilitating effective data training, with the rendering procedure depicted in Fig. \ref{EventNeRF}. Evaluations, both qualitative and quantitative, on various complex synthetic and lifelike scenarios, reveal that EventNeRF surpasses current techniques in rendering clarity, resilience in managing rapid movement, and dim lighting. Despite EventNeRF's notable advancements in managing the event stream's sparse and asynchronous characteristics, it continues to encounter difficulties in handling shadows within singular images. Subsequent research might investigate the integration of user engagement or graphic methods to tackle this problem. Furthermore, this technique might require further enhancements in handling dynamic obstructions and shifting objects.

\subsection{NeRF-based Monocular New View Synthesis}
New viewpoint synthesis aims to recreate fresh perspective images using a collection of images with established viewpoints, holding significant relevance in areas like virtual reality, augmented reality, and self-driving technology. Lately, techniques based on NeRF have achieved notable advancements in synthesizing new viewpoints using monocular vision.

{\bf{MonoNeRD}} \cite{paper100} (2023) proposes a novel monocular 3D target detection framework that achieves a dense 3D geometric and occupancy representation by modeling the scene as a signed distance function (SDF). The method employs volume rendering to recover RGB images and depth maps from the SDF and radial fields, which is the first time volume rendering has been introduced into the M3D domain, demonstrating the potential for implicit reconstruction of image-based 3D perception. MonoNeRD demonstrates superiority in handling long-range and occluded objects, setting a new benchmark for monocular 3D detection.

{\bf{MonoNeRF}} \cite{paper102} (2023) presents a method for learning generic dynamic radiation fields from monocular video. Unlike existing NeRF methods based on multiple views, this method learns both point features and scene streams through video feature extraction, implicit velocity fields, and stream-based feature aggregation modules. This method is capable of learning in multiple scenes and supports new applications such as scene editing, unseen frame synthesis, and fast new scene adaptation. Experiments show that MonoNeRF can render new views in multiple dynamic videos and achieve better performance on unseen frames.

{\bf{HOSNeRF}} \cite{paper103} (2023) proposed a novel 360° free-view rendering method capable of reconstructing dynamic human-object-scene neural radiation fields from a single monocular video. The method effectively estimates the complex object motion during human-object interaction by introducing new object skeletons into the traditional human skeletal hierarchy. In addition, two learnable object state embeddings are introduced for the case of interacting with different objects at different times.HOSNeRF significantly outperforms existing techniques on two challenging datasets, with a 40\% to 50\% improvement on the LPIPS metric.

\subsection{NeRF-based Posture Estimation}
The effective training and utilization of NeRF depend on precise data regarding the camera's location. Acquiring precise camera positions in real-world scenarios is frequently difficult, particularly in outdoor settings or extensive scenes. Consequently, the challenge of conducting proficient NeRF training and determining positions without precise positional data has emerged as a significant subject in contemporary research.

{\bf{IR-NeRF}} \cite{paper105} (2023) proposes a novel NeRF training method without camera poses that improves the robustness of pose estimation on real images by introducing implicit pose regularization. The method constructs a scene codebook that implicitly captures the scene-specific camera pose distribution as a priori, thereby training an effective NeRF in the absence of an accurate camera pose. Extensive experiments on multiple synthetic and real datasets show that IR-NeRF outperforms the current state-of-the-art in new view synthesis and consistently outperforms the state-of-the-art in terms of image quality metrics (PSNR, SSIM, and LPIPS) consistently outperform GNeRF[180].IR-NeRF also performs well in terms of pose estimation accuracy in real images, reducing the estimation of anomalous poses.

{\bf{DBARF}} \cite{paper106} (2023) proposes a new deep bundle adjustment method for optimizing camera poses in conjunction with generalized neural radiation fields (GeNeRFs). The method is able to train with GeNeRFs in a self-supervised manner without the need for an accurate initial camera pose by constructing a residual feature map as an implicit cost function. Unlike existing methods such as BARF \cite{paper180}, DBARF does not rely on scene-specific optimizations, is able to generalize across scenes, and does not require a good initial bit pose. Evaluation of the LLFF \cite{paper033} dataset shows that DBARF achieves higher scores in PSNR, SSIM, and LPIPS metrics.

{\bf{LU-NeRF}} \cite{paper107} (2023) jointly estimates camera pose and scene representation by synchronizing locally unlocalized NeRFs. The method employs a local-to-global learning framework, which first optimizes on a subset of local data (called "mini-scenes"), then brings the poses of these mini-scenes into a global reference frame through a robust pose synchronization step, and finally performs fine-grained optimization of the global poses and scenes. Experimental results show that LU-NeRF outperforms previous techniques on the tasks of camera pose estimation and new view synthesis, and succeeds even in the absence of a priori knowledge of the pose distribution.

\subsection{NeRF-based Other Methods}
In addition to the aforementioned technologies, NeRF has shown potential in deep prior modeling, self-supervised learning frameworks, and end-to-end processing flows.

{\bf{SparseNeRF}} \cite{paper037} (2023) addresses the problem of synthesizing new views from finite perspectives by extracting depth priors from pre-trained depth models or rough depth maps captured by consumer-grade depth sensors. The method introduces local depth ordering and spatial continuity constraints to extract robust depth priors from rough depth maps, thereby improving the performance of new view synthesis while maintaining geometric spatial continuity. Extensive experiments on the LLFF \cite{paper033}, and DTU \cite{paper040} datasets show that SparseNeRF achieves new state-of-the-art performance in new view synthesis, outperforming all existing methods including depth-based models.SparseNeRF relies on pre-trained depth models or depth maps captured by consumer-grade sensors, the accuracy and quality of which can be limited by real-world conditions. Future work may explore how to further improve the accuracy of depth maps and how to better handle occlusion and dynamic scenes.

{\bf{IntrinsicNeRF}} \cite{paper110} (2023) possesses the capability to break down a static scene's multi-view image into consistent elements like reflectance, shading, and residual layers, achieved through intrinsic decomposition in a NeRF-based neural rendering technique. This method facilitates web-based applications like changing scenes, altering lighting, and creating editable new perspectives, while also allowing IntrinsicNeRF to undergo autonomous training for consistent multi-viewpoint intrinsic decomposition, introducing an innovative distance-conscious point-sampling and iterative clustering optimization technique for adaptive reflectance. While IntrinsicNeRF excels in indoor settings (like Replica scenes), its application to outdoor environments with extensive scenes and limited images might result in diminished detail in the rendering outcomes.

\section{NeRF-based Application Scenarios}\label{section:applications}
Since the advent of NeRF technology, it has promoted technological advancement in various fields such as computer vision, virtual reality (VR), augmented reality (AR), etc. Additionally, NeRF has shown significant potential and application value in areas such as robotics, urban planning, autonomous driving navigation, etc. In this section, we will provide a detailed overview of the latest developments up to March 2024, with relevant application fields shown in Table \ref{tab:appilcations_table}.

\begin{table*}[!h]
    \centering
    \caption{Overview of NeRF-based Application Scenarios}
    \label{tab:appilcations_table}
    \begin{tabular}{c||l} \hline
    Applications                                  & \multicolumn{1}{c}{Related Models}                                                                 \\  
    \hline
    Indoor 3D Reconstruction                      & EgoNeRF\cite{paper111}, SurfelNeRF\cite{paper112}, StructNeRF\cite{paper113}, NerfingMVS\cite{paper114}, etc.                                                   \\ \hline
    Portrait Reconstruction                       & ER-NeRF\cite{paper115}, ActorsNeRF\cite{paper116}, HairNeRF\cite{paper117}, FaceCLIPNeRF\cite{paper118}, NeRFInvertor\cite{paper119}, etc.                   \\ \hline
    Human Rendering                               & TransHuman\cite{paper122}, SHERF\cite{paper123}, GM-NeRF\cite{paper124}, NerfCap,\cite{paper125} PersonNeRF\cite{paper126}, AvatarReX\cite{paper127}, etc.                           \\ \hline
    Editable, interactive NeRF                    & Magic3D\cite{paper129}, SKED\cite{paper130}, Instruct-NeRF2NeRF\cite{paper131}, NaviNeRF\cite{paper132}, SPIn-NeRF\cite{paper133}, ICE-NeRF\cite{paper134}, etc. \\ \hline
    3D Perception Techniques & Instance-NeRF\cite{paper137}, Nerflets\cite{paper138}, FeatureNeRF\cite{paper139}, NeRF-RPN\cite{paper140}, NeRF-Det\cite{paper109}, etc.                                     \\ \hline
    Others                                        & StegaNeRF\cite{paper210}(text steganography), CuNeRF\cite{paper142}(medical field), CLNeRF\cite{paper143}(continuous learning), etc.        \\ 
    \hline      
    \end{tabular}
\end{table*}


\subsection{NeRF-based Indoor 3D Reconstruction}
As 3D modeling and virtual reality technologies advance swiftly, the creation of 3D representations of indoor environments has emerged as a key area of study in computer graphics. NeRF technology enables the generation of superior new perspectives from restricted multi-view images by harnessing the intricate lighting and geometric nuances of indoor settings via deep learning models. Nonetheless, the varied and intricate nature of indoor environments, including a range of furniture, differing textures in walls and floors, and fluctuating indoor lighting, poses a challenge to the rendering efficiency of NeRF.

\begin{figure}[h!]
    \centering
    \includegraphics[width=0.45\textwidth]{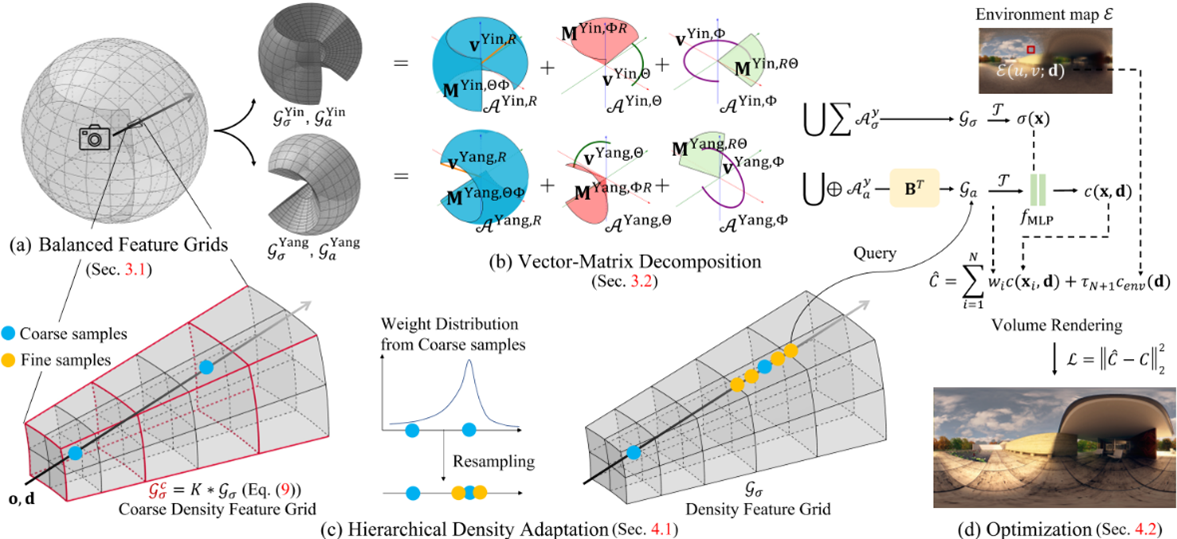}
    \caption{Overview of the EgoNeRF Method.}
    \label{EgoNeRF}
\end{figure}

{\bf{EgoNeRF}} \cite{paper111} (2023) introduces a feasible method for creating extensive real-world settings for virtual reality assets. The system enhances the representation of NeRF's feature grid to more accurately align with egocentric images, utilizing spherical coordinates over the conventional Cartesian coordinates. By merging two evenly distributed grids, this method circumvents singularities close to the poles and employs an exponentially segmented radial grid along with an infinite environment map to depict infinite scenes. Furthermore, a streamlined hierarchical sampling technique has been developed to augment the effective sample count for NeRF volume training, as depicted in Fig. \ref{EgoNeRF}. A thorough analysis of the dataset reveals that EgoNeRF surpasses other comparative techniques across all error measures. Nonetheless, it remains constrained in addressing real-world issues, like the photometric fluctuations resulting from automated camera exposures.

{\bf{SurfelNeRF}} \cite{paper112} (2023) presents a new method for online reconstruction and rendering of large-scale interior scenes. The method combines the complementary advantages of classical 3D reconstruction and NeRF rendering using a flexible and extensible neural voxel (surfel) representation to store geometric attributes and appearance features extracted from the input images. Furthermore, the method extends the traditional voxel-based fusion scheme to incrementally integrate new input frames into the reconstructed global neural scene representation and proposes an efficient differentiable rasterization scheme for rendering the neural voxel radiation field, which significantly improves the training and inference speed. Experimental results show that SurfelNeRF achieves 23.82 PSNR in the no-scene-optimization setting, which outperforms existing techniques. Due to the imperfect quality of the captured or estimated depth, there may be geometric discontinuities or missing parts, which may result in missing parts or incorrect geometry in the rendered output.

{\bf{NerfingMVS}} \cite{paper114} (2023) merges conventional reconstruction methods with pre-existing learning knowledge to directly enhance the implicit volume, circumventing the difficulty of aligning pixels in indoor settings. The essence of this technique lies in steering the NeRF enhancement process through the application of a deep learning-driven prior. Initially, the system fine-tunes the monocular depth network for the intended scene using COLMAP's MVS reconstruction. Subsequently, it employs this adjusted depth to steer the NeRF's volumetric rendering sampling, calculating errors in the resulting images to derive pixel-level confidence maps, thereby enhancing depth accuracy. Nonetheless, the present technique lacks the efficiency for real-time system application, and upcoming efforts will concentrate on integrating this method with current rapid NeRF rendering and convergence techniques for instantaneous reconstruction.

\subsection{NeRF-based Portrait Reconstruction}
In recent years, NeRF-based methods have attracted much attention in the field of portrait reconstruction. The complexity and diversity of faces, as well as the challenges posed by changes in expression, illumination, and pose, make NeRF-based portrait reconstruction still a challenging task. The following are some key directions for improvement, as shown in Table \ref{tab:portrait_reconstruction_table}.

\begin{table*}[!h]
    \centering
    \caption{Overview of Articles Related to Portrait Reconstruction.}
    \label{tab:portrait_reconstruction_table}
    \begin{tabular}{c||c||l} \hline
    Category                                             & Model          & \multicolumn{1}{c}{Highlight}                                                                                     \\ \hline 
    \multirow{8}{*}{\thead{Talking Portrait \\Compositing}}        & ER-NeRF\cite{paper115}        & \multicolumn{1}{m{11cm}}{An efficient region-aware NeRF architecture for generating high-fidelity dynamic portrait synthesis that guides portrait modeling by explicitly exploiting unequal contributions from spatial regions.}                                   \\ \cline{2-3}
                                                         & AvatarMAV\cite{paper181}      & \multicolumn{1}{m{11cm}}{By decoupling expression motion and canonical appearance and using explicit voxel representation, the training speed of NeRF-based head avatar reconstruction is significantly improved.}                                                                          \\ \cline{2-3}
                                                         & HiDe-NeRF\cite{paper182}      & \multicolumn{1}{m{11cm}}{By representing the 3D dynamic scene as a canonical appearance field and an implicit deformation field, it is realized to accurately mimic the motion of the driving image while preserving the identity information of the source image.}                         \\ \cline{2-3}
                                                         & HFFAR\cite{paper183}          & \multicolumn{1}{m{11cm}}{High-quality 3D facial avatars are reconstructed from monocular video by learning a personalized low-dimensional subspace to capture individual features.}                                                                                                         \\ \hline
    \multirow{4}{*}{\thead{Portrait Style \\Migratio}}            & MuNeRF\cite{paper121}         & \multicolumn{1}{m{11cm}}{A robust makeup migration method based on dynamic NeRF is proposed, which is capable of consistently migrating the makeup style of a reference image to a facial image with arbitrary pose and expression while maintaining geometric and appearance consistency.} \\ \cline{2-3}
                                                         & HairNeRF\cite{paper117}       & \multicolumn{1}{m{11cm}}{By considering the underlying head geometry of the input image, the neural rendering technique is utilized to align the two input heads in volumetric space, resulting in a seamless transfer of hairstyles over the target image head.}                           \\ \hline
    \multirow{9}{*}{\thead{Interactive Editable \\Portraits}}                          & FaceCLIPNeRF\cite{paper118}   & \multicolumn{1}{m{11cm}}{A text-driven deformable NeRF-based 3D facial manipulation method realizes fine control and manipulation of NeRF-reconstructed faces using only text by training scene manipulators and position-conditional anchor combiners (PACs).}                             \\  \cline{2-3}
                                                         & LC-NeRF\cite{paper120}        & \multicolumn{1}{m{11cm}}{A locally controllable NeRF facial generation and editing method that achieves fine control over the geometry and texture of localized facial regions by means of a local region generator module and a spatially-aware fusion module.}                            \\ \cline{2-3}
                                                         & AvatarStudio\cite{paper185}   & \multicolumn{1}{m{11cm}}{A text-driven text-based editing method for dynamic 3D human avatars that utilizes NeRF and a text-to-image diffusion model to achieve high-quality editing of dynamic human avatars that is consistent in view and time.}                                         \\ \cline{2-3}
                                                         & NeRF-Art\cite{paper186}       & \multicolumn{1}{m{11cm}}{Manipulating a pre-trained NeRF model's style with basic text cues allows for the manipulation of a 3D scene's visual and geometric diversity.}                                                                                                                    \\ \hline
    \multirow{15}{*}{\thead{High Fidelity Head \\Image Generation}} & NeRFInvertor\cite{paper119}   & \multicolumn{1}{m{11cm}}{Enhancements in 3D geometry and image rendering quality are achieved by refining the generator model and implementing both overt geometric limitations and subtle geometric regularization.}                                                                       \\ \cline{2-3}
                                                         & ActorsNeRF\cite{paper116}     & \multicolumn{1}{m{11cm}}{By conducting pre-training on a variety of human subjects and fine-tuning using a limited set of monocular video frames, we successfully integrated novel viewpoints on unfamiliar actors in different poses.}                                                     \\ \cline{2-3}
                                                         & GazeNeRF\cite{paper187}       & \multicolumn{1}{m{11cm}}{The precision in retargeting the line of sight and head pose is enhanced by dividing the feature volumes between the face and eye areas and using a fixed 3D rotation matrix, all the while preserving the uniformity of the identity features.}                   \\ \cline{2-3}
                                                         & RODIN\cite{paper188}          & \multicolumn{1}{m{11cm}}{The creation of superior, intricately detailed 3D avatars is accomplished through the application of 3D perceptual diffusion on a two-dimensional feature plane, coupled with the use of 3D perceptual convolution and latent condition control.}                  \\ \cline{2-3}
                                                         & CGOF plus plus\cite{paper189} & \multicolumn{1}{m{11cm}}{Precise control of the 3DMM model is achieved using conditionally generated occupation fields(cGOF++) and auxiliary volume loss, enabling the generation of face images with desired 3D shapes, expressions, and poses.}                                           \\ \cline{2-3}
                                                         & LatentAvatar\cite{paper190}   & \multicolumn{1}{m{11cm}}{A latent head NeRF model is proposed, which enables fine control of the head neural model by learning individual-specific latent expression codes.}                                                                                                                 \\ \cline{2-3}
                                                         & NOFA\cite{paper191}           & \multicolumn{1}{m{11cm}}{A disposable 3D face avatar reconstruction method based on NeRF technology is proposed to support high-fidelity 3D facial reconstruction and vivid facial reproduction.} \\ \hline                                                                                          
    \end{tabular}
    \end{table*}
    

\subsubsection{Talking Portrait Compositing}
{\bf{ER-NeRF}} \cite{paper115} (2023) introduces an innovative Conditional NeRF framework designed for effective, high-accuracy dynamic portrait creation. Utilizing spatial regions' distinct contributions to steer dynamic head modeling, a concise and articulate Tri-Plane Hash Representation is developed, and region-sensitive conditional attributes are produced through a Region Attention Module (RAM) to delineate the link between audio attributes and spatial areas. Furthermore, the introduction of rapid and straightforward Adaptive Pose Encoding is suggested to enhance the separation of the head and torso. ER-NeRF is renowned for its ability to produce dynamic portrait videos that are both high-quality and synchronized with audio, offering lifelike details and superior efficiency. ER-NeRF outperforms earlier techniques in objective assessments and human research, attaining instantaneous rendering and rapid convergence using a compact model size. While ER-NeRF excels in instantaneous rendering, additional fine-tuning might be necessary to sustain superior rendering outcomes for poses that exceed the scope of the training dataset.

\begin{figure}[h!]
    \centering
    \includegraphics[width=0.45\textwidth]{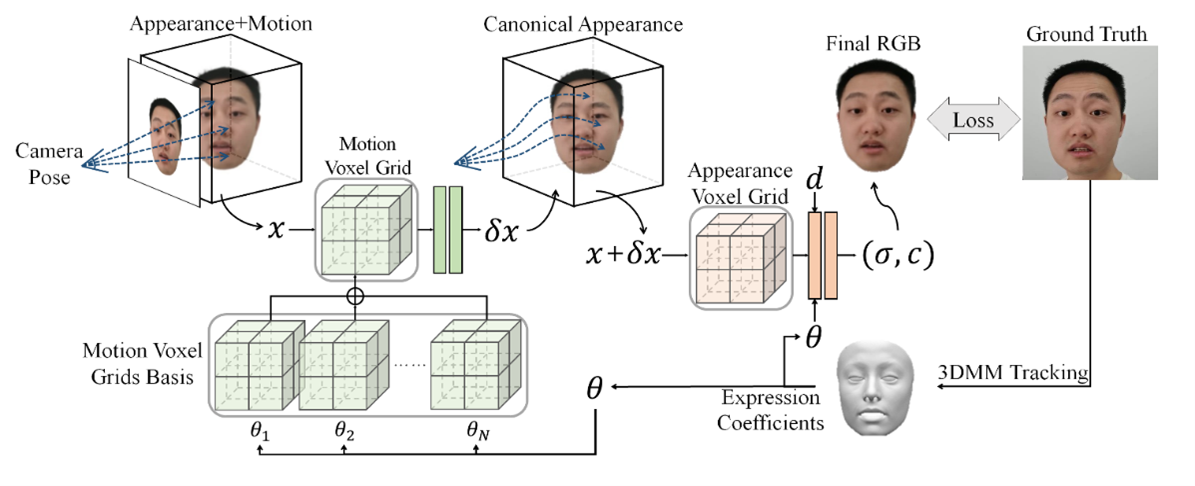}
    \caption{Overview of the AvatarMAV Model Framework.}
    \label{AvatarMAV}
\end{figure}

{\bf{AvatarMAV}} \cite{paper181} (2023) introduces a rapid method for reconstructing 3D head avatars, employing Motion-Aware Neural Voxels to expedite the training procedure. For the first time, this technique separates dynamic expression movement from standard appearance, employing the pre-existing knowledge of the 3DMM expression basis to simulate motion related to expression via voxel meshes and minor MLPs, with the model's structure depicted in Fig. \ref{AvatarMAV}. This method markedly enhances training effectiveness and facilitates the creation of superior 3D avatars in under 5 minutes. Inaccuracies in the camera's pose and expression coefficients can lead to artifacts and blurring, especially when the perspective is off-frontal or the expression is overstated. Subsequent efforts will focus on tackling these matters through a broader parameterized framework.

{\bf{HiDe-NeRF}} \cite{paper182} (2023) introduces an innovative neural radiation field technique for detailed, unobstructed dynamic head synthesis. This technique relies on flexible neural radiation fields, depicting a three-dimensional dynamic environment as a standard appearance field encompassing the standard source face, and a subtle deformation field that mimics the primary pose and expression. Hi-De-NeRF, equipped with a multi-scale generalized appearance module (MGA) and a streamlined expression-aware deformation module (LED), enhances expression precision while preserving identity data. Trials using various benchmark datasets reveal HiDe-NeRF's superiority over current methods in detecting driving movements and preserving source identity data.

\subsubsection{Portrait Style Migration}
{\bf{MuNeRF}} \cite{paper121} (2024) introduces a makeup migration technique based on dynamic NeRF, known as MuNeRF, capable of seamlessly transitioning makeup styles from a standard image to a facial image in various poses and expressions. This technique guarantees uniformity in geometry and looks by incorporating a hidden 3D model and an innovative mixed makeup loss function, enhancing the precision and uniformity of PatchGAN \cite{paper192}'s makeup creation using UV texture maps. MuNeRF has proven its dominance in various makeup genres, particularly regarding the visual authenticity and uniformity of facial images, as evidenced by comprehensive experiments and user research. The real-time rendering capabilities of the NeRF framework prevent MuNeRF from facilitating real-time makeup migration.

{\bf{HairNeRF}} \cite{paper117} (2023) employs a neural rendering method to position two input heads in three-dimensional space, enabling the adaptation of the transferred hairstyles to match the head of the target image. In contrast to standard techniques, this approach yields steadier and more authentic outcomes in both the task of transferring hairstyles and reconstructing them. The numerical data indicates that HairNeRF matches other techniques in naturalness, realism, and FID, and surpasses them in certain respects. Despite this, HairNeRF shows significant promise in hairstyle transfer and opens up fresh avenues for upcoming studies.

\subsubsection{Interactive Editable Portraits}
\begin{figure}[h!]
    \centering
    \includegraphics[width=0.45\textwidth]{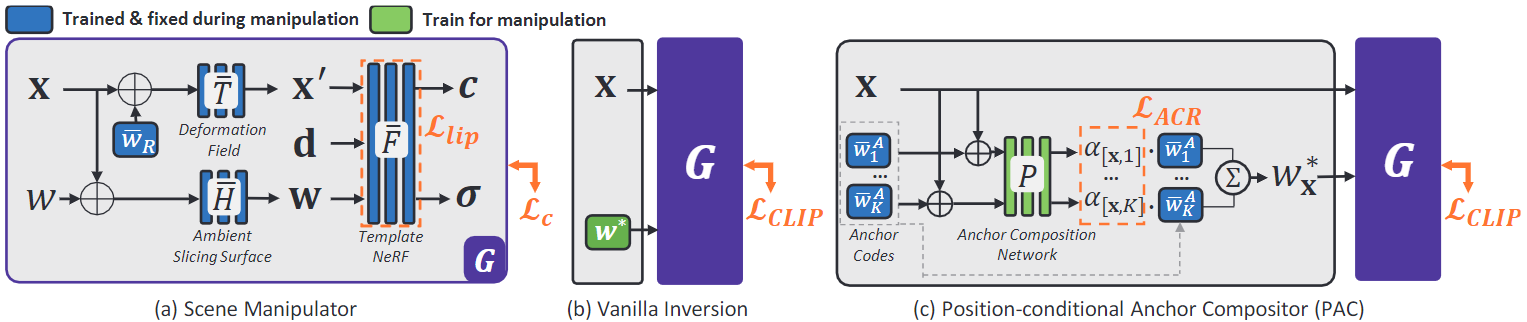}
    \caption{(a) Network structure of scene manipulator G. (b) Vanilla inversion method for manipulation. (c) Position-conditional Anchor Compositor (PAC) for manipulation.}
    \label{FaceCLIPNeRF}
\end{figure}

{\bf{FaceCLIPNeRF}} \cite{paper118} (2023) employs a text-based 3D facial alteration technique utilizing malleable NeRF. This technique manages facial distortion by instructing a scene manipulator to use hidden codes and adapt to depict different latent codes for scene manipulation using a Positional Conditional Anchor Combiner(PAC), as depicted in the module architecture of Fig. \ref{FaceCLIPNeRF}. For the first time, the concept of employing text to alter NeRF-reconstructed faces has been introduced, tackling the constraints of conventional deformable NeRFs in merging localized deformations noted in various cases. Via both qualitative and quantitative studies, FaceCLIPNeRF successfully mirrors the visual characteristics of descriptive and emotionally charged texts, preserving the visual integrity and essence of 3D faces.

{\bf{SketchFaceNeRF}} \cite{paper184} (2023) introduces a two-dimensional, sketch-oriented 3D facial NeRF creation and modification technique, integrating visual and three-dimensional data into drawings through the Sketch Tri-plane Prediction network, and creating superior facial NeRFs using a pre-trained generator. This approach demonstrates resilience to various drawing styles and facilitates control over appearances. SketchFaceNeRF surpasses current methods for generating and editing faces based on sketches in terms of both visual clarity and consistency in 3D views. Yet, in the case of excessively abstract or cartoonish drawings, the created 3D faces might fail to depict excessively exaggerated features.

{\bf{LC-NeRF}} \cite{paper120} (2023) introduces a method for locally adjustable facial creation and modification, enabling precise management of the shape and texture of specific facial areas via a local region generator unit and a spatially conscious fusion module. This technique permits users to modify specific facial areas like hair, nose, mouth, etc., autonomously, without impacting other parts. LC-NeRF surpasses current facial editing techniques in maintaining the stability of those areas and ensuring consistent facial identity. Furthermore, LC-NeRF facilitates text-driven facial editing, allowing for precise manipulation of facial attributes by directly refining possible codes from the text. Yet, precise management of specific internal textures, like hair and facial lines, remains unfeasible at present. Subsequent research aims to investigate methods for more precise regulation of local texture elements.

\subsubsection{High Fidelity Head Image Generation}
{\bf{NeRFInvertor}} \cite{paper119} (2023) introduced a broadened version of the NeRF-GAN inversion technique, aiming to produce animations that are high-accuracy, maintain 3D consistency, and preserve identity from an individual actual image. This technique eliminates both geometric and visual distortions by fine-tuning the underlying code and diminishing the identity discrepancy through a two-dimensional loss function, and by incorporating both overt and subtle three-dimensional regularization. The experimental findings indicate that NeRFInvertor surpasses current inversion techniques and single-use NeRF models in preserving authenticity, accuracy, and three-dimensional uniformity.

\begin{figure}[h!]
    \centering
    \includegraphics[width=0.45\textwidth]{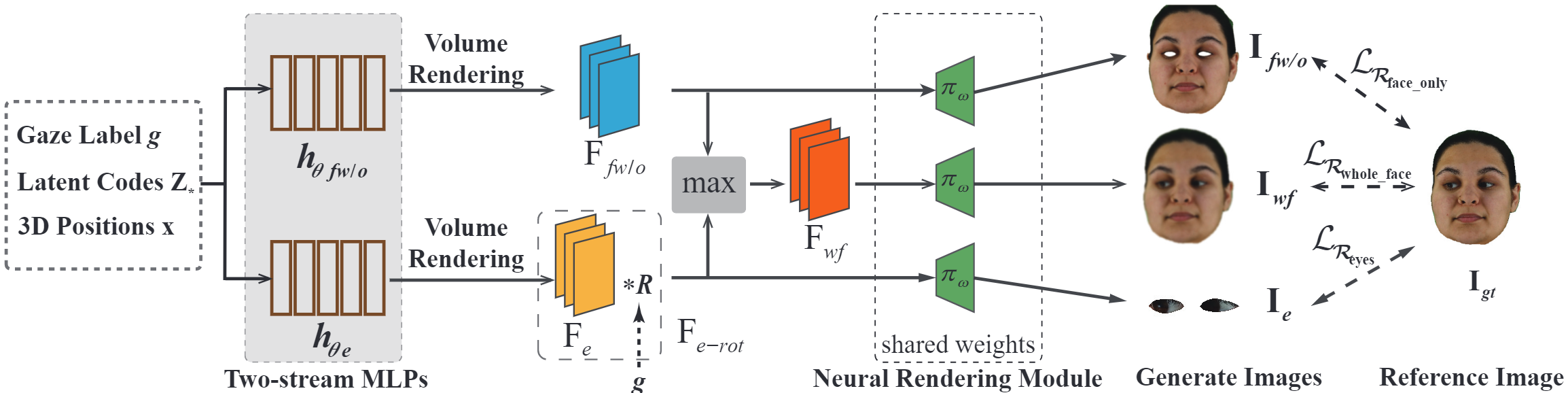}
    \caption{Overview of GazeNeRF Pipeline.}
    \label{GazeNeRF}
\end{figure}

{\bf{GazeNeRF}} \cite{paper187} (2023) introduces a method for 3D-aware gaze retargeting, aiming to produce superior gaze retargeting images. This was achieved by forecasting the facial and eye region volumetric characteristics using a dual-stream MLP framework and employing a 3D rotation matrix for precise gaze angle adjustment, with the implementation process depicted in Fig. \ref{GazeNeRF}. Research indicates that GazeNeRF surpasses current NeRF baseline and 2D gaze retargeting techniques across various datasets, notably in the accuracy of gaze and head pose retargeting and maintaining identity. Although GazeNeRF excels in gaze retargeting tasks, it faces a common constraint with NeRF models: prolonged training durations and subsequent efforts will focus on lessening the training load.

{\bf{RODIN}} \cite{paper188} (2023) accomplishes effective three-dimensional perceptual diffusion through the expansion of several two-dimensional feature maps from a 3D NeRF model onto a singular two-dimensional feature plane. Utilizing essential methods like 3D perceptual convolution, latent conditioning, and hierarchical synthesis, the model enhances the creation of 3D digital avatars, with RODIN excelling in producing high-quality 3D avatars, particularly in aspects like hairstyles and facial hair. Despite RODIN's notable advancements in creating 3D avatars, its training methodology might be skewed by data, necessitating meticulous choice of training data to prevent biased depictions. Furthermore, the RODIN model's issues with sampling speed and 3D data bottlenecks require additional investigation and enhancement.

\subsection{NeRF-based Human Rendering}
Lately, the domain of human rendering has experienced expansion due to the advent of NeRF technology. The intricate nature of modeling the human body, encompassing aspects like joint suppleness, muscular deformation, and evolving clothing styles, introduces new hurdles in employing NeRF. Below are several crucial areas for enhancement, as depicted in Table \ref{tab:human_rendering_table}.

\begin{table*}[!h]
    \centering
    \caption{Overview of Human Rendering Related Articles.}
    \label{tab:human_rendering_table}
    \begin{tabular}{c||c||l} \hline
    Category                                     & Model        & \multicolumn{1}{c}{Highlight}                                                                                                                                                                                                          \\ \hline
    \multirow{10}{*}{\thead{Less View \\Human Rendering}}   & SHERF\cite{paper123}        & \multicolumn{1}{m{12cm}}{A generalized human NeRF based on a single image enables high-quality rendering and animation from arbitrary viewpoints and poses by extracting and encoding 3D human representations in canonical space.}                              \\ \cline{2-3}
                                                 & OccNeRF\cite{paper193}      & \multicolumn{1}{m{12cm}}{A novel neural rendering approach achieves high-quality rendering of dynamic human bodies in heavily occluded scenes through a surface-based rendering strategy that integrates geometric and visibility priors.}                       \\ \cline{2-3}
                                                 & GM-NeRF\cite{paper124}      & \multicolumn{1}{m{12cm}}{A model-based generalized neural radiation field framework for synthesizing high-fidelity new-view images of arbitrary performers from multi-view images by combining geometrically guided attentional mechanisms and perceptual loss.} \\ \cline{2-3}
                                                 & CHPC\cite{paper194}         & \multicolumn{1}{m{12cm}}{Robust capture of dynamic clothing and human motion from sparse views or monocular video is achieved by tracking clothing and human motion separately and incorporating a physically aware clothing simulation network.}                \\ \cline{2-3}
                                                 & NerfCap\cite{paper125}      & \multicolumn{1}{m{12cm}}{A dynamic NeRF based method for human performance capture.}                                                                                                                                                    \\ \hline
    \multirow{1}{*}{\thead{Real-time \\ Human Rendering}}   & AvatarReX\cite{paper127}    & \multicolumn{1}{m{12cm}}{A combinatorial representation is proposed to create separate implicit fields for face, hand, and body parts.}                                                                                                                           \\ \cline{2-3}
                                                 & SAILOR\cite{paper128}       & \multicolumn{1}{m{12cm}}{3D reconstruction and rendering of the human body by fusing radiation and occupation fields.}                                                                                                                                           \\ \hline
    \multirow{3}{*}{\thead{Interactive Human \\Rendering}} & IntrinsicNGP\cite{paper136} & \multicolumn{1}{m{12cm}}{An intrinsic coordinate-based hash coding method for efficiently training NeRF for high-quality human interactive rendering.}                                                                                 \\ \cline{2-3}
                                                 & CFVVR\cite{paper195}        & \multicolumn{1}{m{12cm}}{A controlled free-viewpoint video reconstruction method based on neural radiation fields and motion maps.}                                                                                                                              \\ \hline
    \multirow{5}{*}{Others}                      & TransHuman\cite{paper122}   & \multicolumn{1}{m{12cm}}{A Transformer-based human representation for neural human rendering with high generalization capabilities.}                                                                                                                             \\ \cline{2-3}
                                                 & PersonNeRF\cite{paper126}   & \multicolumn{1}{m{12cm}}{The method reconstructs a 3D representation of an individual from a diverse collection of photos and enables the free-viewpoint rendering of the person.}                                                                                   \\ \cline{2-3}
                                                 & HandNeRF\cite{paper196}     & \multicolumn{1}{m{12cm}}{By introducing depth-guided density optimization and neural feature distillation strategies, high-fidelity rendering of animated gesture images and videos from arbitrary viewpoints is achieved.}                     \\ \hline                                     
    \end{tabular}
    \end{table*}
    

\subsubsection{Less View Human Rendering}
{\bf{OccNeRF}} \cite{paper193} (2023) enhances the accuracy of human body depiction in scenes with significant occlusion by employing a surface-oriented rendering approach that merges geometric and visibility precedents. Furthermore, a novel loss function has been implemented to preserve the geometric integrity in areas that are obscured. Trials using both simulated and actual occlusion data reveal that OccNeRF markedly surpasses current advanced methods in rendering accuracy, particularly in areas with occlusions. OccNeRF can lead to minor distortions owing to the necessity of fine-tuning additional parameters and diminishing training data because of occlusion. In terms of reasoning, OccNeRF operates at a slower pace compared to current techniques. Subsequent research has the potential to surmount these constraints through the enhancement of geometric prior and cross-scene training methodologies.

\begin{figure}[h!]
    \centering
    \includegraphics[width=0.45\textwidth]{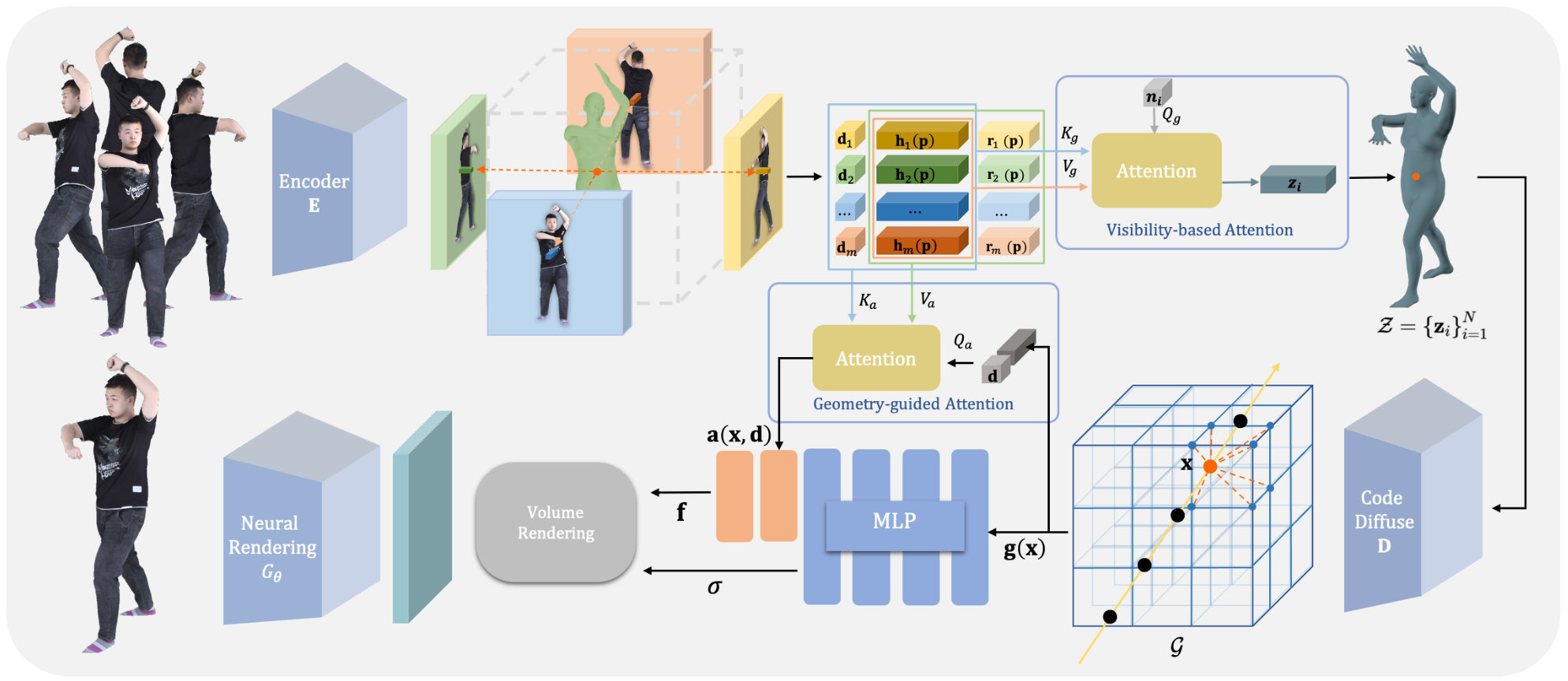}
    \caption{The Architecture of GM-NeRF Method.}
    \label{GM_NeRF}
\end{figure}

{\bf{GM-NeRF}} \cite{paper124} (2023) proposes a new generalized framework for synthesizing high-fidelity new-view images of arbitrary performers from multi-view images. The framework mitigates the misalignment between inaccurate geometric priors and pixel space by introducing a geometry-based attention mechanism that registers the appearance code of a multi-view 2D image to a geometric agent. In addition, the method employs neural rendering and partial gradient backpropagation to improve the perceptual quality of the synthesized images, the framework diagram of which is shown in Fig. \ref{GM_NeRF}. Experiments on the synthetic dataset THuman 2.0 \cite{paper055} show that GM-NeRF outperforms existing techniques in both new-viewpoint synthesis and geometric reconstruction.

{\bf{CHPC}} \cite{paper194} (2023) introduces an innovative method based on dual-layer NeRF to record the movements of a dynamic human body clad in a regular outfit. This technique adeptly replicates both dynamic costumes and human forms by independently monitoring the costume and human body movements, while simultaneously enhancing the deformation field and the standard dual-layer NeRF. Furthermore, the introduction of a garment simulation network, attuned to physical conditions, aids in creating realistic garment dynamics and interactions between the body and clothing. The experimental findings demonstrate the technique's superior capability in accurately recording body and clothing movements in dynamic videos, surpassing current methods. Given that the technique depends on the detailed visual limitations derived from the video, its effectiveness can be compromised in scenarios where the video input quality is subpar or the lighting is less than optimal.

\subsubsection{Real-time Human Rendering}
{\bf{AvatarReX}} \cite{paper127} (2023) introduces an innovative approach for mastering NeRF-based full-body avatars using multi-view video data. This technique offers not just the ability to expressively manipulate the body, hands, and face, but also facilitates live animation and rendering. The inherent structural design of parametric mesh templates is adeptly employed via a combinatorial avatar depiction, ensuring the representation's adaptability. Furthermore, this technique suggests a unique pipeline for delayed rendering, facilitating instantaneous rendering by distinguishing between geometry and visual aspects. The experimental findings indicate AvatarReX excels in creating synthesized outcomes with varied identities and manages an extensive array of clothing styles, body motions, and facial expressions.

{\bf{SAILOR}} \cite{paper128} (2023) introduces a technique that merges brightness with occupied field to record human actions instantaneously. This technique adeptly combines RGBD imagery from various perspectives through the creation of a dual-layered tree framework, ensuring superior rendering of fresh viewpoints. Additionally, SAILOR incorporates a deep noise-reducing network and a neural hybrid module to enhance rendering accuracy and minimize artifacts. SAILOR surpasses current techniques in the metrics of PSNR, SSIM, and LPIPS when applied to both the THuman 2.0 [55] dataset and actual captured data. Experimental findings indicate that with an increase in the number of input views, there's a gradual enhancement in the texture details rendered, particularly in areas that are invisible or smaller.

\subsubsection{Interactive Human Rendering}
\begin{figure}[h!]
    \centering
    \includegraphics[width=0.45\textwidth]{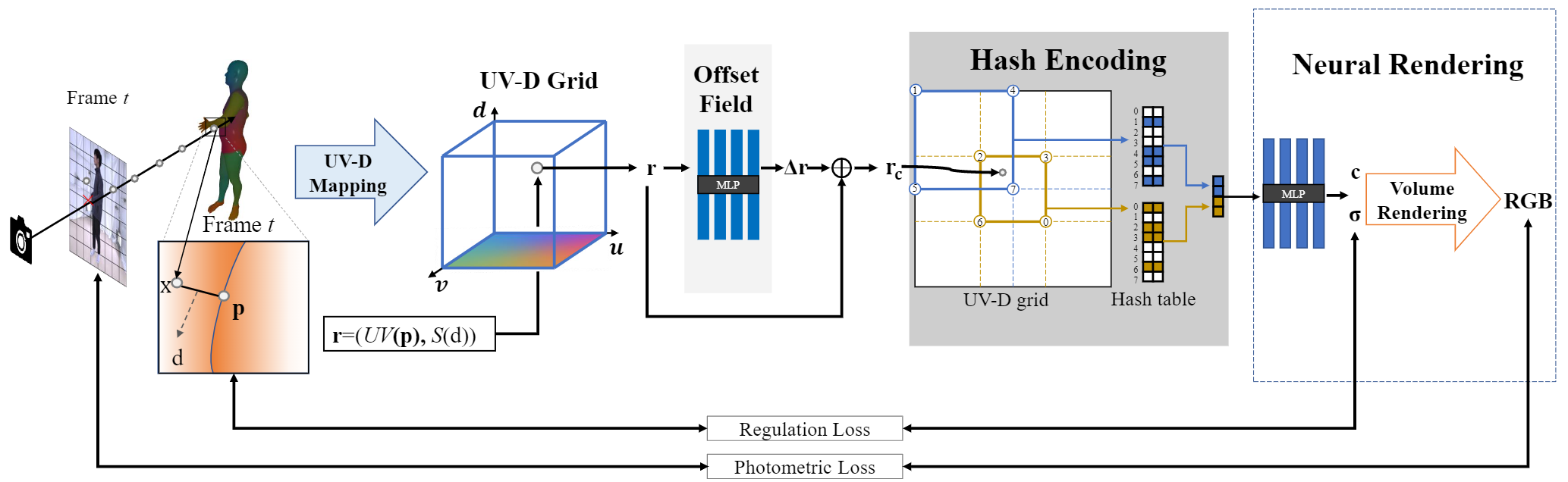}
    \caption{Overview of IntrinsicNGP Method.}
    \label{IntrinsicNGP}
\end{figure}

{\bf{IntrinsicNGP}} \cite{paper136} (2023) introduces a unique coordinate-driven hash coding technique for generating new perspectives in human NeRF. This technique consolidates data between frames on dynamic entities by incorporating continuous and optimizable intrinsic coordinates rather than direct Euclidean coordinates in InstantNGP. Furthermore, IntrinsicNGP incorporates a modifiable offset field reliant on intrinsic coordinates to enhance the outcomes, with a comprehensive summary of the model depicted in Fig. \ref{IntrinsicNGP}. In-depth experiments on various datasets reveal that IntrinsicNGP attains precise rendering of human performers' videos within minutes.

{\bf{CFVVR}} \cite{paper195} (2022) introduces a method for creating videos from a controlled free viewpoint, utilizing kinematic maps and neural radiation fields   (NeRF). This technique attains adaptable perspective management and streamlines the reconstruction process through the creation of a directed motion map for the captured sequence. Furthermore, the integration of overt surface distortion and subtle neural depiction of scenes leads to the suggestion of localized surface-directed NeRF, enhancing realism in rendering and editing body shapes in limited viewpoints. The experimental findings indicate that our suggested technique surpasses current NeRF approaches regarding the quality of rendering and manageability. Nonetheless, this technique primarily concentrates on adjustable free-viewpoint video rendering instead of neuralized body reconstruction, potentially resulting in rendering test actions that vary greatly from motion map actions.

\subsubsection{Others}
{\bf{TransHuman}} \cite{paper122} (2023) introduces a framework based on Transformer for human representation, aiming at broadening the scope of neural human rendering. This structure encapsulates the worldwide interconnections among human body components by manipulating the illustrated SMPL model in canonical space and altering it through Deformable Partial Radiance Fields in observation space for encoding query points. Furthermore, the Fine-grained Detail Integration (FDI) module enhances the rendering quality by merging detailed data from the reference image. TransHuman demonstrates notable improvements in pose generalization, identity generalization, and cross-dataset generalization, surpassing current methods.

{\bf{PersonNeRF}} \cite{paper126} (2023) proposes a method for reconstructing personalized 3D models from a collection of personal photographs, capable of rendering images with any combination of new viewpoints, body poses, and appearances. The method does this by constructing a shared motion weight field that allows changing the appearance under different viewing conditions while using a shared pose-dependent motion field. Compared to HumanNeRF \cite{paper197}, PersonNeRF achieves significant performance gains across all datasets, especially in visualization results, enabling the creation of consistent geometry, sharp details, and good rendering.

\begin{figure}[h!]
    \centering
    \includegraphics[width=0.45\textwidth]{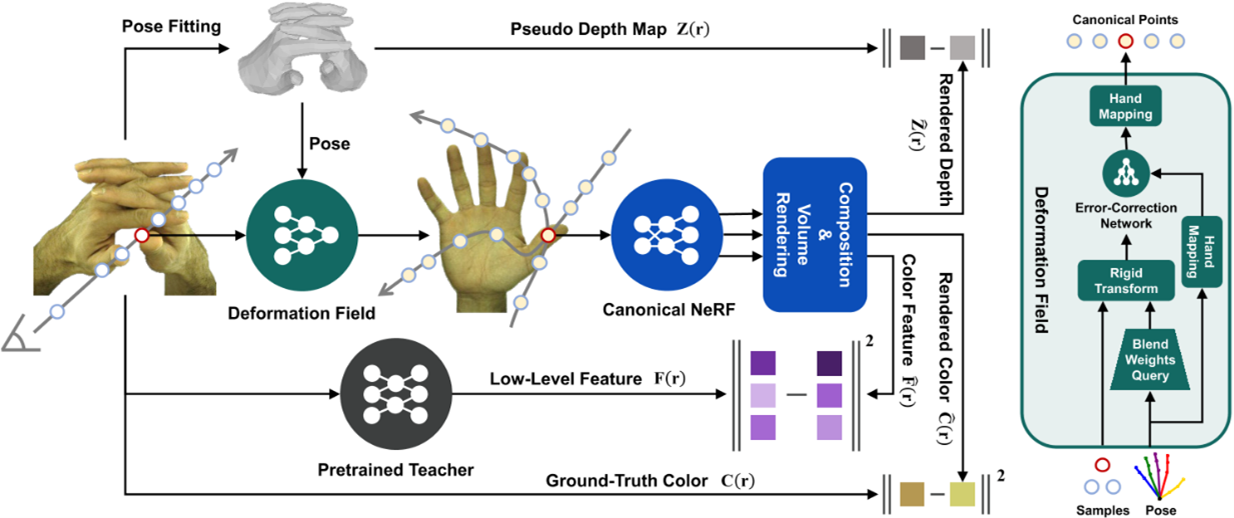}
    \caption{Overview of HandNeRF Method.}
    \label{HandNeRF}
\end{figure}

{\bf{HandNeRF}} \cite{paper196} (2023) introduces an innovative method for accurately replicating the shape and structure of a hand in interaction using the NeRF, facilitating the creation of lifelike-animated images and videos of hand movements from various perspectives. By integrating a pose-based deformation field, the system warps observational rays into a common canonical space, enhancing a pose-independent canonical hand NeRF within this space. Furthermore, this technique creates pseudo-depth maps based on the pose beforehand to aid in occlusion-aware density learning and introduces a neural feature distillation approach for optimizing color across different domains, with the model's framework depicted in Fig. \ref{HandNeRF}. In-depth tests using the extensive InterHand 2.6M \cite{paper060} dataset reveal HandNeRF's superiority over the standard method. Furthermore, by adapting to different poses, the model reduces the creation of artifacts and geometric inaccuracies in new poses, enhancing its efficiency in synthesizing new poses.

\subsection{Editable, interactive NeRF}
Traditional NeRF models are usually static, and once training is complete, the content and structure of the scene are fixed, making further editing and interaction difficult. To overcome this limitation, researchers have begun to explore editable and interactive NeRF models that allow users to make modifications and interact dynamically in the rendered scene, thus greatly expanding the scope of NeRF applications. The following are some of the key directions for improvement, as shown in Table \ref{tab:editable_interactive_table}.

\begin{table*}[!h]
    \centering
    \caption{ Overview of Editable, Interactive NeRF-Related Articles.}
    \label{tab:editable_interactive_table}
    \begin{tabular}{c||c||l} \hline
    Category                                  & Model              & \multicolumn{1}{c}{Highlight}                                                                                                                                                                                                                                                            \\ \hline
    \multirow{5}{*}{\thead{Text-based \\3D Generation}} & Magic3D\cite{paper129}            & \multicolumn{1}{m{12cm}}{An efficient text-to-3D content creation framework for fast and high-quality 3D mesh model generation through a two-stage optimization strategy using a low-resolution diffusion prior and a high-resolution potential diffusion model.}                                                  \\ \cline{2-3}
                                              & Latent-NeRF\cite{paper198}        & \multicolumn{1}{m{12cm}}{A Neural Radiation Field (NeRF) model for direct manipulation in potential space for text-guided 3D shape and texture generation.}                                                                                                                                                        \\ \cline{2-3}
                                              & Text2NeRF\cite{paper199}          & \multicolumn{1}{m{12cm}}{A text-driven 3D scene generation framework that combines NeRF and a pre-trained text-to-image diffusion model enables the generation of diverse 3D scenes with complex geometries and high-fidelity textures from textual cues.}                                                         \\ \hline
    \multirow{11}{*}{\thead{Text-to-3D \\Editing}}       & SKED\cite{paper130}               & \multicolumn{1}{m{12cm}}{A 3D editing technique based on sketches and textual cues is capable of generating localized and meaningful edits by utilizing a pre-trained NeRF and a small number of guided sketches.}                                                                                                 \\ \cline{2-3}
                                              & \thead{Instruct- \\NeRF2NeRF\cite{paper131}} & \multicolumn{1}{m{12cm}}{A text-command-based NeRF scene editing technique for consistent 3D editing of large-scale real-world scenes by iteratively updating the input image and optimizing the underlying scene.}                                                                                               \\ \cline{2-3}
                                              & LERF\cite{paper200}               & \multicolumn{1}{m{12cm}}{An approach that combines natural language processing with NeRF enables natural language querying and interaction with complex 3D scenes by embedding linguistic embedding of CLIP models in NeRF.}                                                                                       \\ \cline{2-3}
                                              & LENeRF\cite{paper201}             & \multicolumn{1}{m{12cm}}{A textual cueing-based 3D content editing framework for fine-grained and localized manipulation of 3D features by estimating 3D attention fields, supporting real-time editing, and maintaining high-quality visual effects.}                                                              \\ \cline{2-3}
                                              & NeRF-Art\cite{paper186}           & \multicolumn{1}{m{12cm}}{A text-driven NeRF-based stylization method enables simultaneous stylization of the appearance and geometric structure of pre-trained NeRF models by incorporating the semantic capabilities of CLIP models.}                                                                            \\ \hline
    \multirow{4}{*}{\thead{3D Color \\Editing}}         & ICE-NeRF\cite{paper134}           & \multicolumn{1}{m{12cm}}{Editing the color of a specific region is achieved by fine-tuning the weights of the model using a pre-trained model and a coarse mask provided by the user.}                                                                                                                             \\ \cline{2-3}
                                              & Seal-3D\cite{paper202}            & \multicolumn{1}{m{12cm}}{Interactive editing of the NeRF model was realized by designing multiple pixel-level editing tools.}                                                                                                                                                                    \\ \cline{2-3}
                                              & PaletteNeRF\cite{paper203}        & \multicolumn{1}{m{12cm}}{An editing technique for NeRF appearances utilizing 3D color breakdown.}                                                                                                                                                                                                                  \\ \hline
    \multirow{8}{*}{\thead{3D Geometric \\Editing}}     & ClimateNeRF\cite{paper204}        & \multicolumn{1}{m{12cm}}{A scene model editing framework that combines physical simulation and NeRF for generating realistic 3D consistent renderings of extreme weather effects such as haze, flooding, and snow.}                                                                                                 \\ \cline{2-3}
                                              & RO-NeRF\cite{paper205}            & \multicolumn{1}{m{12cm}}{The method guides 2D image restoration through a user-supplied mask and employs a confidence-based view selection process to ensure the consistency of the restored NeRF in 3D space.}                                                                                                    \\ \cline{2-3}
                                              & VQ-NeRF\cite{paper206}            & \multicolumn{1}{m{12cm}}{The continuous reflection field is discretized by a vector quantization mechanism, which improves the accuracy of reflection decomposition and facilitates material editing.}                                                                                                             \\ \cline{2-3}
                                              & Interactive NeRF\cite{paper207}   & \multicolumn{1}{m{12cm}}{User-controlled shape deformation of NeRF scene content is achieved by extracting the explicit triangular mesh representation and passing its deformation to the implicit volumetric representation while supporting real-time interactive editing and synthesis of novel perspectives.}  \\ \hline
    \end{tabular}
    \end{table*}
    

\subsubsection{Text-based 3D Generation}
{\bf{Magic3D}} \cite{paper129} (2023) introduces a dual-phase optimization approach that hastens the creation of coarse modes via a low-resolution diffusion prior and a thinly distributed 3D hash mesh framework, subsequently enhancing textured 3D mesh models through a high-resolution latent diffusion model. Utilizing this method, superior 3D mesh models are produced in under 40 minutes, doubling the speed of DreamFusion  \cite{paper208}, and attaining greater resolution. Nonetheless, this method depends on extensive datasets and substantial computational power, posing challenges in environments with limited resources.

\begin{figure}[h!]
    \centering
    \includegraphics[width=0.45\textwidth]{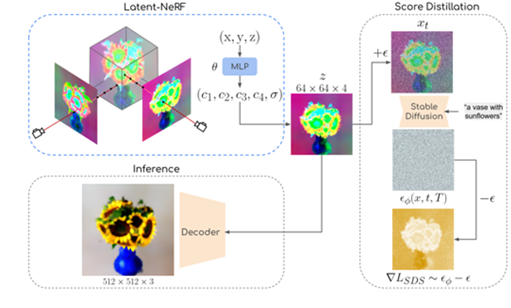}
    \caption{Overview of the Training Process for Latent-NeRF.}
    \label{Latent_NeRF}
\end{figure}

{\bf{Latent-NeRF}} \cite{paper198} (2023) represents a NeRF framework functioning directly within the realm of potential space for generating 3D shapes and textures guided by text. Introducing NeRF into a potential space eliminates the need to encode rendered RGB images into this space during each bootstrapping phase. Furthermore, Sketch-Shape and Latent-Paint, two types of shape bootstrapping, have been implemented to enhance control over the creation process. Fig. \ref{Latent_NeRF} illustrates the model's training methodology. Research indicates the efficacy of Latent-NeRF across various situations, yielding superior 3D models relative to techniques like DreamFusion. The RSketch-Shape and Latent-Paint approaches are superior in offering extra control, particularly in creating intricate geometric patterns that align with textual indicators.

{\bf{Text2NeRF}} \cite{paper199} (2023) introduces a framework for creating 3D scenes driven by text, integrating NeRF with an established text-to-image diffusion model. This technique can create a variety of three-dimensional settings featuring intricate geometric patterns and detailed textures based on textual indicators. The introduction of a progressive scene repair and update   (PIU) approach guarantees texture and geometric uniformity across various perspectives. In contrast to current techniques, Text2NeRF is superior in creating lifelike, multi-view uniform, and varied 3D environments.

\subsubsection{Text-to-3D Editing}
{\bf{SKED}} \cite{paper130} (2023) introduces a 3D editing method based on sketch-guided text, enabling users to alter an existing Neural Radiation Field   (NeRF) using at least two guided sketches from varied perspectives. This technique employs an already trained diffusion model to adhere to textual cue semantics, producing the intended edits through a unique loss function, all the while preserving the density and brightness of the base instances. SKED proves its proficiency in modifying numerous pre-trained NeRF instances, as evidenced by both qualitative and quantitative studies. This technique effortlessly incorporates diverse accessories, objects, and artifacts into the initial NeRF, with the method's components' contributions confirmed via quantitative analysis and ablation research.

{\bf{Instruct-NeRF2NeRF}} \cite{paper131} (2023) introduces a text-based command-driven NeRF scene modification technique, facilitating uniform 3D editing of actual scenes through repeated updates of input images and scene optimization. This technique employs the InstructPix2Pix \cite{paper209} model, enabling users to intuitively edit 3D scenes using natural language commands, eliminating the necessity for specialized tools or expertise. Research demonstrates Instruct-NeRF2NeRF's proficiency in executing diverse editing operations across multiple real-life scenarios, encompassing alterations in the environment and specific object modifications.

{\bf{LERF}} \cite{paper200} (2023) proposes an approach to embedding language into NeRFs that enables 3D scenes to respond to natural language queries by learning dense multi-scale linguistic fields within NeRFs. This approach does not need to rely on region proposals or masks, supports long-tailed open vocabulary queries, and can handle multiple types of linguistic queries such as visual attributes, abstract concepts, text, and long-tailed objects.LERF demonstrates its ability to handle natural language queries in a variety of real-world scenarios and significantly outperforms existing open vocabulary detection frameworks to localize the relevant parts of a 3D scene compared to the pixel-aligned LSeg. In addition, LERF can produce a more localized 3D relevance activation map in 3D compared to 2D CLIP embeddings.

\subsubsection{3D Color Editing}
{\bf{ICE-NeRF}} \cite{paper134} (2023) introduces a dynamic color modification system through the partial refinement of an already-trained NeRF model and a coarse mask provided by the user. This technique employs Activation Field Regularization (AFR) and Single Mask Multi-view Rendering (SMR) to tackle the issue of entanglement in implicit representations and the disruption of consistency across multiple views during the fine-tuning process. ICE-NeRF not only excels in decomposition and uniform color editing across multiple views but also markedly cuts down processing duration. Its effectiveness across various NeRF models is evident, with ICE-NeRF surpassing current NeRF color editing techniques in decomposition efficacy and time-saving. Despite ICE-NeRF's effectiveness in composite and forward-facing scenarios, it encountered unwanted color changes in 360-degree views.

{\bf{Seal-3D}} \cite{paper202} (2023) introduces an innovative interactive method and system for editing at the pixel level, aimed at implicit neural representations (like Neural Radiation Field NeRF) in 3D environments. Rapid responses to interactive editing and immediate previews are attained by aligning editing instructions with the NeRF model's initial space through a proxy function, utilizing a dual-phase training approach (inclusive of local pre-training and comprehensive fine-tuning). Seal-3D accommodates diverse editing styles, such as geometric (like bounding box, brush, and anchor tools) and color modification, without the need for a specific proxy framework, enhancing both interactivity and ease of use. The outcomes of experiments reveal that Seal-3D attains superior editing quality, demonstrating quicker convergence and immediate previews, all the while preserving the intricacies of the editing zone, in contrast to current techniques.

{\bf{PaletteNeRF}} \cite{paper203} (2023) introduces an innovative method for editing the appearance of NeRF using a palette approach. This technique breaks down a scene's visual elements into a collection of common palette bases via 3D color decomposition, enabling users to intuitively alter a 3D scene's look by adjusting the palette. Furthermore, this technique incorporates functions based on perspective to detect color discrepancies (like highlight shading), optimizes fundamental functions and color schemes during training, and introduces novel regularization elements to promote both sparsity and spatial uniformity in the breakdown. PaletteNeRF, in contrast to current NeRF coloration techniques, facilitates more detailed local color modification while preserving the authenticity of the original scene. Enhancements to the technique might be necessary to accommodate intricate lighting effects dependent on perspective, like specular reflections, and to broaden its application to dynamic NeRFs. Subsequent efforts could involve breaking down and modifying specular highlights based on frequency, and also broadening its scope to dynamic NeRFs.

\subsubsection{3D Geometric Editing}
{\bf{ClimateNeRF}} \cite{paper204} (2023) introduces an innovative approach that merges physical simulations and NeRF models to create authentic extreme weather scenarios. This technique attains a lifelike depiction of weather phenomena like floods, snow, and haze through enhanced geometric precision and the NeRF model's light field. Furthermore, ClimateNeRF offers management of variables with physical significance, like water levels, to mimic the impacts of climate change. ClimateNeRF's comparative analysis with current 2D image editing and 3D NeRF stylization methods reveals through its experiments on various datasets that its videos surpass other techniques in realism, perspective uniformity, and physical feasibility.

{\bf{RO-NeRF}} \cite{paper205} (2023) introduces a method for object removal from NeRF models, employing two-dimensional image restoration methods directed by a mask provided by the user. This algorithm operates on a confidence-driven view selection method, choosing the 2D restored image for NeRF creation, guaranteeing 3D consistency in the NeRF produced. This technique surpasses current methods by consistently creating satisfactory restoration images across multiple views. This technique outperforms standard methods in producing satisfactory scene completions by eliminating objects from the scene. The efficacy of this technique is constrained by the 2D restoration method's performance, which falters when the mask obscures a significant part of the trajectory, leading to a scarcity of realistic perspectives to select from.

{\bf{VQ-NeRF}} \cite{paper206} (2023) represents a dual-branch neural network framework integrated with vector quantization (VQ) to decompose and edit reflectivity fields in three-dimensional settings. Differing from conventional continuous models, VQ-NeRF breaks down the continuous reflectivity field into distinct substances via the VQ process, enhancing decomposition precision and streamlining the editing of materials. Furthermore, the paper presents a dropout-oriented VQ code word ranking approach to autonomously ascertain the scene's material count and minimize repetitiveness. Trials involving computer-generated (CG) and actual scenes reveal that VQ-NeRF outperforms in various functions, including reconstructing scenes, decomposing reflectivity, editing materials, and relighting scenes. VQ-NeRF stands out from current techniques for its precise breakdown of materials and simpler editing, particularly in scenarios involving intricate materials and lighting scenarios.

\subsection{NeRF for 3D Perception Techniques}
The NeRF model was initially proposed for 3D scene reconstruction and has shown significant potential for applications in this area. In recent years, researchers have begun to explore and develop NeRF-based frameworks for applications to more complex computer vision tasks such as 3D target detection and 3D segmentation.

{\bf{Instance-NeRF}} \cite{paper137} (2023) designs for a learning-oriented NeRF 3D instance segmentation process, can learn scene segmentation in 3D using a pre-trained NeRF of multiview RGB images, depicted through the NeRF model's instance field components. This technique utilizes a 3D proposal-driven mask prediction network to create distinct 3D instance masks and oversee the training of these masks by aligning them with 2D segmentation masks, thus facilitating the creation of uniform 2D segmentation maps and ongoing 3D segmentations for pure inference. Tests using both artificial and actual NeRF datasets reveal that Instance-NeRF excels in segmenting unseen views, surpassing earlier NeRF segmentation techniques and rival 2D segmentation approaches. This technique depends on an already trained NeRF model and necessitates extensive scene data for training to achieve generalization.

{\bf{Nerflets}} \cite{paper138} (2023) presents an innovative 3D depiction of scenes, featuring localized neural radiation fields trainable solely with 2D supervision. These Nerflets aid in reconstructing a scene's panoramic, density, and radial aspects by preserving their specific spatial locations, orientations, and distances. This method is suitable not just for 2D activities like creating new viewpoints and segmenting panoramas, but also for 3D endeavors including segmenting panoramas and engaging in interactive editing. Nerflets excel in various tasks like creating new viewpoints, synthesizing panoramic views, segmenting and reconstructing 3D panoramas, and engaging in interactive editing in both indoor and outdoor settings. Nerflets, in contrast to current NeRF techniques, provide notable benefits in efficiency and structural awareness, enable the extraction of panoramic and photometric renderings from various perspectives, and facilitate tasks uncommon in NeRF.

{\bf{FeatureNeRF}} \cite{paper139} (2023) acquires standard NeRF through the refinement of existing visual base models like DINO and Latent Diffusion. It leverages the conversion of 2D pre-trained models into 3D dimensions and extracts intricate features from NeRF MLPs using neural rendering, facilitating the forecasting of ongoing 3D semantic feature volumes from individual images for diverse subsequent tasks. FeatureNeRF proves to be efficient in extracting 3D semantic features for both 2D/3D semantic keypoint transfer and segmenting object parts. The experimental findings reveal that FeatureNeRF surpasses standard methods in these activities, particularly in the 3D task, proving that its acquired 3D model holds precise semantic data. Despite FeatureNeRF's notable advancements in 3D semantic comprehension, its dependence on extensive 2D image datasets for preliminary training could restrict its use in settings with limited data.

\subsection{NeRF-based Other Applications}
In addition to some of the applications mentioned above, NeRF has a wide range of applications in scenarios such as text steganography, style migration, and medical images, as shown in Table \ref{tab:other_applications_table}.

\begin{table*}[!h]
    \centering
    \caption{Other Applications}
    \label{tab:other_applications_table}
    \begin{tabular}{c||c||l} \hline
    Category                  & Model     & \multicolumn{1}{c}{Highlight}                                                                                                                                   \\ \hline
    Text Steganography        & StegaNeRF\cite{paper210} & \multicolumn{1}{m{12cm}}{An innovative steganography technique for embedding and recovering customized, imperceptible information in Neural Radiation Field (NeRF) renderings while maintaining the original visual quality of the rendered image provides a new approach to copyright protection and ownership marking of NeRF content.} \\ \hline
    \thead{NeRF Copyright \\Protection} & CopyRNeRF\cite{paper211} & \multicolumn{1}{m{12cm}}{A novel digital watermarking technique realizes effective copyright protection of NeRF models by embedding copyright information in NeRF models while guaranteeing the rendering quality and improving the robustness of extracted information under different rendering strategies and   2D distortions.}      \\ \hline
    Medical Field             & CuNeRF\cite{paper142}    & \multicolumn{1}{m{12cm}}{The CuNeRF framework is proposed to realize arbitrary-scale super-resolution reconstruction of medical images by using techniques such as cube sampling, isotropic volume rendering, and cube-based hierarchical rendering.}                                                                                     \\ \hline
    Continuous Learning       & CLNeRF\cite{paper143}    &\multicolumn{1}{m{12cm}}{ A new dataset, World Across Time (WAT), and an efficient sequential learning system, CLNeRF, are presented, which allows step-by-step learning from sequential scene scans without the need to store historical images.}                                                                      \\ \hline                   
    \end{tabular}
\end{table*}
    

\section{Discussion}\label{section:discussion}
Neural Radiance Fields, as an emerging 3D scene representation method, has attracted much attention in the fields of computer vision and graphics. However, despite its remarkable achievements in rendering quality and detail, NeRF still faces a series of challenges that point to future directions.

\subsection{Discussion on Computational Efficiency}
With the progression of deep learning methods, it's anticipated that future research will concentrate on enhancing the computational effectiveness of NeRF and similar techniques. Such research could investigate innovative sampling methods, enhance network configurations, incorporate pre-existing geometric understanding, and create algorithms for more efficient rendering. Going forward, the pursuit of computational efficiency will concentrate on enhancing the speed of rendering and diminishing the resource usage of NeRF.

Researchers might investigate improved sampling and integration techniques to lessen the computational demands for each image rendering. As an illustration, NerfAcc[88] integrates various sampling techniques using a unified transmittance estimator, enabling quicker sampling speeds and lower rendering quality.

Conversely, further studies are anticipated to concentrate on enhancing network configurations, like MIMO-NeRF \cite{paper097}, through the application of multiple-input-multiple-output (MIMO), multilayer perceptrons (MLPs), aiming to decrease the frequency of MLP operations during the rendering process, thereby boosting the overall rendering speed. Furthermore, integrating the latest developments in deep learning, including Transformer architecture and unsupervised learning methods, could pave the way for enhanced efficiency in NeRF.

\subsection{Discussion on Less View Rendering}
Presently, the domain of combining less-view and single-view views is swiftly expanding as a focal point in the study of computer vision and graphics. The advent of methods like NeRF has enabled scientists to create superior 3D imagery from a significantly restricted set of perspectives. Even with NeRF's impressive multi-view synthesis feats, its efficacy remains constrained for limited or single views due to inadequate training data, potentially resulting in overfitting and errors in geometric reconstruction.

Contemporary research is delving into diverse regularization techniques to enhance synthesis quality amidst a lack of data. As an illustration, research has been undertaken to improve model generalization through the implementation of geometric priors (GeoNeRF \cite{paper213}), the use of Generative Adversarial Networks (GAN) (PixelNeRF \cite{paper214}), or the enhancement of rendering methods (ViP-NeRF \cite{paper156}). While these methods have achieved advancements in decreasing the duration of training and enhancing the quality of rendering, they continue to encounter obstacles like sparse views, managing occlusions, and recovering geometric details. Subsequent studies might concentrate on creating more effective training approaches, enhancing network structures for improved scene detail capture, and investigating both unsupervised and self-supervised learning techniques to lessen dependence on extensive labeled data. Furthermore, the integration of physical simulation with scene comprehension in hybrid methodologies could lead to novel advancements in the domain, enhancing areas like virtual reality, augmented reality, and self-driving vehicles.

\subsection{Discussion on Rendering Quality}
Regarding the quality of rendering, contemporary research focuses on two primary categories high-resolution rendering and the generalization potential of models. The task of handling extensive data and computational tasks while preserving intricate details remains a significant hurdle in creating high-resolution, high-quality images (such as those exceeding 4K) through model optimization. UHDNeRF \cite{paper073} and RefSR-NeRF \cite{paper076} regarding the quality of rendering, contemporary research focuses on two primary categories of high-resolution rendering and the generalization potential of models. The task of handling extensive data and computational tasks while preserving intricate details remains a significant hurdle in creating high-resolution, high-quality images (such as those exceeding 4K) through model optimization. UHDNeRF and RefSR-NeRF refine their network structures to enhance the model's detection accuracy. However, UHDNeRF elevates the model's rendering by merging explicit and implicit scene depictions, boosting detail efficiency in 4K UHD resolution, whereas RefSR-NeRF amplifies NeRF's high-frequency details by incorporating high-resolution reference images for creating super-resolution views. Regarding its capacity to generalize, NeRF's proficiency in handling unfamiliar scenes and data is constrained, necessitating enhancement through improved network design and training approaches. NeRF-SR mitigates this issue by boosting NeRF's efficiency on new views via oversampling and combined optimization techniques, while NeRFies bolsters the model's generalization by incorporating adaptable neural radiation fields for dynamic scenarios. capability.076 refine their network structures to enhance the model's detection accuracy. However, UHDNeRF elevates the model's rendering by merging explicit and implicit scene depictions, boosting detail efficiency in 4K UHD resolution, whereas RefSR-NeRF amplifies NeRF's high-frequency details by incorporating high-resolution reference images for creating super-resolution views. Regarding its capacity to generalize, NeRF's proficiency in handling unfamiliar scenes and data is constrained, necessitating enhancement through improved network design and training approaches. NeRF-SR mitigates this issue by boosting NeRF's efficiency on new views via oversampling and combined optimization techniques, while NeRF bolsters the model's generalization by incorporating adaptable neural radiation fields for dynamic scenarios. capability.

\subsection{Discussion on Imaging Barriers}
Regarding the enhancement of imaging obstacles, the primary focus of researchers has been on tackling the challenge of processing objects that are reflective and transparent. Given that NeRF often results in fuzzy or warped images when interacting with objects that exhibit reflective or transparent characteristics. In response to this challenge, MS-NeRF \cite{paper172} and Ref-NeRF \cite{paper095} tackle the issue by addressing the issue of multi-view consistency. MS-NeRF manages reflective and transparent elements by depicting the scene as a feature field with several parallel areas, whereas Ref-NeRF enhances the structured and parameterized representation of reflection direction based on view by incorporating NeRF's capability to manage reflective surfaces, leading to more accurate rendering outcomes. To address a broader spectrum of rendering issues in intricate lighting scenarios, including dynamic range illumination, shadows, and overall lighting impacts, further studies and methods might be required to enhance NeRF's efficiency. Subsequent research should investigate techniques for integrating precise physical lighting models with NeRF and also create fresh datasets and assessment standards to evaluate and confirm the efficacy of these approaches in intricate lighting scenarios.

\subsection{Discussion on Application Scenarios}
Regarding real-world uses, recent research has primarily concentrated on interactive rendering, crafting portraits and faces, and the authentic reconstruction of scenes, as elaborated below:

\subsubsection{Interactive rendering techniques}
Present studies in interactive rendering methods concentrate on boosting the efficiency of rendering, enriching the user's editing process, and broadening the scope of multimodal interaction features. Nonetheless, several obstacles and constraints persist in these domains. Enhancements are still required in the intuitiveness and adaptability of user editing interfaces to enable lay users to execute proficient editing tasks without intricate training.

Regarding multimodal interaction, enhancing the integration of various inputs like text, images, and audio into research is crucial for a more intuitive and natural editing process. Furthermore, current methods continue to struggle with broad applicability, potentially diminishing the model's flexibility and the quality of editing for unfamiliar scenes and objects.

Future studies could investigate these avenues to tackle these issues. Initially, enhancements in rendering's real-time and efficiency can be achieved through optimization algorithms like NerfAcc \cite{paper088} and by employing more effective hardware acceleration methods, including GPUs and TPUs. Furthermore, enhancing the design of the user interface for better intuitiveness and ease of use can reduce the difficulty for users to edit, thereby boosting both the precision and contentment in editing, as implemented by ICE-NeRF \cite{paper134} and NaviNeRF \cite{paper132}. Enhancing the model's multimodal fusion features is feasible, enabling it to more effectively comprehend and react to various inputs. Ultimately, to enhance the model's ability to generalize and sustain superior rendering and editing across various applications, it might be necessary to build datasets across different domains, implement meta-learning methods, and innovate in the regularization techniques of models.

Through these efforts, future interactive rendering technologies will be able to better meet user needs and provide more powerful and flexible tools for a wide range of application areas.

\subsubsection{Portrait Reconstruction}
Face synthesis technology holds great potential in the future, particularly for improving realism and the experience of user interaction. The emergence of techniques like FaceCLIPNeRF \cite{paper118} highlights the capability of text-based descriptions in accurately managing 3D facial expressions and characteristics. Not only does this method retrieve data from still pictures, but it also preserves uniformity from various perspectives, paving the way for crafting customized media content. Conversely, the NeRFInvertor \cite{paper119} method showcases the creation of superior animation of authentic identities from a solitary image, offering significant potential for use in gaming, film, and virtual reality. Furthermore, the creation of GazeNeRF \cite{paper187} demonstrates the ability to alter facial attributes, like eye positioning, using 3D perception methods to improve the interactivity and authenticity of virtual characters. Ultimately, the RODIN framework introduces innovative opportunities for generating and modifying digital avatars via 3D diffusion networks, enhancing the efficiency of crafting bespoke and high-accuracy 3D characters. Such advancements in technology signal upcoming advancements in face synthesis, focusing on real-time processing, diversity, and tailoring for users, yet they also introduce fresh challenges regarding privacy safeguards and ethical considerations.

\subsubsection{Human Rendering}
Presently, the domain of human rendering is experiencing a twofold growth, encompassing both technological advancements and the broadening of applications. Initially, from a technical standpoint, new research findings like TransHuman \cite{paper122} and GM-NeRF \cite{paper124} exemplify frameworks showcasing superior new-view synthesis, even with limited data, by educating conditional NeRFs using videos from multiple views. Not only do these methods improve the immediacy and broad applicability of rendering, but they also offer robust technical assistance for immediate applications like virtual reality (VR) and augmented reality (AR). Furthermore, methods like PersonNeRF \cite{paper126} allow for the customization of visuals from various perspectives, stances, and looks through the creation of individualized 3D models using a set of personal photos, offering a novel, personalized approach for social media, digital entertainment, and e-commerce.

Secondly, regarding the growth of applications, the progression in human rendering technology is catalyzing transformations across various sectors. The SAILOR framework, for instance, offers not just superior rendering effects but also grants users editorial and creative liberty, granting content creators increased creative room and the capacity to craft more varied and detailed visual content. Furthermore, as data compression and transmission technologies progress, it's anticipated that future human rendering will facilitate effective data transfer in network settings with limited bandwidth, thereby ensuring the smooth operation of superior VR and AR experiences on mobile devices. The progress indicates a growing significance of human rendering technology in providing engaging experiences and tailored content, introducing novel applications across various sectors such as entertainment, education, and healthcare.

Numerous obstacles confront the NeRF domain, yet it harbors immense growth prospects. With ongoing technological progress, the significance of NeRF in shaping the future of 3D scene modeling and rendering is set to rise.

\section{Conclusion}\label{section:conclusion}
Following Mildenhall et al.'s proposal of the NeRF framework. The model, in its groundbreaking research, has significantly enhanced various aspects including processing velocity, output integrity, and training data needs, thereby surpassing numerous constraints of its original form. The NeRF method's success is due to its capacity to recreate a continuous 3D landscape from limited perspectives and produce superior images from diverse viewpoints. The advent of this technology brings a novel aspect to the realm of computer vision. This innovation paves the way for innovative approaches in viewpoint synthesis, 3D reconstruction, and neural rendering within computer vision, with NeRF technology demonstrating significant promise in diverse areas like style migration, image editing, avatar development, and 3D urban environment modeling. As NeRF modeling garners increasing attention in academic and industrial circles, a multitude of researchers have allocated substantial research resources, aiding in the release of various preprints and scholarly works. This document methodically examines NeRF's latest advancements in technology and its real-world uses, offering a thorough examination and perspective on its prospective paths and challenges. The focus of this document is to inspire scholars in this domain, aiming to foster ongoing advancements and innovations in NeRF-related technologies.


\bibliographystyle{IEEEtran}
\bibliography{bibfile.bib}

\vfill

\end{document}